\documentclass{article} 
\usepackage{iclr2025_conference,times}

\usepackage{etoolbox}
\newbool{singcol}
\setbool{singcol}{false}  

\newenvironment{figurecustom}
{%
    \ifbool{singcol}{\begin{figure*}[!t]}{\begin{figure}[H]}%
}
{%
    \ifbool{singcol}{\end{figure*}}{\end{figure}}%
}

\usepackage{balance} 
\usepackage{hyperref}       
\usepackage{url}            
\usepackage{booktabs}       
\usepackage{amsfonts}       
\usepackage{nicefrac}       
\usepackage{microtype}      
\usepackage{xcolor}         
\usepackage{amsmath}        
\usepackage{amsthm}         
\usepackage{tabularx}
\usepackage{graphicx}
\usepackage{subcaption}
\usepackage{tikz}
\usetikzlibrary{positioning}
\usepackage{wrapfig} 
\usepackage{pifont}
\usepackage{textcomp}
\usepackage{float}
\usepackage{enumitem}
\usepackage{multicol}
\usepackage{placeins}

\newcommand{\BibTeX}{\rm B\kern-.05em{\sc i\kern-.025em b}\kern-.08em\TeX}

\newcommand{\E}{{\mathcal{E}}}

\newcommand{\Pp}{{\mathcal{P}}}
\newcommand{\Ss}{{\mathcal{S}}}

\newcommand{\argmax}{\mathop{\mathrm{argmax}}}

\newtheorem{lemma}{Lemma}
\title{Noisy Zero-Shot Coordination: Breaking The Common Knowledge Assumption In Zero-Shot Coordination Games
}

\author{
Usman Anwar\textsuperscript{\textnormal{1}†},
Ashish Pandian\textsuperscript{\textnormal{2}†},
Jia Wan\textsuperscript{\textnormal{3}},
David Krueger\textsuperscript{\textnormal{4}}
Jakob Foerster \textsuperscript{\textnormal{5}}\\
\textsuperscript{1}University of Cambridge,
\textsuperscript{2}UC Berkeley,
\textsuperscript{3}MIT,\\
\textsuperscript{4}MILA \& Université de Montréal,
\textsuperscript{5}University of Oxford.\\
\textsuperscript{†}Equal contribution.
}

\preprintcopy
\begin{document}
\maketitle

\begin{abstract}
Zero-shot coordination (ZSC) is a popular setting for studying the ability of reinforcement learning (RL) agents to coordinate with novel partners. Prior ZSC formulations assume the \textit{problem setting} is common knowledge: each agent knows the underlying Dec-POMDP, knows others have this knowledge, and so on ad infinitum. However, this assumption rarely holds in complex real-world settings, which are often difficult to fully and correctly specify. 
Hence, in settings where this common knowledge assumption is invalid, agents trained using ZSC methods may not be able to coordinate well. 
To address this limitation, we formulate  the \textit{noisy zero-shot coordination} (NZSC) problem. In NZSC, agents observe different noisy versions of the ground truth Dec-POMDP, which are assumed to be distributed according to a fixed noise model. Only the distribution of ground truth Dec-POMDPs and the noise model are common knowledge. We show that a NZSC problem can be reduced to a ZSC problem by designing a meta-Dec-POMDP with an augmented state space consisting of all the ground-truth Dec-POMDPs. For solving NZSC problems, we propose a simple and flexible meta-learning method called NZSC training, in which the agents are trained across a distribution of coordination problems – which they only get to observe noisy versions of. We show that with NZSC training, RL agents can be trained to coordinate well with novel partners even when the (exact) problem setting of the coordination is not common knowledge\footnote{Code: \url{https://github.com/ashishp166/Noisy-Zero-Shot-Coordination}}.

\end{abstract}
\newcommand{\zsc}[0]{\text{ZSC}}

\section{Introduction}

Cooperation and coordination are central to human society \citep{editorial2018cooperative, smith1997major}. Humans are able to work together with novel partners, even under uncertain conditions, to achieve common goals. This is evident in a variety of human activities, from driving a car to playing a team sport to carrying out moonshot projects such as the Apollo Program \citep{pennisi2005did,melis2010human}. While AI has had many surprising successes in recent years, developing AI agents that can successfully coordinate with novel partners in complex environments remains an outstanding challenge in artificial intelligence research~\citep{open_problems_in_coop_ai}.

\newcommand{\figurehspace}{
 \ifbool{singcol}{\hspace{10pt}}{\hspace{80pt}}
}

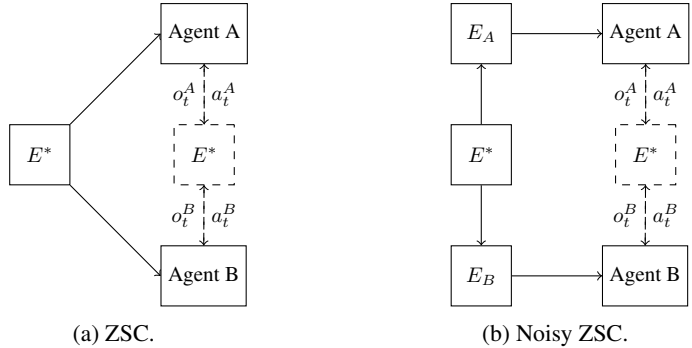
\begin{figure}[t]
\ifbool{singcol}{}{\centering}
\begin{subfigure}{0.20\textwidth}
\begin{tikzpicture}[scale=0.8, transform shape, node distance=1cm and 1.5cm]
\node[draw, minimum size=1cm] (E) {$E^*$};
\node[draw, minimum size=1cm, above right=of E] (A) {Agent A};
\node[draw, minimum size=1cm, below right=of E] (B) {Agent B};
\node[draw, dashed, minimum size=1cm, below=of A] (Estar) {$E^*$};
\draw[->] (E) -- (A.west);
\draw[->] (E) -- (B.west);
\draw[->, dashed] (Estar) -- node[left] {$o^A_t$} (A);
\draw[->, dashed] (A) -- node[right] {$a^A_t$} (Estar);
\draw[->, dashed] (Estar) -- node[left] {$o^B_t$} (B);
\draw[->, dashed] (B) -- node[right] {$a^B_t$} (Estar);
\coordinate[right=1.5cm of E] (X);
\coordinate[right=4.0cm of E] (Y);
\end{tikzpicture}
\caption{ZSC.}
\label{fig:sub1}
\end{subfigure}
\figurehspace
\begin{subfigure}{0.20\textwidth}
\begin{tikzpicture}[scale=0.8, transform shape, node distance=1cm and 1.5cm]
\node[draw, minimum size=1cm] (E) {$E^*$};
\node[draw, minimum size=1cm, above=of E] (EA) {$E_A$};
\node[draw, minimum size=1cm, below=of E] (EB) {$E_B$};
\node[draw, minimum size=1cm, right=of EA] (A) {Agent A};
\node[draw, minimum size=1cm, right=of EB] (B) {Agent B};
\node[draw, dashed, minimum size=1cm, below=of A] (Estar) {$E^*$};
\draw[->] (E) -- (EA.south);
\draw[->] (E) -- (EB.north);
\draw[->] (EA) -- (A.west);
\draw[->] (EB) -- (B.west);
\coordinate[right=1.5cm of E] (X);
\coordinate[right=4.0cm of E] (Y);
\draw[->, dashed] (Estar) -- node[left] {$o^A_t$} (A);
\draw[->, dashed] (A) -- node[right] {$a^A_t$} (Estar);
\draw[->, dashed] (Estar) -- node[left] {$o^B_t$} (B);
\draw[->, dashed] (B) -- node[right] {$a^B_t$} (Estar);
\end{tikzpicture}
\caption{Noisy ZSC.}
\label{fig:sub2}
\end{subfigure}
\caption{In zero-shot coordination, the environment $E^*$ is assumed to be common knowledge (CK). In noisy zero-shot coordination, agents still have to act and coordinate in $E^*$ but $E^*$ is no longer CK. Instead, each agent has a distinct (private) model of the problem setting which is assumed to be a noisy copy of $E^*$.}
\label{fig:main}
\end{figure}

A setting that has become popular in recent times to study coordination between novel partners is zero-shot coordination~\citep[ZSC;][]{hu2020other, OBL}.
In ZSC, the central idea is to modify the training process in such a way that any two independently trained agents end up learning the mutually compatible (meta) conventions and thus are able to coordinate successfully at test time. 
However, ZSC makes the strong and unrealistic assumption that all the coordinating agents have complete and precise knowledge of the \textit{problem setting} \citep{hu2020other, treutlein2021new, OBL}.
In other words, the problem setting is assumed to be \textit{common knowledge} (CK).
In \textit{games} like Overcooked~\citep{carroll2019utility}, or Hanabi~\citep{bard2020hanabi} -- which are commonly studied benchmarks for ZSC~\citep{hu2020other,lupu2021trajectory,zhao2021maximum,OBL} -- this assumption is indeed valid.
However, this assumption is clearly not valid for most real-world settings as it is often difficult to correctly \& fully specify a complex problem setting~\citep{amodei2016concrete, hendrycks2021unsolved, goal_misgeneralization_lauro}.
For instance, consider a disaster response scenario where multiple AI-powered robots from different manufacturers are deployed to assist in an earthquake-stricken area.
Unlike controlled game environments, disaster zones present complex, dynamic settings that cannot be fully specified in advance.
Each robot may have slightly different sensor capabilities or interpretations of the environment, violating the common knowledge assumption crucial to traditional ZSC approaches.

In this work, we demonstrate that when common knowledge assumption does not hold, agents trained to coordinate under the standard ZSC setup may perform poorly.
To address this limitation, we propose a novel problem formulation called \textit{noisy zero-shot coordination} (NZSC).
In NZSC, the agents still have to solve the coordination problem in the same problem setting, but agents do not get to observe it directly. Instead, each agent observes a perturbed (``noisy'') version of the problem setting with the perturbation sampled from that agent's noise model.
In contrast to ZSC where the entire problem setting is common knowledge, NZSC assumes that only the underlying problem distribution and noise models are common knowledge.
Each agent then privately observes its own problem instance, sampled from this shared distribution and perturbed by the agent's noise model.
To refer back to our disaster response example, while ZSC would assume that \textit{all} robots observe the same coordination problem, NZSC makes the more realistic assumption that different robots will observe different versions of the coordination problem, which are derived by perturbing the coordination problem according to some noise model.

Our main contributions are:

\begin{itemize}[noitemsep,topsep=0pt,parsep=0pt,partopsep=0pt]
\item We argue that zero-shot coordination is possibly an unrealistic model for real-world coordination problems between novel agents due to assuming that the problem setting for the coordination problem is common knowledge across all (coordinating) agents.
\item We introduce a novel problem formulation -- the noisy zero-shot coordination (NZSC) problem -- which removes the assumption that the problem setting is CK for all agents attempting to coordinate zero-shot. 
\item We show that NZSC can be reduced to a standard ZSC problem. By making use of this reduction, we show that existing ZSC algorithms can in principle be repurposed for solving NZSC problems.





\end{itemize}

\section{Related Work}
\textbf{ZSC \& Ad-Hoc Teamwork}: The Zero-shot coordination (ZSC) problem was introduced by \citet{hu2020other} who then proposed \textit{other-play} as a solution to the ZSC problem. \citep{treutlein2021new} provided a formal treatment of this setup as the \textit{label-free coordination} problem.
Separately, \citet{bullard2021quasi} focused on identifying symmetries in a given Dec-POMDP to learn a coordination strategy that may be resilient to the arbitrary breaking of symmetries.
\textit{Off-belief learning} (OBL) promotes ZSC by preventing agents from learning arbitrary conventions by replacing the real rewards seen during a transition, by reward from a fictitious trajectory sampled from a belief model which is independent of the policy\citep{OBL}. However, OBL can be highly inefficient
\citep{nekoei2023towards}. 

Number of prior works have taken inspiration from human learning and reasoning processes to propose methods for ZSC. \citet{cui2021k} use K-level reasoning \citep{costa2006cognition} to train ZSC capable policies in Hanabi \citep{bard2020hanabi}.
\citet{Ma2022LearningTC} show that attention-based architectures have a better inductive-bias towards ZSC.
\citet{yu2023learning} show that assuming humans are biased, and training RL agents to coordinate with similarly biased policies enables better ZSC performance.
Other works \citep{carroll2019utility,jacob2022modeling,bakhtin2022mastering} use human-data in various ways to learn policies that perform better at ZSC with humans.
Language has also emerged as a popular medium to regularize policies towards behaviors that are more favored by humans~\citep{hu2023language,guan2023efficient}.


A closely related setting to zero-shot coordination is ad-hoc teamwork \citep{ad_hoc_teamwork_survey, barrett2011empirical, barrett2015cooperating}. The goal in ad-hoc teamwork is to train an agent which does well when placed in a team of unknown agents \citep{AAAI10-adhoc}. This ultimately amounts to learning the best response to a population of agents. However, training a diverse population remains an open challenging problem and has attracted considerable attention \citep{strouse2021collaborating, lupu2021trajectory, cui2023adversarial, zhao2021maximum, lou2023pecan, szot2023adaptive, hammond2024symmetry, yan2024efficient}. Different from all the above mentioned works, we consider the challenge of zero-shot coordination when the problem setting is not common knowledge.

\textbf{Robustness To Misspecification In RL}: Within single agent reinforcement learning, many works consider the misspecification in reward function \citep{clark2016faulty,krakovna2020specification,pan2022effects,skalse2022defining} and propose algorithms to sidestep its effects \citep{shah2019preferences, hadfield2016cooperative, malik2021inverse}. 
Robust reinforcement learning considers the misspecification in environment dynamics \citep{roy2017reinforcement, panaganti2022robust}; techniques such as domain randomization \citep{tobin2017domain, zhao2020sim} and unsupervised environment design~\citep{dennis2020emergent} have been proposed to counter this misspecification.
While several works study robust MARL, the common forms of robustness studied are to adversarial partners \citep{shen2019robust, li2023byzantine}, state uncertainties \citep{he2023robust} and partner agent's policies \citep{li2019robust, sun2022romax, van2020robust}.
However, in this work, we study the misspecification in the \textit{problem setting} for different agents with a focus on ZSC, which to the best of our knowledge has not been studied previously.

\section{Background}
\label{sec:background}
\textbf{Multi-Agent Reinforcement Learning:} A general fully cooperative multi-agent reinforcement learning problem (MARL) with $n$-agents is represented as a Decentralized Partially Observable Markov Decision Process (Dec-POMDP) \citep{marl-book}, which is defined as a tuple $\langle D, S, \mathcal{A}, I, T, \{O^i\}_{1:n}, \{\Omega^i\}_{1:n}, R, h \rangle$. Here, $D$ refers to the set of $n$ agents involved in the game, $S$ is the finite set of states over which the Dec-POMDP is defined, $\mathcal{A}$ is the shared action space and $T:S \times \mathcal{A} \times S \rightarrow [0,1]$ is the state transition function that maps a state-action pair $(s,a)$ to the probability of reaching a new state $s'$, i.e., $T(s,a,s')=Pr(s'|s,a)$. $I\in \Pp(S)$ refers to the initial state distribution at stage $t=0$.  $O^i$ and $\Omega^i$ respectively represent the set of observations and the observation function for any agent $i$. $R$ represents the common reward function for all agents. The variable $h$ represents the horizon of the game.

In each round of the game, every agent $i$ receives an observation $o_t^i \in O^i$. Based on their respective action-observation history, players select actions $a^i_t\sim\pi_i(\cdot|o^i_t, o^i_{t-1}, a_{t-1},..., a_1, o^i_{0})$, then each player receives a reward $r^i_t(s_t,a_t=a^1_t, \ldots, a^n_t)$, and the game transitions into a new state $s_{t+1}$. Each player $i$ aims to maximize the expected reward of the team, and the optimal policy for player $i$, $\pi^*_i$, usually depends on the policies $\pi_{-i} = \{\pi_1,\ldots,\pi_{i-1},\pi_{i+1},\ldots,\pi_n\}$ of the other players.

\textbf{Common Knowledge}: Common knowledge (CK) for a group of agents consists of facts that all
agents know and ``each individual knows that all other individuals
know it, each individual knows that all other individuals know that
all the individuals know it, and so on'' \citep{osborne1994course}. Specifically, in MARL, for a problem setting, described by some Dec-POMDP $E$ to be CK means that all agents have the \textit{same} knowledge of $E$; and that everyone knows that everyone has the same knowledge of $E$ and so on.

\section{Noisy Zero-Shot Coordination}
\label{sec:noisy_zsc}
A core part of the problem formulation of ZSC is the assumption that the ground truth Dec-POMDP in which agents need to coordinate is CK~\citep{treutlein2021new}.
For agents trained using RL, this is equivalent to assuming that all the independently trained agents have an identical (and exact) \textit{copy} of the environment simulator.
In practice, if agents are trained in a simulator, then this will have to be realized by having a prior agreement among all parties training the agents on the implementation details of the simulator (or exact sharing of the code).
This is hard to achieve when there are several different parties involved, as is assumed in ZSC, who are not known to each other.
In contrast, it is more realistic to assume that all parties training their agent share only a high-level understanding of the problem but are not able to share low-level details of their respective training environments -- resulting in different agents (trained by different principals) training on different \textit{noisy} specifications of the problem setting but still having to act and coordinate in the (unknown) ground truth Dec-POMDP at \textit{test time}.

To address this, we propose the problem setting of \textit{noisy zero-shot coordination} (NZSC) which removes the CK assumption and leads to a more realistic ZSC problem setting.
In NZSC, agents are not informed about the \textit{ground-truth} Dec-POMDP they are acting in at test time.
Instead, the agents each observe a different \textit{noisy} Dec-POMDP which reveals partial information about the ground-truth Dec-POMDP.
However, the distribution over ground truth Dec-POMDPs and the noise models (used to generate noisy Dec-POMDPs given a ground truth Dec-POMDP) are assumed to be CK.
This is in line with the spirit of ZSC where agents may share high-level details, but are disallowed any agreements on low-level details.


\subsection{NZSC Problem Formulation}
Formally, consider a distribution $P(E)$ over the space of ground truth Dec-POMDPs $\mathcal{E}$ where 
$E_i \in \mathcal{E}$ is given by the tuple
$\langle D, S, \mathcal{A}, I,$ 
$T, \{O^i\}_{1:n}, \{\Omega^i\}_{1:n}, {R^i}_{1:n}, h \rangle$\footnote{See Section~\ref{sec:background} for a more descriptive presentation of Dec-POMDP.}. 
Further, let an agent $A_i$'s noise model be $P^{A_i}(E_i|E^*)$ which gives a distribution over the \textit{noisy} Dec-POMDP that $A_i$ might observe given that the ground truth Dec-POMDP is $E^*$.
Given a ground truth Dec-POMDP $E^* \sim P(E)$, an agent $A_1$ observes $E^{A_1} \sim P^{A_1}(E_1|E^*)$, agent $A_2$ observes $E^{A_2} \sim \Pp^{A_2}(E^{A_2}|E^*)$ and so on. 
While both the distribution over ground truth Dec-POMDPs $\Pp(E)$ and noise models of all agents are CK - an agent only knows the concrete realization of its own noisy Dec-POMDP i.e., $\{E_{i}\}_{1:n}$ are \textit{not} CK. The goal in noisy-ZSC is for agents $\{A_i\}_{1:n}$ to coordinate zero-shot in the ground truth Dec-POMDP $E^*$ at test-time.

We next present the reduction of noisy ZSC to standard ZSC. To simplify the presentation, we restrict our presentation to the two agent setup (agents $A$ and $B$) here and further assume that all the Dec-POMDPs under consideration have identical state space and horizon. In appendix \ref{appendix:theory_stuff}, we relax these assumptions.

\subsection{Reduction of NZSC to ``Standard'' ZSC}
\label{subsec:reduction_to_zsc}
Standard ZSC assumes that all agents at test time act in a Dec-POMDP which is CK to all agents during their independent training. We show that NZSC can be reduced to a standard ZSC problem by creating a CK meta Dec-POMDP. This meta Dec-POMDP has an augmented state space and observation space, covering the space of Dec-POMDPs $\mathcal{E}$. Specifically, the aforementioned two-agent NZSC problem is equivalent to standard ZSC with both the agents acting in the following (CK) meta-Dec-POMDP 
$M = \langle \tilde D, \tilde S, \tilde I, \mathcal{\tilde A}, \{\tilde O^A, \tilde T, \tilde O^B\}, \{\tilde \Omega^A, \tilde \Omega^B\}, \tilde R, \tilde h \rangle$ where
\begin{itemize}
\item $\tilde D = \{A,B\}$
\item $\tilde S = S \times E$ is the state space of the meta Dec-POMDP
\item $\tilde I = I(S)P(E)$
\item $\mathcal{\tilde A} = \mathcal{A}$
\item $\tilde{T}: \tilde S \times \mathcal{\tilde A} \times \tilde S \rightarrow [0, 1]$ is the transition function where 
    \begin{align*}
        \tilde{T}(\tilde s, a, \tilde s') = \tilde{T}[(s,E_i),a,(s',E_j)] = \begin{cases}
        T_i(s,a,s') & i=j \\
        0 & i \ne j
        \end{cases}
    \end{align*}
\item $\mathcal{\tilde O}^A = \mathcal{O}^A \times E$ is the observation space for agent $A$. Observation space for agent $B$ is defined similarly.
\item $\tilde{\Omega}^A: \tilde{\mathcal{O}}^A \times  \tilde{S}  \rightarrow [0,1]$ gives the observation probability function for agent $A$ where 
    \begin{align*}
        \tilde{\Omega}^A(\tilde o, \tilde s, a) &= \tilde{\Omega}^A((o, E_i), (s,E_j),a) \\&= 
        \begin{cases}
        [\Omega^A_{E_i}(o,s,a), E_A] & i=j \\
        \text{undefined} & i \ne j
        \end{cases}
    \end{align*}
    where $E_A$ is the noisy Dec-POMDP observed by agent $A$ and $\Omega^A_{E_i}$ denotes the observation function of agent $A$ in the Dec-POMDP $E_i \in \mathcal{E}$. The observation function for agent $B$, $\tilde \Omega^B$, is defined similarly as follows 
    \begin{align*}
        \tilde{\Omega}^B(\tilde o, \tilde s, a) = 
        \begin{cases}
        [\Omega^B_{E_i}(o,s,a), E_B] & i=j \\
        \text{undefined} & i \ne j
        \end{cases}
    \end{align*} 
    where $E_B$ is the noisy Dec-POMDP observed by agent $B$. 
\item $\tilde R: \tilde S \times A \times \tilde S \rightarrow (-\infty, \infty)$ is the reward function given by 
   \begin{align*}
        \tilde{R}(\tilde s, a, \tilde s') = \tilde{R}[(s,E_i),a,(s',E_j)] = \begin{cases}
        R_i(s,a,s') & i=j \\
        \text{undefined} & i \ne j
        \end{cases}
    \end{align*}
\item $\tilde{h} = h$ 
\end{itemize}

Intuitively, this reduction works by creating a new problem setting that is CK, where respective observations of noisy Dec-POMDPs $E_A$ and $E_B$ are modeled as the private knowledge of agents $A$ and $B$ and the ground truth Dec-POMDP is modeled as part of the (augmented) state space. Agents then keep a belief over what the true Dec-POMDP may be which they refine as they interact with the ground truth Dec-POMDP at test time. In other words, we convert uncertainty over problem setting, which is novel and not previously studied within MARL, into uncertainty over state space, which is often encountered in MARL. 

The main benefit of this reduction is that it allows the reuse of reinforcement learning algorithms developed for standard ZSC problem \textit{out of the box} for NZSC. We exploit this by using IPPO-OP, a standard ZSC algorithm \citep{hu2020other}, in our experiments.

\section{Methods}
As discussed in the previous section, we can attempt to solve NZSC through repurposing standard ZSC methods.
In this section, we present two methods for solving NZSC problems -- noisy other-play (NOP) and noisy maximum entropy population-based training (NMEP).
These methods are respectively based on popular ZSC methods -- other-play~\citep[OP;][]{hu2020other} and maximum entropy population-based training~\citep[MEP;][]{zhao2021maximum}.


Formally, when solving a coordination problem $E \sim P(E)$, the ZSC methods optimize the following objective where $P(\pi_2)$ is a distribution over the opposing policies that the learning policy $\pi_1$ is trained to coordinate with, and $J(\pi_1, \pi_2, E)$ denotes the expected return of policies $\pi_1$ and $\pi_2$ when evaluated on the coordination problem $E$.
\begin{align}
    \argmax_{\pi_1} \mathbb{E}_{E \sim P(E)} \big[ \mathbb{E}_{\pi_2 \sim P(\pi_2)}\big[J(\pi_1(\cdot | E), \pi_2(\cdot | E), E)\big]\big].
    \label{eq:zsc}
\end{align}

We can similarly write the NZSC training objective as follows:

\ifbool{singcol}%
{\begin{align}
    \argmax_{\pi_1} \quad &\mathbb{E}_{E \sim P(E), E_1 \sim P(E_1|E), E_2 \sim P(E_2|E)} \Big[ \nonumber \\
    &\mathbb{E}_{\pi_2 \sim P(\pi_2)} \Big[ J\big(\pi_1(\cdot|E_1, P(E_1 | E), P(E_2 | E)), \nonumber \\
    & \hspace{5.5em} \pi_2(\cdot|E_2, P(E_1 | E), P(E_2 | E)), E\big) \Big] \Big]. \tag{2}
    \label{eq:nzsc}
\end{align}}%
{\begin{align}
    \argmax_{\pi_1} \quad \mathbb{E}_{E \sim P(E), E_1 \sim P(E_1|E), E_2 \sim P(E_2|E)} \Big[ 
    &\mathbb{E}_{\pi_2 \sim P(\pi_2)} \Big[ J\big(\pi_1(\cdot|E_1, P(E_1 | E), P(E_2 | E)), \nonumber \\
    & \pi_2(\cdot|E_2, P(E_1 | E), P(E_2 | E)), E\big) \Big] \Big]. \tag{2}
    \label{eq:nzsc}
\end{align}}
As we discussed before, in NZSC, the agents do not get to observe $E$ directly but rather observe its noisy versions $E_1$ and $E_2$ generated from the respective noise models $P(E_1|E)$ and $P(E_2|E)$ of the two agents.
This is reflected in the NZSC objective given in which the policies are conditioned on their observed noisy version and the noise models.
This objective is not directly amenable to ZSC methods that assume the coordination problem is common knowledge.
However, by applying the reduction given in the previous section, we can create a meta-DecPOMDP $M$ for any $E \sim P(E)$, $E_1 \sim P(E_1|E)$ and $E_2 \sim P(E_1|E)$.
Let $P(M)$ be the corresponding distribution over meta-DecPOMDPs; then, we can rewrite the above NZSC objective as follows:
\ifbool{singcol}{}{\vspace{2pt}}
\begin{align}
    \argmax_{\pi_1} \mathbb{E}_{M \sim P(M)} \big[ \mathbb{E}_{\pi_2 \sim P(\pi_2)}\big[J(\pi_1(\cdot | M), \pi_2(\cdot | M), E)\big]\big].
    \label{eq:nzsc.modified}
\end{align}
\ifbool{singcol}{}{\vspace{2pt}}
We now describe the two methods we use in this work for training agents under NZSC problem formulation.
To emphasize the point that we apply these methods to meta-Dec-POMDPs,
we call these methods NZSC methods, but functionally they remain identical to their ZSC counterparts.

\textbf{Noisy-Other-Play}:
When we create a meta-Dec-POMDP $M$ from some Dec-POMDP $E$ and its noisy versions $E_1, E_2$, $M$ inherits the symmetries present in $E, E_1$ and $E_2$.
These symmetries could potentially inhibit coordination among novel partners if broken arbitrarily by them.
Noisy-Other-Play (NOP), similar to Other-Play \citep[OP;][]{hu2020other}, enables agents to learn symmetry invariant strategies by randomizing over the known symmetries during training.
We test NOP in a one-shot lever game and iterated lever game (see Section~\ref{sec:environments} for details on these games).
We provide a discussion of symmetries in the lever game in Appendix~\ref{appendix:environments}.

\textbf{Noisy-MEP}: MEP is a popular ZSC method that works by first training a diverse population of agents, and then training the ZSC agent to learn to coordinate with all the agents present in the population~\citep{zhao2021maximum}.
In order to train a diverse population of agents, MEP trains agents (in self-play setting) using an augmented reward $R(s_t, a_t) - \alpha \log \left(\frac{1}{n} \sum_{i=1}^n \pi^{(i)}(a_t|s_t)\right)$ where $\{pi_1, ..., \pi_N\}$ is the current population of the agents.
Once the population has been trained, the ZSC agent is trained by prioritizing training against agents that are hard to collaborate with.
Noisy-MEP (NMEP) works similarly, but with one main difference: we use a curriculum-based approach in which the learning agent is initially trained in a noise-free setting, and then we gradually increase the noise to the maximum level throughout the training.
We found this modification crucial as without it the agents failed to learn performant coordination policies.

\textbf{Meta-NZSC}:
In some of our experiments, we consider more complex noise models, e.g., hierarchical noise models.
Specifically, we assume that there is a meta-distribution over the noise models.
When sampling a noisy observation of the Dec-POMDP, we assume that first a noise model is sampled from this distribution, and then the noisy observation of the ground-truth Dec-POMDP is sampled from this noise model.
As the hierarchical noise models can still be expressed as distributions over the noisy versions of Dec-POMDPs, no modification to the approaches described above is required.
However, for ease of reference, we would describe training when using hierarchical noise models as meta-NZSC training.

\textbf{Training Details}: For all our experiments, we use an LSTM-based network architecture and use Independent Proximal Policy Optimization (IPPO) as the reinforcement learning algorithm. We defer further details on our training setup to Appendix~\ref{appendix:training_details}.

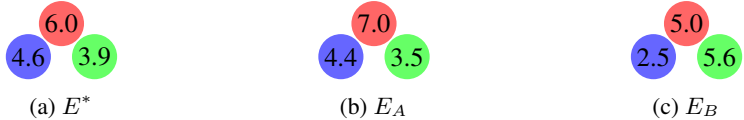
\begin{figure}[t]
\centering
\begin{subfigure}{.31\columnwidth}
\centering
\begin{tikzpicture}[scale=0.22,main_node/.style={circle,fill=blue!60,minimum size=0.6cm,inner sep=0pt}] 
    \node[main_node] (1) at (0,0) {4.6};
    \node[main_node,fill=red!60] (2) at (2,2) {6.0};
    \node[main_node,fill=green!60] (3) at (4,0) {3.9};
\end{tikzpicture}
\caption{$E^*$}
\end{subfigure}%
\hspace{-0.5em}%
\begin{subfigure}{.31\columnwidth}
\centering
\begin{tikzpicture}[scale=0.22,main_node/.style={circle,fill=blue!60,minimum size=0.6cm,inner sep=0pt}] 
    \node[main_node] (11) at (0,0) {4.4};
    \node[main_node,fill=red!60] (21) at (2,2) {7.0};
    \node[main_node,fill=green!60] (31) at (4,0) {3.5};
\end{tikzpicture}
\caption{$E_A$}
\end{subfigure}%
\hspace{-0.5em}%
\begin{subfigure}{.31\columnwidth}
\centering
\begin{tikzpicture}[scale=0.22,main_node/.style={circle,fill=blue!60,minimum size=0.6cm,inner sep=0pt}] 
    \node[main_node] (12) at (0,0) {2.5};
    \node[main_node,fill=red!60] (22) at (2,2) {5.0};
    \node[main_node,fill=green!60] (32) at (4,0) {5.6};
\end{tikzpicture}
\caption{$E_B$}
\end{subfigure}
\caption{In the noisy lever game, depicted here, there are three levers corresponding to three different reward values. Agents get the reward if they both pull the same lever. (a) shows the ground truth game $E^*$. (b) shows $E_A$, the noisy version of $E^*$ observed by player A. Similarly, (c) shows $E_B$, the noisy version of $E^*$ observed by player B.}
\label{fig:nlg_game}
\end{figure}
\section{Environments}
\label{sec:environments}
To better study the effects of misspecification of problem setting on the coordination performance of ZSC agents, we design multiple environments with different types of misspecifications.
In the one-shot noisy lever game, and the iterated lever game agents have to deal with uncertainty over the payoffs.
In the coordinated exploration environment, agents have to deal with uncertainty over the speed (operationalized as maximum timesteps an agent can take within an episode) of the partner agent.
In the battleship environment, agents have to deal with uncertainty over the observation function of the partner agent.

\subsection{One-Shot Noisy Lever Game}
This is a simple one-step lever game (depicted in Figure~\ref{fig:nlg_game}) consisting of three levers; 
with each lever $i$ having an associated reward value $R_i^*$ sampled from a predefined distribution $P(R_i)$ at the start of every round.
However, in the noisy Dec-POMDP observed by agents $A_1$ and $A_2$, the reward values are corrupted according to public noise models $P^{A_1}(R_i|R_i^*)$ and $P^{A_2}(R_i|R_i^*)$.
The challenge for the agents is to coordinate on pulling the same lever $i$ to earn reward $R_i^*$.
Pulling different levers results in a penalty set to some constant $c$ ($-2$ in our experiments).
By default, we set $P(R_i^*)$ to be the univariate normal distribution with mean $r_{mean}=5$ and standard deviation $\sigma^*=2$ and the noise models to be additive normal noise with mean $0$ and standard deviations $\sigma_1$ and $\sigma_1$ respectively.
Specifically,  $P^{A_1}(R|R_i^*) = R_i^* + \epsilon^{A_1}; \epsilon^{A_1} \sim N(0, \sigma^A)$ with $P^{A_2}$ defined similarly.
This models a commonly found situation where coordinating agents have the option to pursue multiple goals but true payoffs of the goals are unknown and every agent observes a different noisy payoff.
For example, a team of product designers trying to coordinate which new feature they should prioritize building, or researchers within a team trying to coordinate on their next project given multiple options.

\subsection{Iterated Noise Lever Game (I-NLG)}
This is an extension of NLG to a longer horizon - which is set to be $16$ in our experiments. Identical to NLG, agents get to observe noisy Dec-POMDPs at the first timestep.
For all subsequent timesteps, agents get to observe the action the other agent took at the last timestep.

\subsection{Coordinated Exploration Environment}
The Coordinated Exploration Environment (CEE) (shown in Figure~\ref{fig:cee.environment}) is an 8x8 gridworld environment that posits a multi-agent explore or exploit dilemma to the agents.

The environment contains four mines; in every episode, three of these mines are placed at random locations throughout the gridworld but the location of the fourth mine is kept fixed in every episode.
Mining the fourth mine always gives a small constant reward value of $1$. 
But for the other three mines, the reward values are sampled every episode from normal distributions with means of $10$, $10$, and $20$, respectively, and a standard deviation of $2$. Mining requires both agents to be on the same mine square.
The reward values observed by the agents are corrupted by additive noise sampled from standard Gaussian with a standard deviation of $2$.

Agents begin at the start position (0, 0) and operate with partial observability, perceiving only a 3x3 subgrid centered around their current location. 
They observe the other agent, and noisy reward value for any mine, only when the other agent and the mine are within their 3x3 subgrid.
Different agents may have different `speeds' at which they move through the environment.
Each agent knows its own speed, but does not know what the speed of the other agent (it is coordinating with) is a priori.
We simulate different speeds by fixing the maximum number of actions agents can take in a fixed time-window, while the horizon for the environment is kept fixed at $32$.
Specifically, we have agents with three speed levels: low, medium, and high. High speed agents take one action for every environment timestep. Medium speed agents take one action every two environment timesteps.
Low speed agents take one action every four environment timesteps.
In the `empty' timesteps, the agent is forced to do a stay action.
Mismatch between the speeds at which human players and AI agents act is one of the reasons that AI agents trained in isolation can fail to coordinate with human players in Overcooked~\citep{carroll2019utility,yu2023learning}.

Agents with low speed tend to have insufficient time to explore the environment, so, are better-off simply exploiting and mining the low-reward mine.
Agents with higher speeds, however, are better off exploring and finding the high-reward mine.
Furthermore, the environment contains a special key which if collected by both agents causes rewards to be multiplied by three times.
In order to collect the key, agents need to be simultaneously positioned on the square containing the key.

However, the key is placed in the environment to be farthest away from the agents, hence, agents need to make a strategic decision about whether collecting the key is feasible or not based on their own and the partner's speed.
\usetikzlibrary{positioning,calc}

\newcommand{\figurescale}{%
    \ifbool{singcol}{0.5}{0.7}
}

\begin{figure}
\centering
\begin{tikzpicture}[scale=\figurescale]
    \draw (0,0) grid (8,8);
    \fill[blue] (2,2) rectangle (3,3);
    \node[white] at (2.5,2.5) {A1};
    \fill[purple] (5,5) rectangle (6,6);
    \node[white] at (5.5,5.5) {A2};
    
    \node[black, font=\tiny] at (0.5, 0.8) {1};
    \node at (0.5, 0.3) {\includegraphics[scale=0.014]{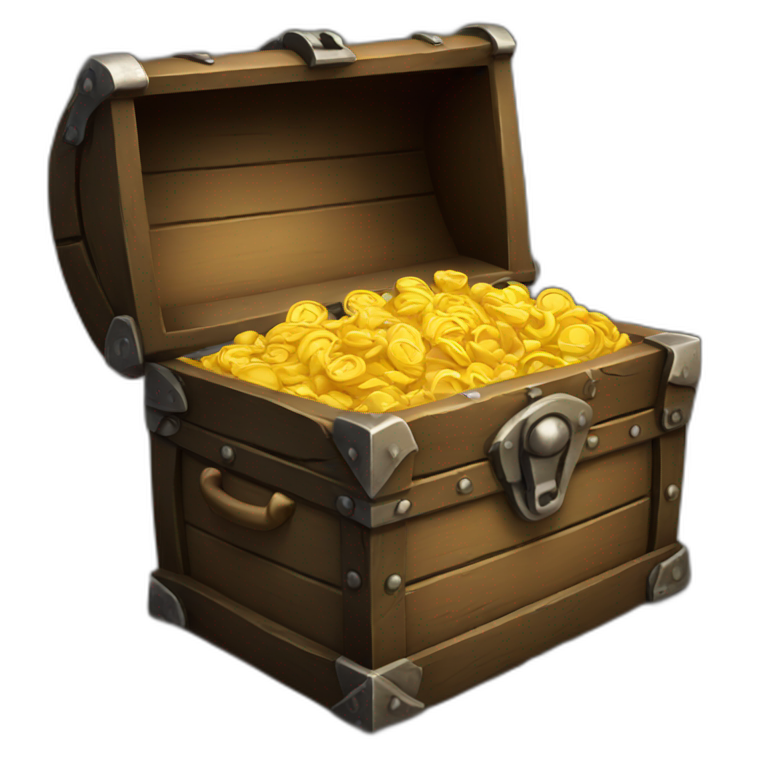}};
    \node at (5.5, 3.3) {\includegraphics[scale=0.014]{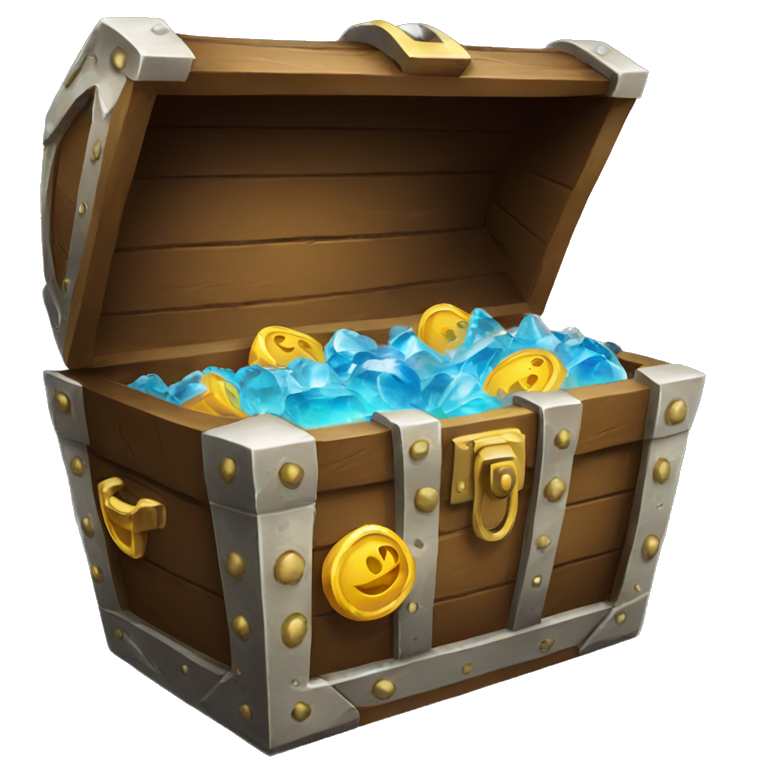}};
    \node[black, font=\tiny] at (5.5, 3.8) {8.93};
    \node at (2.5, 6.3) {\includegraphics[scale=0.014]{figures/env_pics/treasure-chest-random.png}};
    \node[black, font=\tiny] at (2.5, 6.8) {22.34};
    \node at (4.5, 0.3) {\includegraphics[scale=0.014]{figures/env_pics/treasure-chest-random.png}};
    \node[black, font=\tiny] at (4.5, 0.8) {11.54};
    \node at (7.5, 0.5) {\includegraphics[scale=0.02]{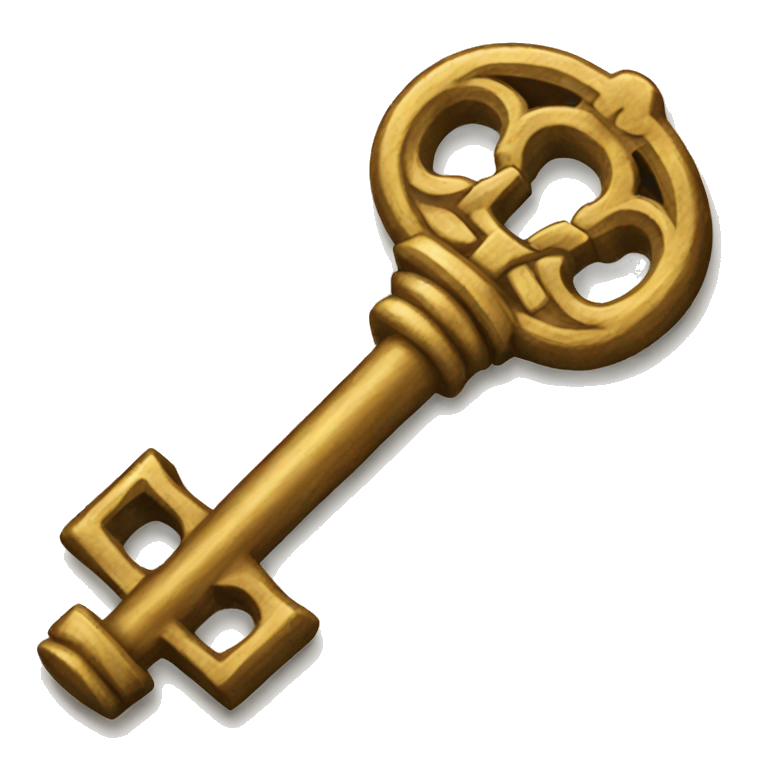}}; 
    
    \foreach \i in {1,...,3} {
        \foreach \j in {1,...,3} {
            \fill[gray!30] (\i,\j) rectangle (\i+1,\j+1);
        }
    }
    \fill[blue] (2,2) rectangle (3,3);
    \node[white] at (2.5,2.5) {A1};
    
    \foreach \i in {4,...,6} {
        \foreach \j in {4,...,6} {
            \fill[gray!30] (\i,\j) rectangle (\i+1,\j+1);
        }
    }
    \fill[purple] (5,5) rectangle (6,6);
    \node[white] at (5.5,5.5) {A2};
    
    \draw[->, thick] (0.3, 7.5) -- (0.7, 7.5);
\end{tikzpicture}
    \caption{Visualization of Coordinated Exploration Environment (CEE). The agent start position is denoted by $\rightarrow$, while gold chests denote different mines. The mine in the bottom left gives reward of $1$ when mined and is always located in the same square. While for the other three mines, the reward values are sampled randomly from normal distribution with means $20$, $10$ and $10$ and the location of these mines is randomized in every episode. The bottom right corner contains the key which if collected allows agents to $3$x any future rewards collected. The agents observe 3x3 gird around them.}
    \label{fig:cee.environment}
\end{figure}

\subsection{SyncSight Environment}
SyncSight Environment (SSE) (shown in Figure~\ref{fig:synsight.environment}) simulates a visual coordination game in which the agents must balance maximizing their potential reward with maintaining mutual visibility to successfully coordinate their actions. 
It is structured as a 3x12 grid with a central barrier, creating two 3x6 subgrids.
Two agents are confined to their respective subgrids, unable to cross the barrier.
For clarity, we label the columns in each subgrid from 1 to 6, with column 1 being closest to the barrier.

Each grid square is associated with a random reward value, sampled at the start of the episode. The reward structure is mirrored on either side of the barrier, meaning corresponding squares in both subgrids have identical values. Reward values for different columns are sampled from normal distributions with a standard deviation of 2 but varying means. The mean increases by 2 for each column moving away from the barrier, starting at 4 for column 1 and ending at 24 for column 6.
Agents observe noisy reward values where noise is sampled from standard normal with standard deviation of $2$.

An agent's observation is determined by its view size, which can range from 2 to 12. Agents always face the barrier and can observe squares and (noisy) reward values in up to n columns in front of them, where n is their view size. Each agent is assigned a secret token value, randomly chosen at each time step from a set of 9 predefined values. Agents always know their own secret token but can only observe the other agent's token when it is within their view. Agents know their own view size but observe a noisy version of the other agent's view size. This noisy observation is sampled from a categorical distribution that assigns $0.5$ probability to the true view size $n$, $0.15$ probability each to view sizes $n-1$ and $n+1$ and $0.1$ probability to view sizes $n-2$ and $n+2$. At each time step, agents decide on two actions: a movement direction (up, down, left, right, or stay) and a guess of the opposing agent's secret token.

Agents receive a small penalty of -1 for non-coordination at each time step, unless they occupy corresponding squares in their subgrids and correctly predict each other's secret tokens. In this case, they receive the reward associated with their current square. The reward structure incentivizes agents to coordinate on squares farther from the barrier, where expected rewards are higher. However, moving too far apart may prevent them from observing each other's tokens, resulting in failed coordination. The optimal strategy balances maximizing distance from the barrier while remaining within each other's view.
\ifbool{singcol}{}{\begin{figure}
    \centering
\begin{tikzpicture}[scale=0.7]
    \draw (0,0) grid (12,3);
    \newcommand{\mylist}{24,20,16,12,8,4,4,8,12,16,20, 24}
    \foreach \val [count=\xi from 0] in \mylist {
        \node[font=\Large] at (\xi+0.5,3.3) {\val};
    }
    \fill[blue!40] (2,0) rectangle (4,3);
    \foreach \x in {4,...,8} {
        \fill[purple!40] (\x,0) rectangle (\x+1,3);
    }
    \draw (0,0) grid (12,3);
    \fill[black!60] (5.9,-0.1) rectangle (6.1,3.1);
    \draw[very thick, black] (6,0) -- (6,3);
    \fill[blue] (2,0) rectangle (3,1);
    \node[white] at (2.5,0.5) {A1};
    \fill[red] (8,1) rectangle (9,2);
    \node[white] at (8.5,1.5) {A2};
    \node[below] at (3,0) {Agent 1 Side};
    \node[below] at (9,0) {Agent 2 Side};
\end{tikzpicture}
\caption{A depiction of SyncSght Environment (SSE). Black line in the middle is the impermissible barrier that divides the grid into two symmetric subgrids. The agent A1 is shown to have view size of $2$, while the agent A2 is shown to have view size 5. The numbers at the top of each column correspond to the mean of the normal distribution from which the reward values for the squares in that column are sampled.}
\label{fig:synsight.environment}
\end{figure}
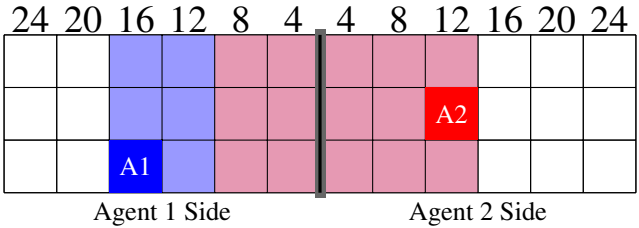}

\vspace{-2pt}
\section{Experiments \& Results}
As is standard in the ZSC literature, we use cross-play performance as a proxy for ZSC performance of the trained agents.
In cross-play (XP), two independently trained agents are evaluated as a team at test time.
Since in this work we are primarily interested in the ability of the agents to coordinate under noise, we evaluate XP performance across noise models.
In Figures~\ref{fig:self-play}~-~\ref{fig:meta-nzsc}, we respectively show returns for agents trained using self-play, NZSC and meta-NZSC.
Due to the space constraints, we use two models per noise model, and consider three distinct noise models per environment.
In Appendix~\ref{appx:sec.additional_results}, we provide additional results which show both return as well as final step coordination rate for $3$ agents per noise model and a larger number of noise models where possible.
In Figures~\ref{fig:self-play}~-~\ref{fig:meta-nzsc}, we use tick-labels to indicate the noise model of the agent.

In NZSC training (and self-play training) for OS-NLG and I-NLG, we consider a fixed noise model in which $\sigma_1$ and $\sigma_2$ values are kept fixed throughout training.
However, in meta-NZSC training, we uniformly sample $\sigma_1$ and $\sigma_2$ from the range $0-5$.
Similarly, in NZSC training (and self-play training) for CEE and SIE environments, the agents have to contend with (different) noisy observations of the reward values but the agents are only trained with homogeneous partner agents that have the same speed and view size respectively.
In meta-NZSC training, however, the agents are trained with additional noise over the speed and the view sizes of the partner agent for the two environments.

\textbf{Agents trained via self-play perform poorly}:
In Figure~\ref{fig:self-play}, we plot cross play return for agents trained via self-play.
As is expected for self-play trained agents, in all environments, these agents coordinate well with themselves, but generally coordinate poorly with other agents.

\ifbool{singcol}{\begin{figure}
    \centering
\begin{tikzpicture}[scale=0.7]
    \draw (0,0) grid (12,3);
    \newcommand{\mylist}{24,20,16,12,8,4,4,8,12,16,20, 24}
    \foreach \val [count=\xi from 0] in \mylist {
        \node[font=\Large] at (\xi+0.5,3.3) {\val};
    }
    \fill[blue!40] (2,0) rectangle (4,3);
    \foreach \x in {4,...,8} {
        \fill[purple!40] (\x,0) rectangle (\x+1,3);
    }
    \draw (0,0) grid (12,3);
    \fill[black!60] (5.9,-0.1) rectangle (6.1,3.1);
    \draw[very thick, black] (6,0) -- (6,3);
    \fill[blue] (2,0) rectangle (3,1);
    \node[white] at (2.5,0.5) {A1};
    \fill[red] (8,1) rectangle (9,2);
    \node[white] at (8.5,1.5) {A2};
    \node[below] at (3,0) {Agent 1 Side};
    \node[below] at (9,0) {Agent 2 Side};
\end{tikzpicture}
\caption{A depiction of SyncSght Environment (SSE). Black line in the middle is the impermissible barrier that divides the grid into two symmetric subgrids. The agent A1 is shown to have view size of $2$, while the agent A2 is shown to have view size 5. The numbers at the top of each column correspond to the mean of the normal distribution from which the reward values for the squares in that column are sampled.}
\label{fig:synsight.environment}
\end{figure}}{}
\textbf{NZSC training helps agents coordinate better}: 
In Figure~\ref{fig:nzsc}, we show cross play return for NZSC-trained agents (see equation~\ref{eq:nzsc}).
Specifically, we use noisy-other-play (NOP) for OS-NLG and I-NLG and noisy maximum entropy population-based training (NMEP) for CEE and SSE.
These agents are more robust relative to self-play agents and coordinate optimally with other agents when the noise model of the partner agent matches their own noise model.
However, notably when the two agents have different noise models, and thus there is information asymmetry, these agents under-perform in the sense that their coordination performance is only as good as if both agents had the higher noise model, and they fail to leverage the fact that one of them has access to less noisy observation of the coordination problem.

\newcommand{\figurewidth}{0.24\textwidth}
\begin{figurecustom}
\begin{subfigure}{\figurewidth}
    \includegraphics[width=\textwidth]{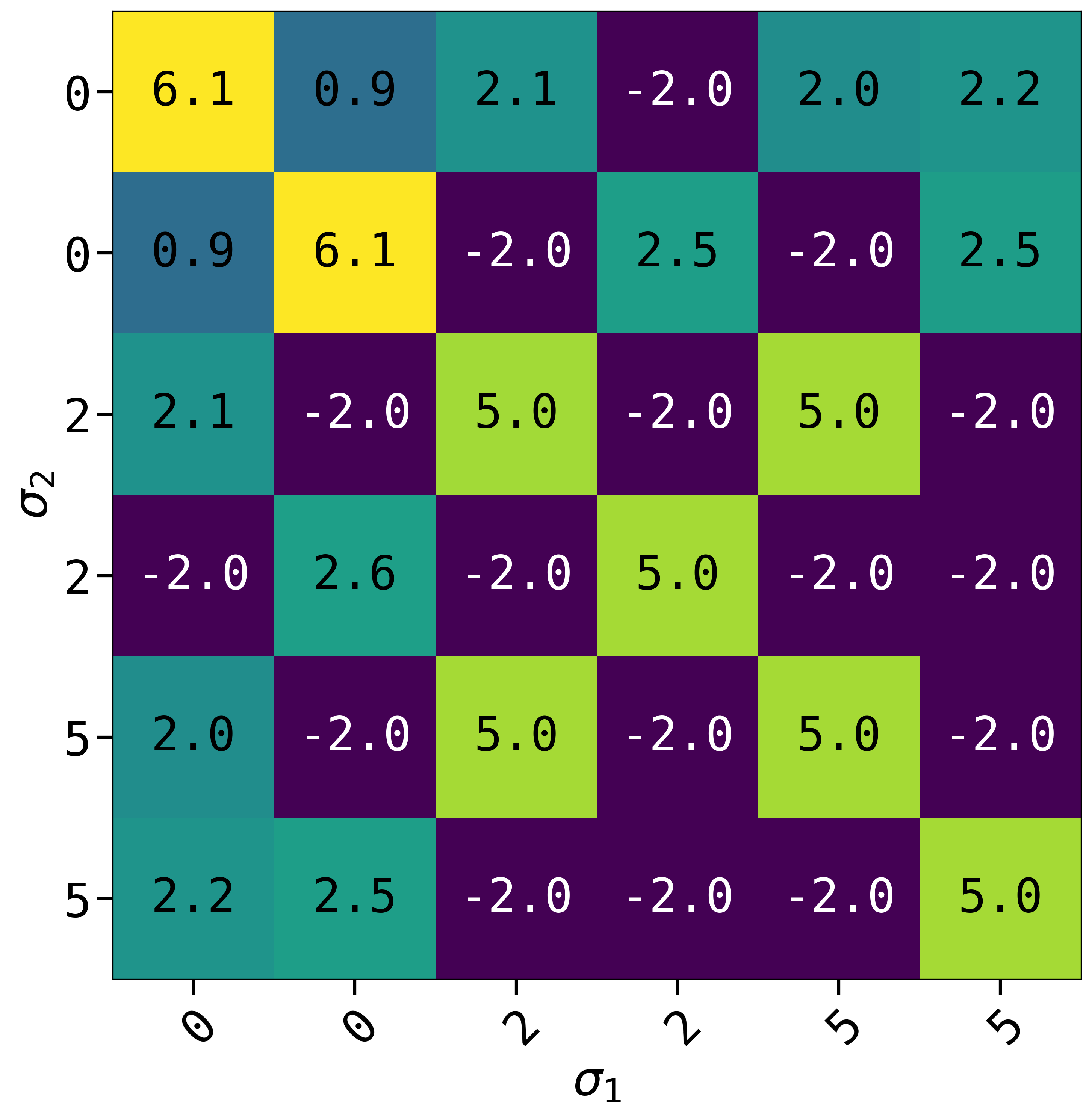}
    \caption{OS-NLG.}
    \label{fig:self-play.osnlg}
\end{subfigure}
\begin{subfigure}{\figurewidth}
    \includegraphics[width=\textwidth]{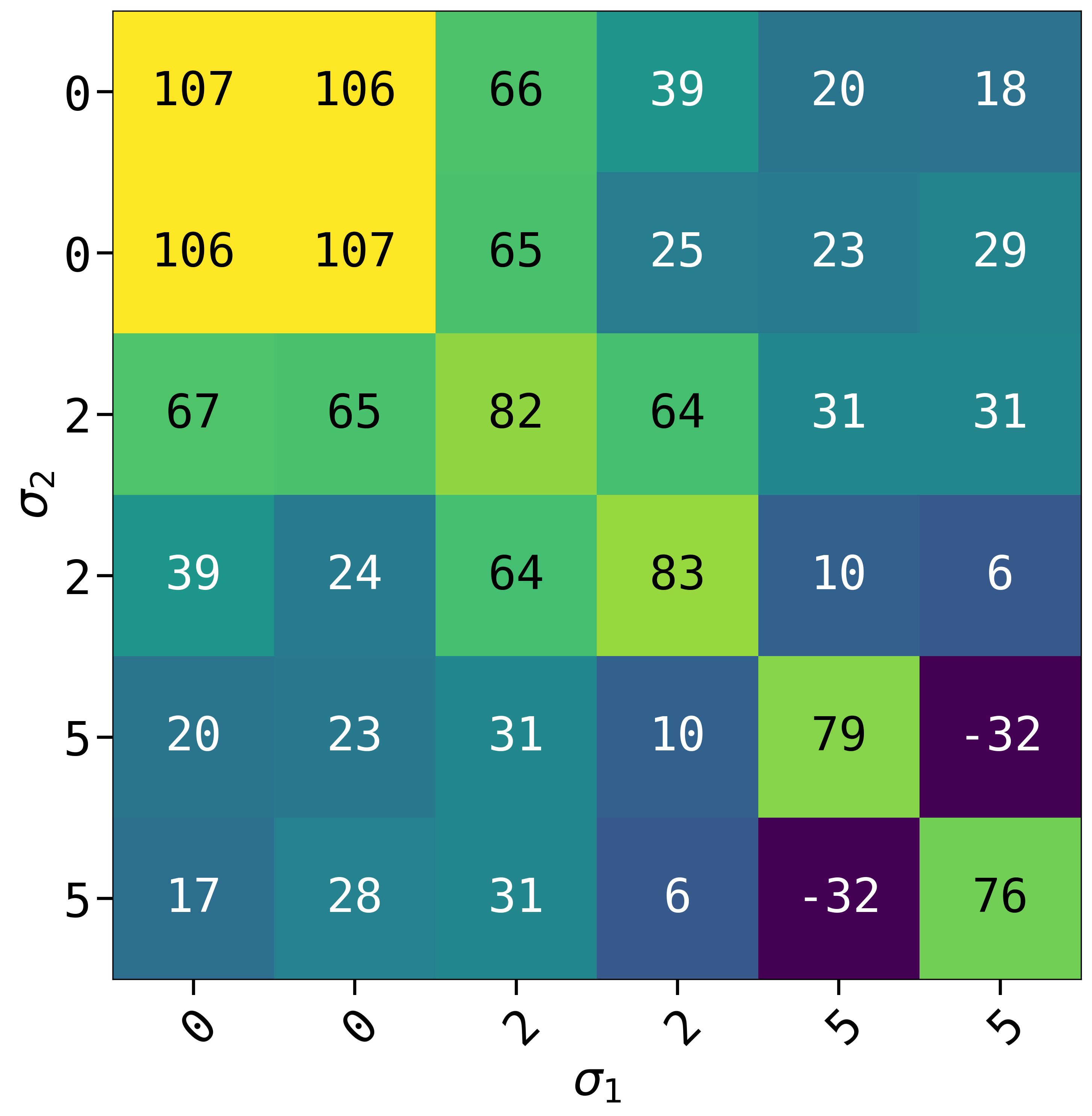}
    \caption{I-NLG.}
    \label{fig:self-play.inlg}
\end{subfigure}
\begin{subfigure}{\figurewidth}
    \includegraphics[width=\textwidth]{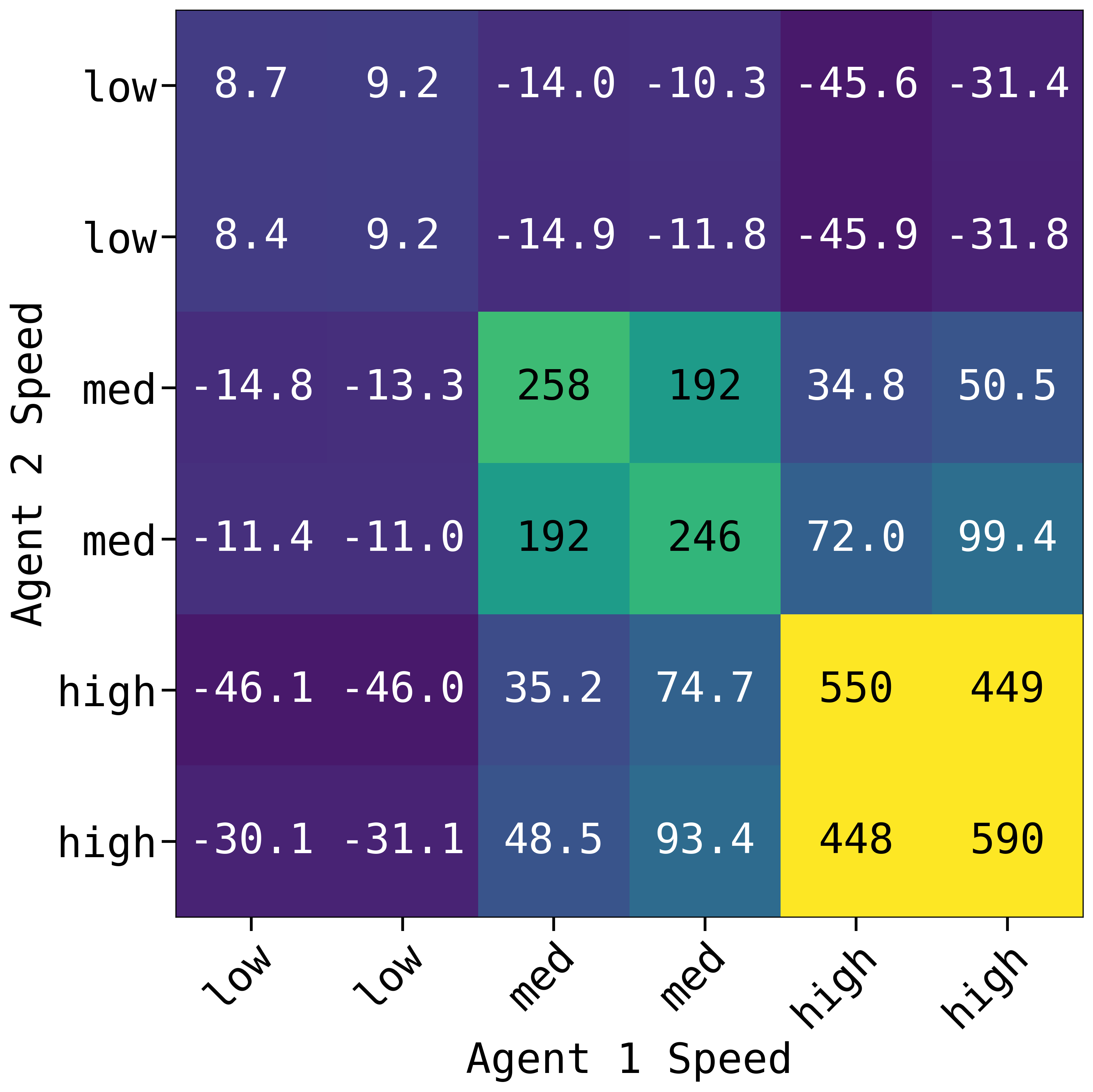}
    \caption{CEE.}
    \label{fig:self-play.cee}
\end{subfigure}
\begin{subfigure}{\figurewidth}
    \includegraphics[width=\textwidth]{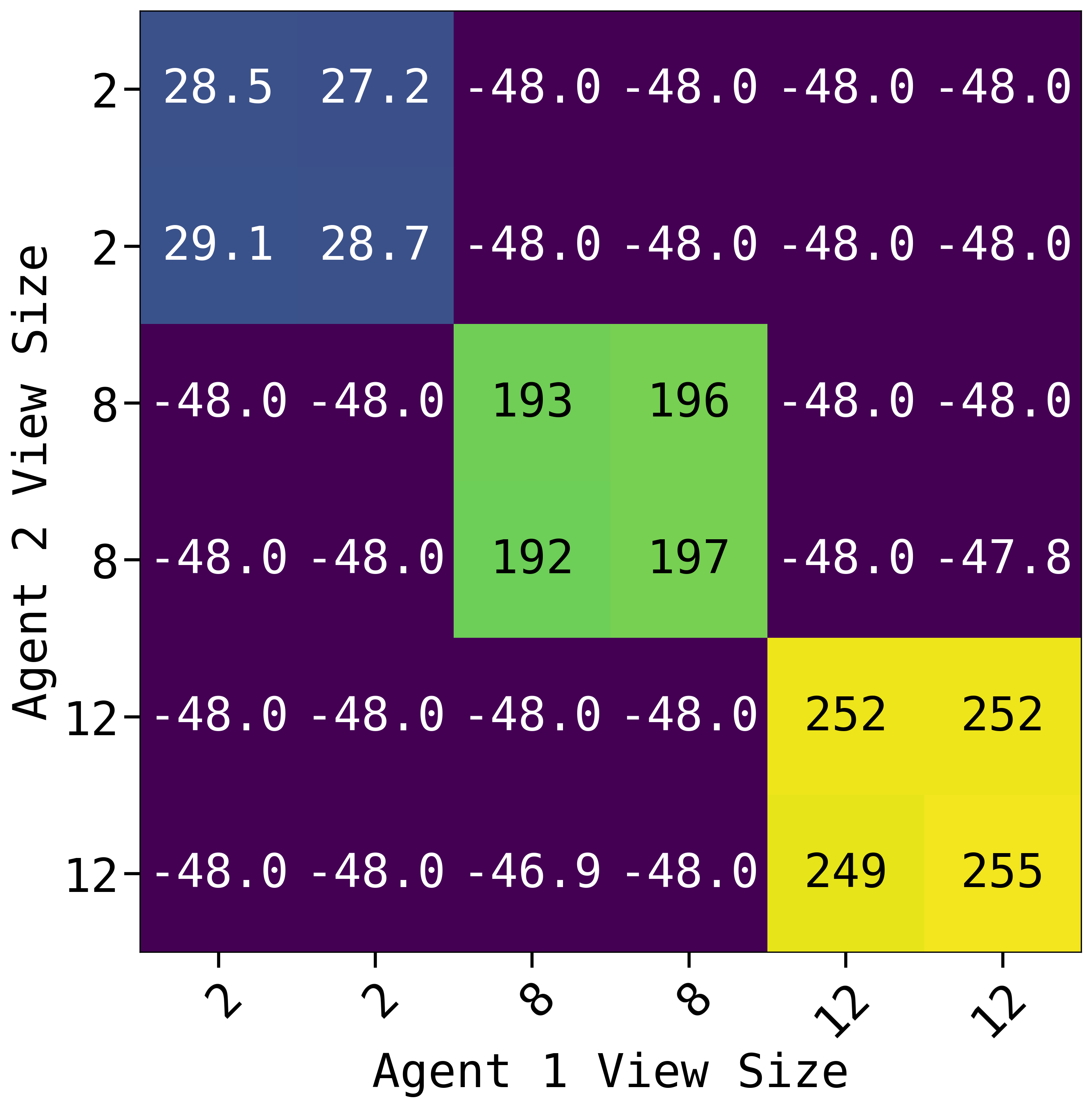}
    \caption{SSE.}
    \label{fig:self-play.see}
\end{subfigure}
\caption{Cross-play return for agents trained via Self-Play.}
\label{fig:self-play}
\end{figurecustom}
\begin{figurecustom}
\begin{subfigure}{\figurewidth}
    \includegraphics[width=\textwidth]{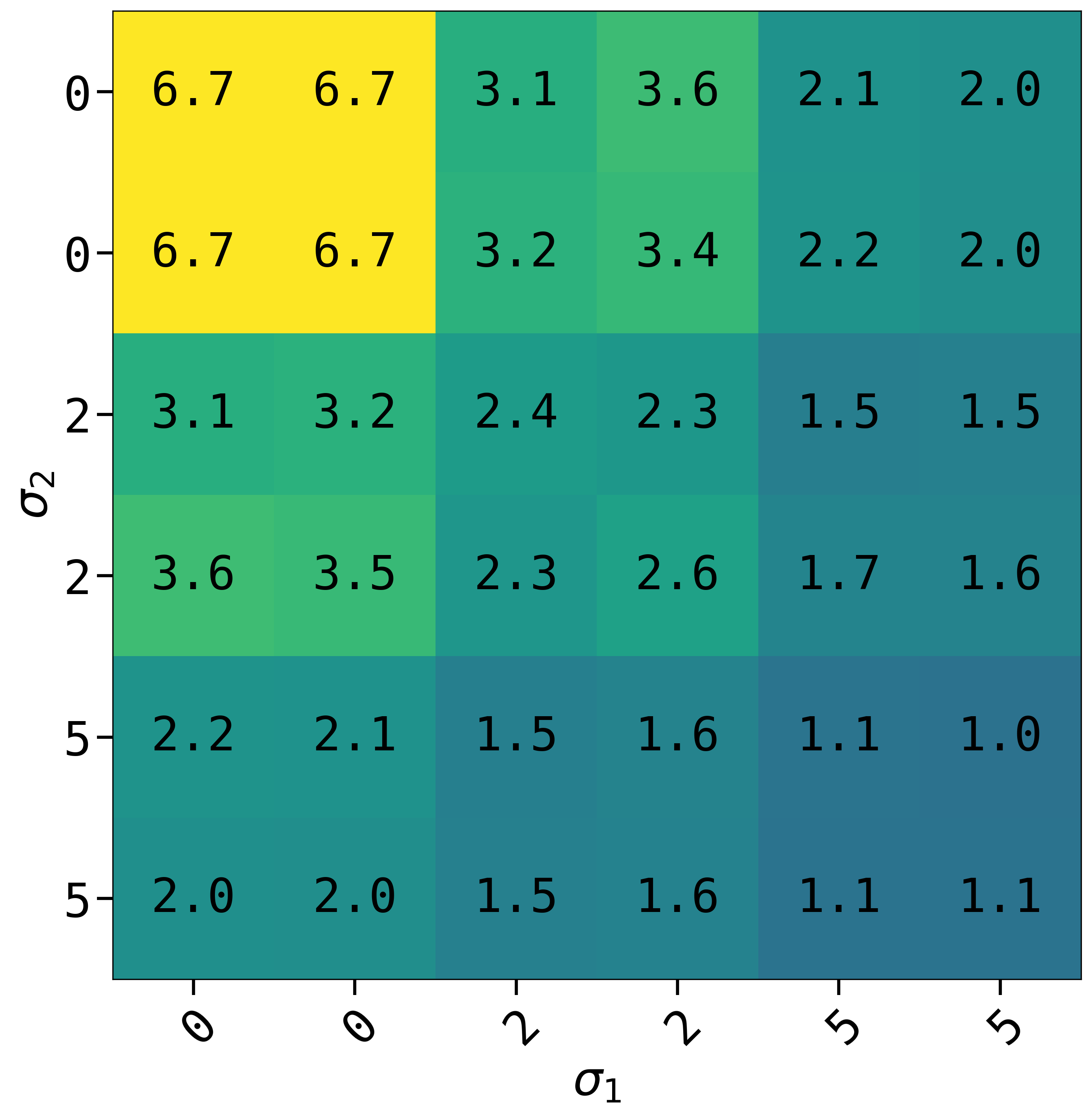}
    \caption{OS-NLG.}
    \label{fig:nzsc.osnlg}
\end{subfigure}
\begin{subfigure}{\figurewidth}
    \includegraphics[width=\textwidth]{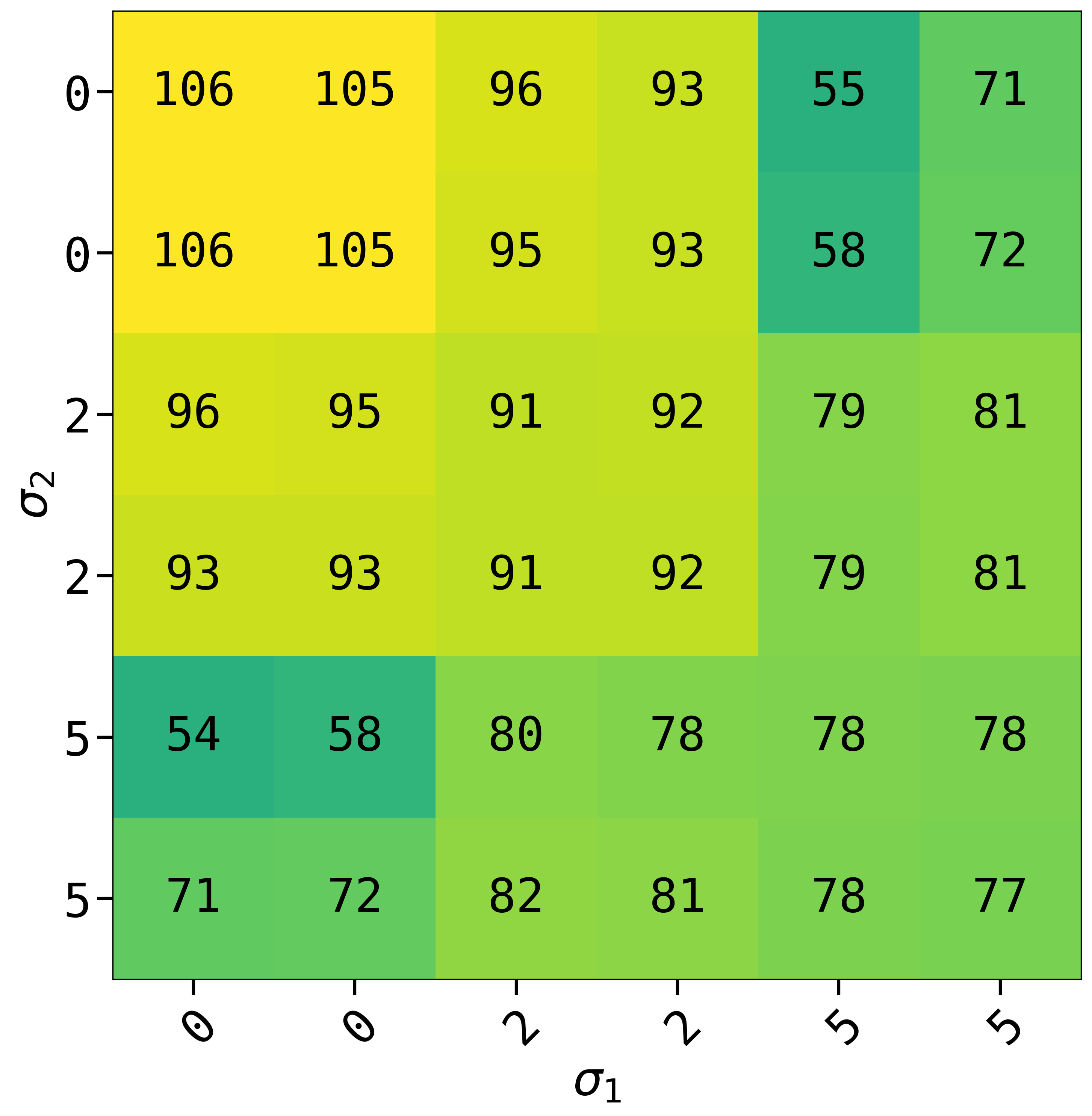}
    \caption{I-NLG.}
    \label{fig:nzsc.inlg}
\end{subfigure}
\begin{subfigure}{\figurewidth}
    \includegraphics[width=\textwidth]{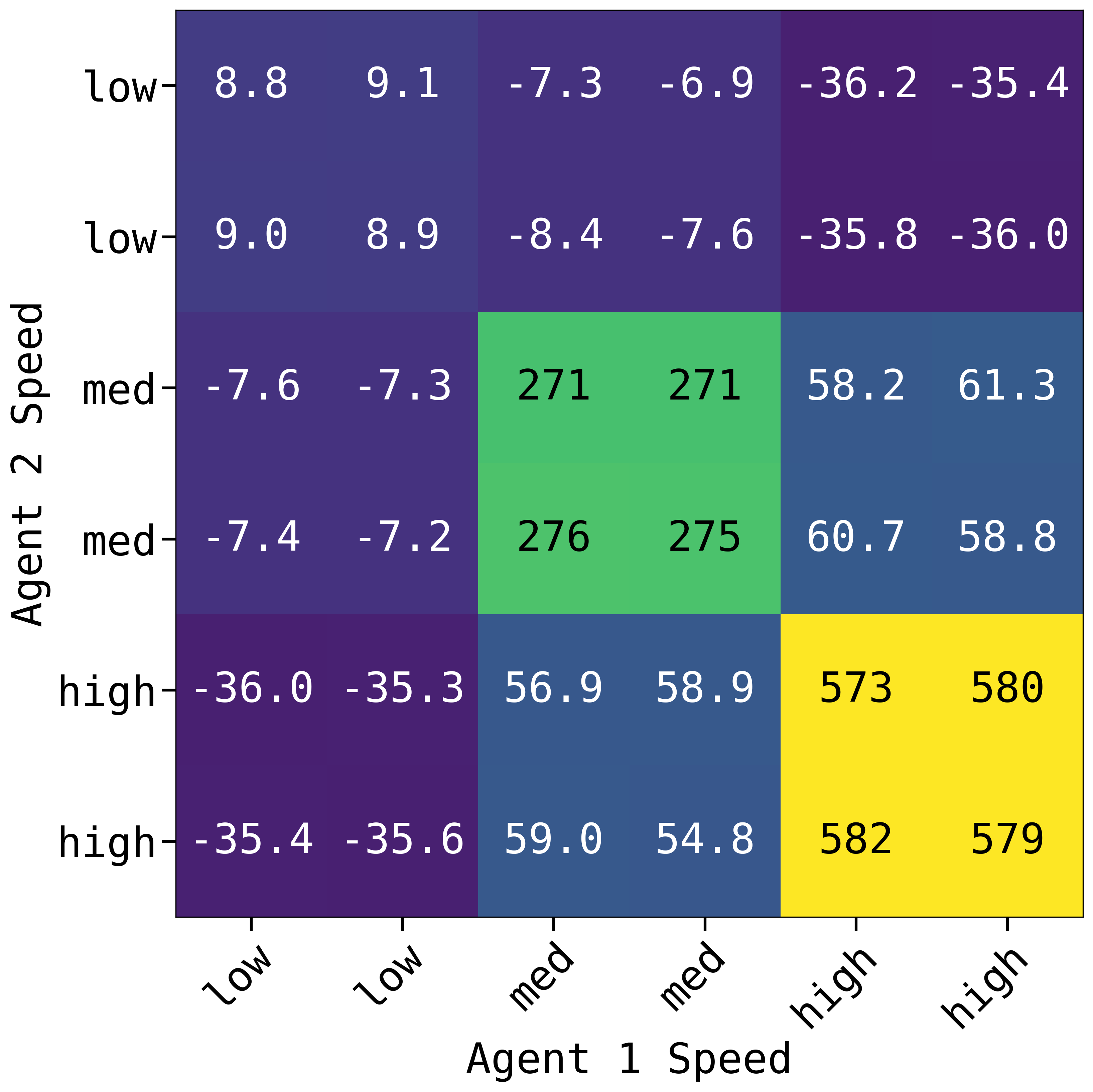}
    \caption{CEE.}
    \label{fig:nzsc.cee}
\end{subfigure}
\begin{subfigure}{\figurewidth}
    \includegraphics[width=\textwidth]{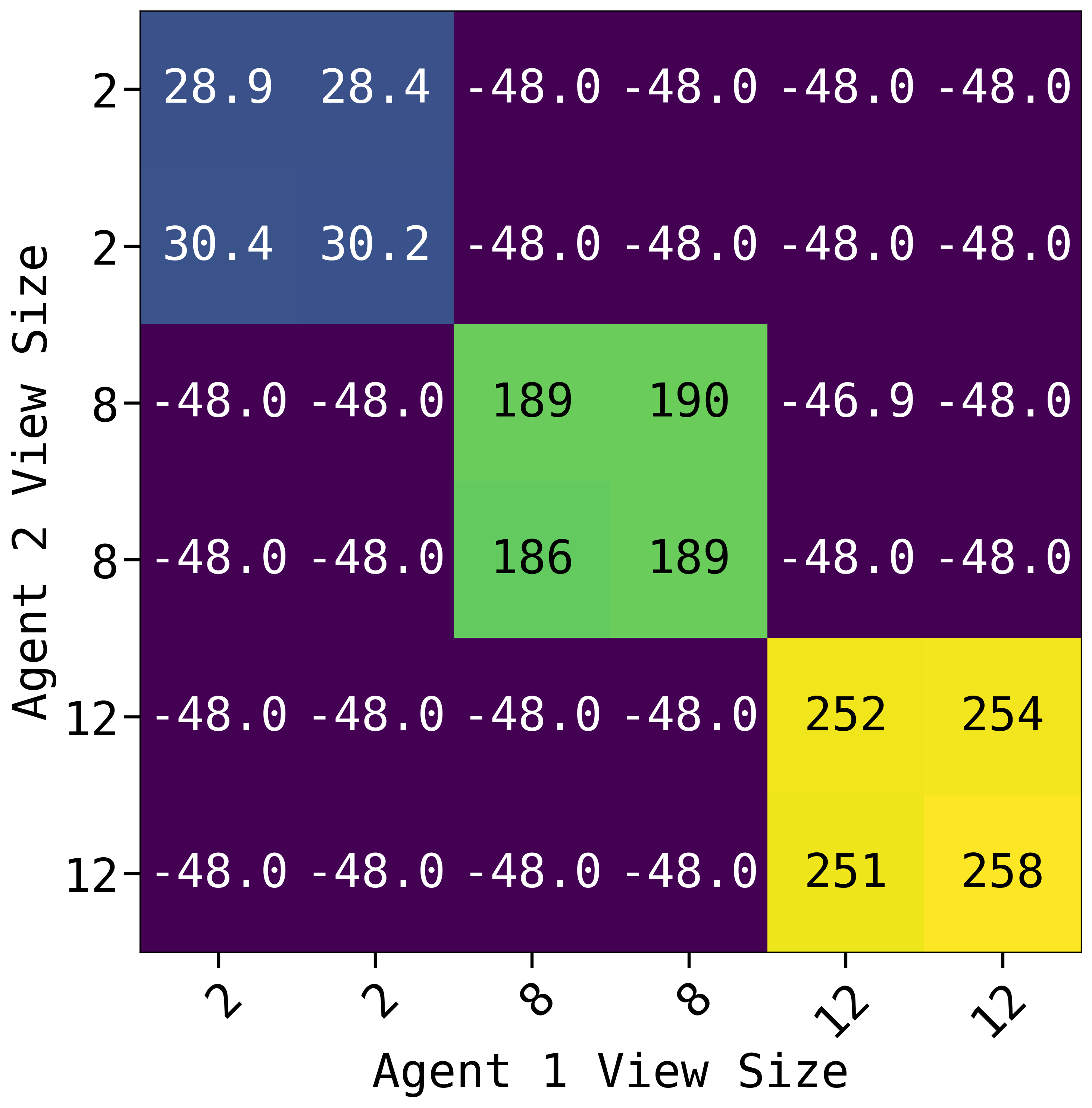}
    \caption{SSE.}
    \label{fig:nzsc.sse}
\end{subfigure}
\caption{Cross-play return for agents trained via NZSC; specifically NOP for OS-NLG and I-NLG, and NMEP for CEE and SEE. These agents are co-trained with agents whose noise model is fixed and known, hence, they coordinate well with agents with that known noise model, they do not perform well when paired with agents whose noise models may be different.}
\label{fig:nzsc}
\end{figurecustom}
\begin{figurecustom}
\begin{subfigure}{\figurewidth}
    \includegraphics[width=\textwidth]{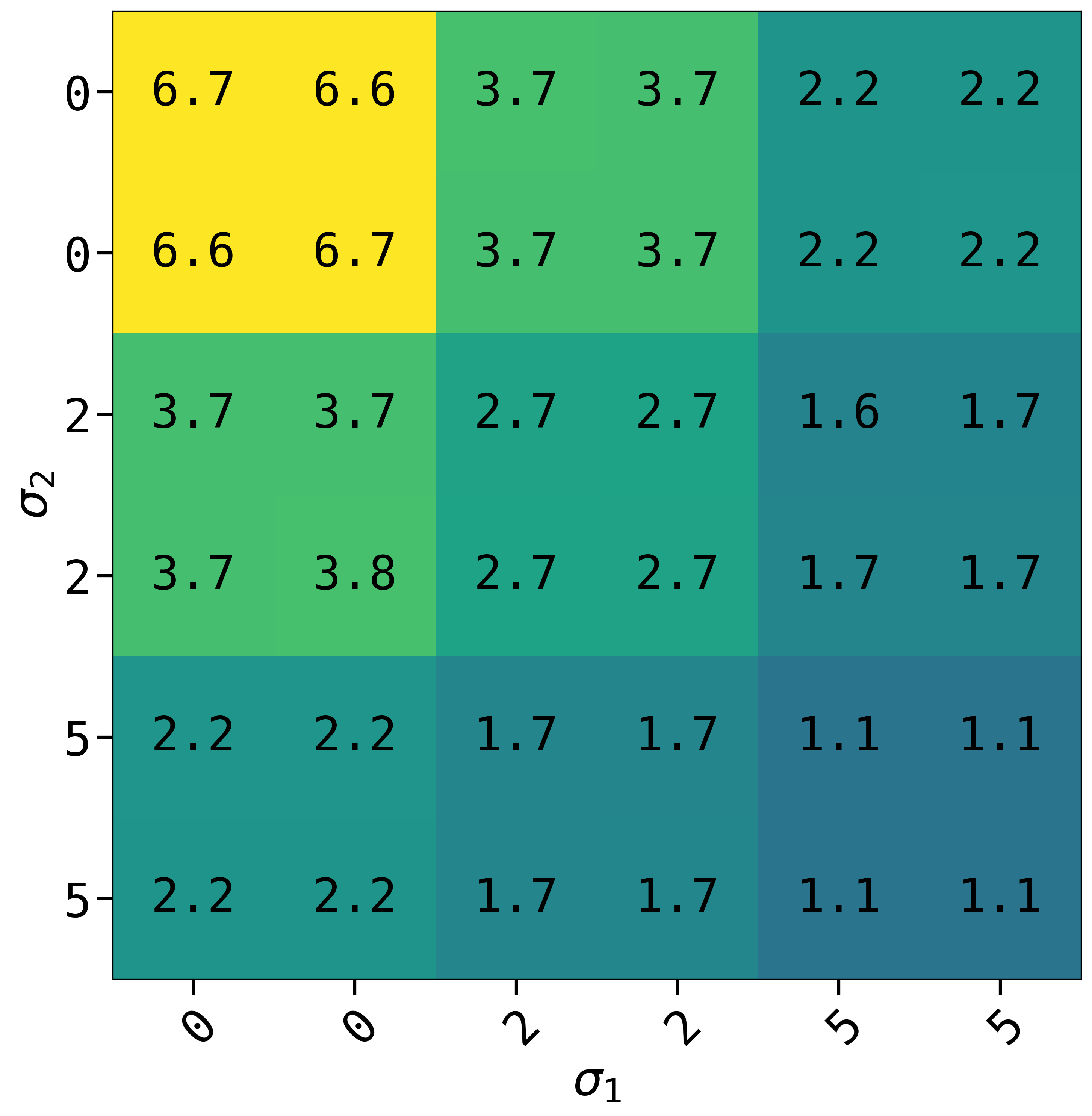}
    \caption{OS-NLG.}
    \label{fig:meta-nzsc.osnlg}
\end{subfigure}
\begin{subfigure}{\figurewidth}
    \includegraphics[width=\textwidth]{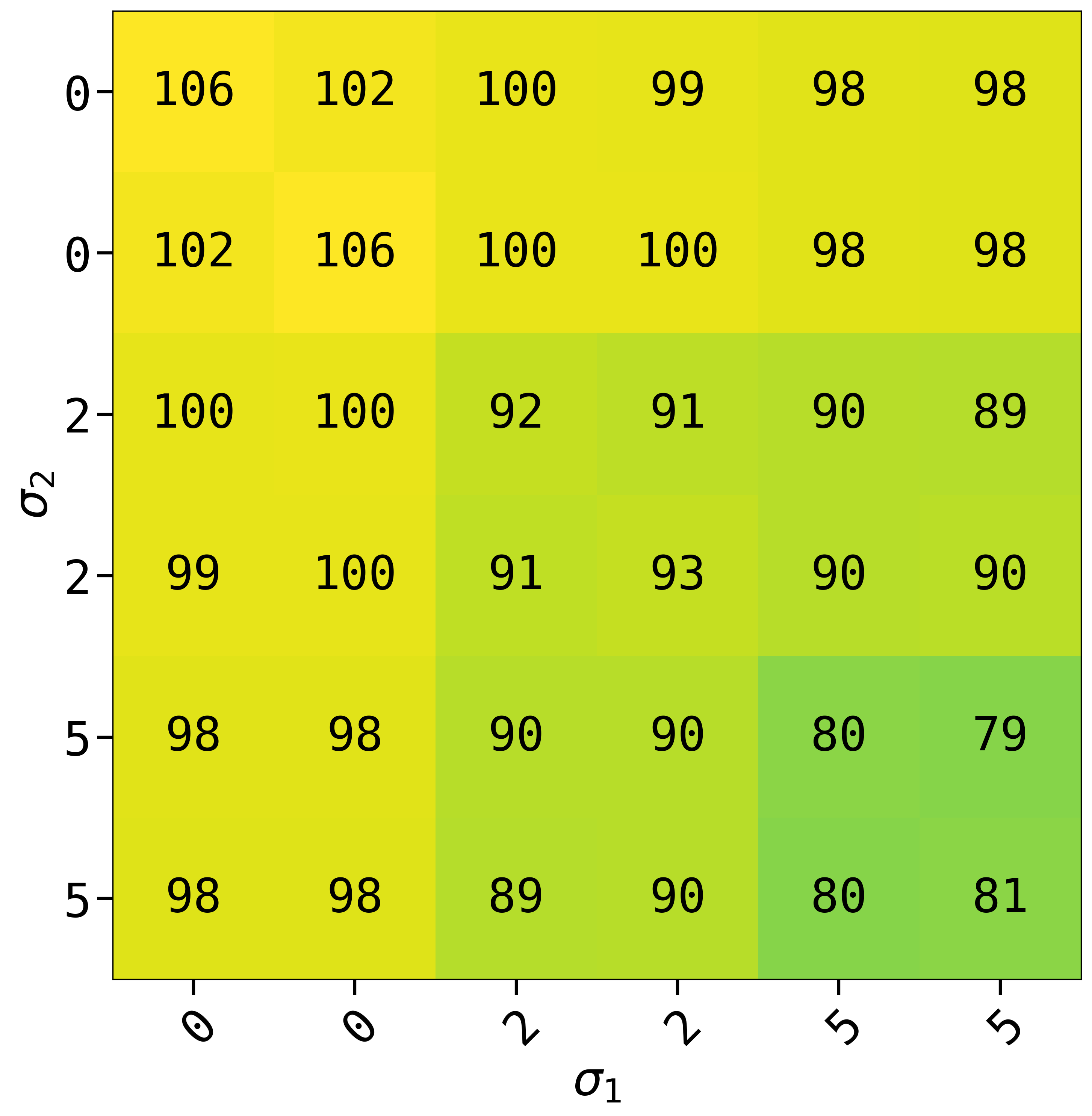}
    \caption{I-NLG.}
    \label{fig:meta-nzsc.inlg}
\end{subfigure}
\begin{subfigure}{\figurewidth}
    \includegraphics[width=\textwidth]{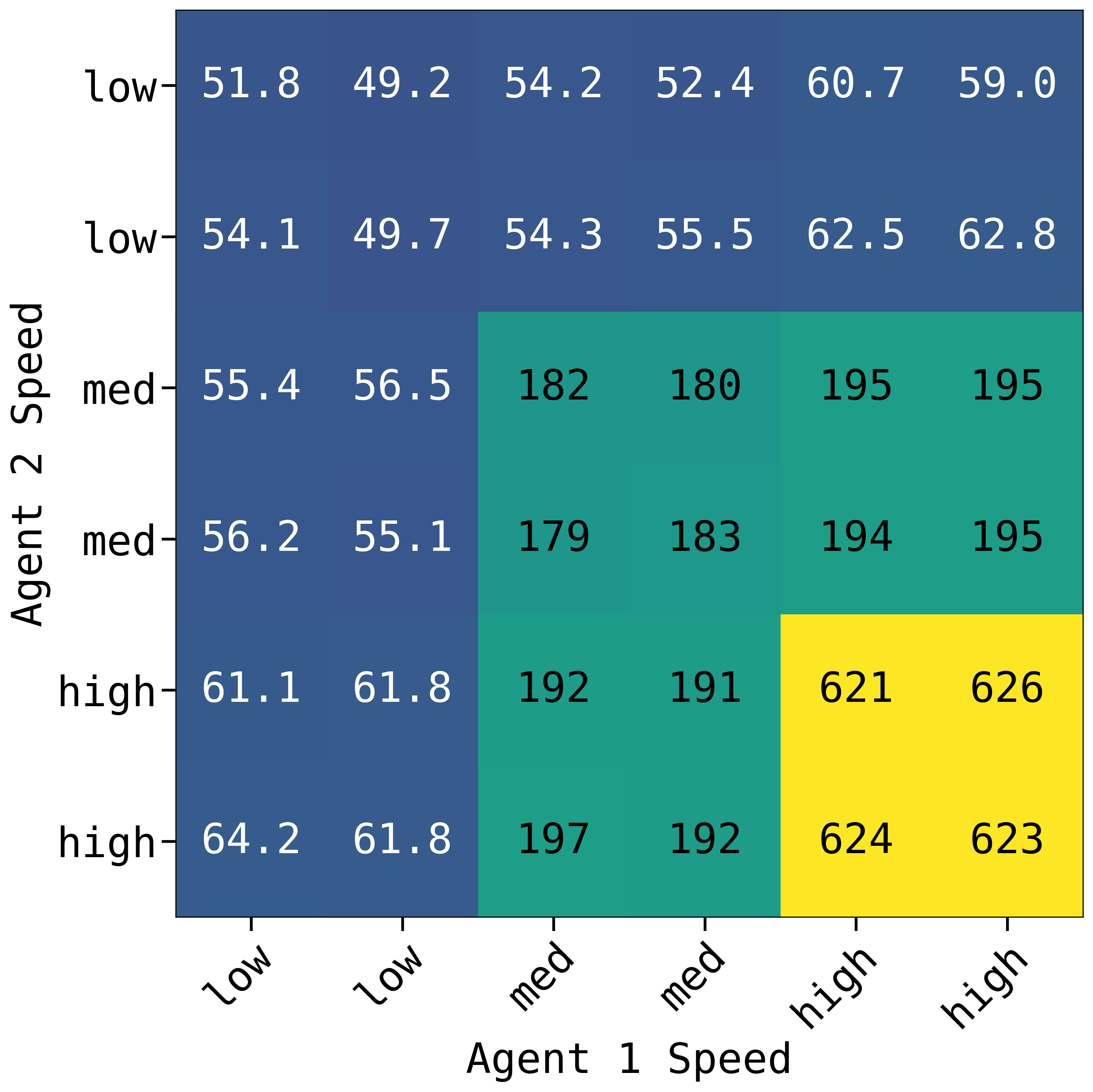}
    \caption{CEE.}
    \label{fig:meta-nzsc.cee}
\end{subfigure}
\begin{subfigure}{\figurewidth}
    \includegraphics[width=\textwidth]{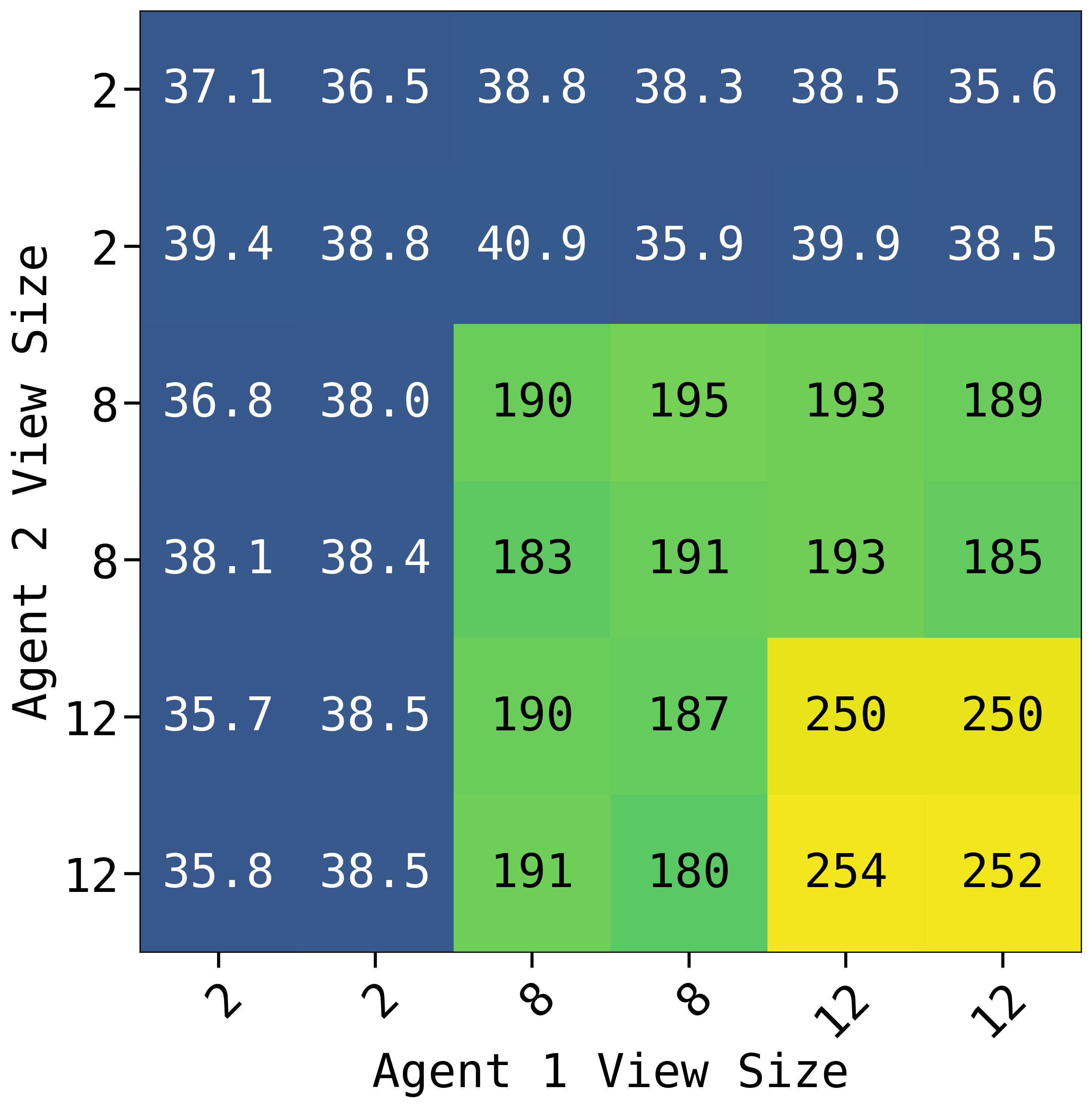}
    \caption{SSE.}
    \label{fig:meta-nzsc.sse}
\end{subfigure}
\caption{Cross-play return for agents trained via meta-NZSC; specifically meta-NOP for OS-NLG and I-NLG, and meta-MEP for CEE and SSE. Agents trained via meta-NZSC can robustly coordinate even when the noise models of the partner agent is only noisily known.}
\label{fig:meta-nzsc}
\end{figurecustom}
\textbf{Meta-NZSC trained agents coordinate optimally even when agents may have different noise models}: 
Figure~\ref{fig:meta-nzsc} shows coordination performance of meta-NZSC trained agents.
Recall that in meta-NZSC, the noisy models the agents may have are also sampled from a distribution, as such meta-NZSC should help agents learn effective coordination policies regardless of the noise models of the coordinating agents.
Indeed, meta-NZSC trained agents proficiently coordinate with each other.
Meta-NZSC trained agents perform particularly better than NZSC trained agents when there is an information asymmetry present.
For example, consider the case of two meta-NZSC trained agents in I-NLG with respective noise models corresponding to $\sigma_1=0, \sigma_2=5$.
Despite the fact that agent $2$'s observation of the coordination problem is highly noisy, the agents leverage the fact that agent $1$ can actually observe the noiseless version of the coordination problem and thus are able to coordinate at the performance level comparable to if they both had $\sigma_1=0,\sigma_2=0$.
In general, meta-NZSC trained agents are able to adapt optimally based on the characteristics of their opposing agent, which NZSC-trained agents are not able to do.
Importantly, this distinction is only obvious in environments with multiple timesteps.
In OS-NLG, the behavior of NZSC trained agents and meta-NZSC trained agents is quite similar.
This is because in a one-shot game, the agents have no way of exchanging information with each other and must rely only on their own observation of the coordination problem for deciding on their action.

Interestingly, note that for CEE environment, when coordinating with other agents with low speeds, meta-NZSC trained agents attain better returns than agents trained with NZSC or even self-play.
This is because in meta-NZSC, the agents with `low' speeds are trained with agents with `medium' and `high' speeds.
Training with these agents forces these agents to be more exploratory and they end up learning about the higher rewards that randomized mines provide, and learn to mine them when they might see them but otherwise go to the mine with the fixed location but lower reward.

\vspace{-8pt}
\subsection{Robustness Analysis}
In this section, we provide experiments that shed light on the robustness of NZSC.
We primarily use I-NLG as a testbed in this section due to it being relatively computationally efficient environment, while also being a non-trivial environment from NZSC perspective.

\renewcommand{\figurewidth}{
 \ifbool{singcol}{0.23\textwidth}{0.3\textwidth}}
\renewcommand{\figurehspace}{
 \ifbool{singcol}{}{\hspace{50pt}}
}
\begin{figure}
\centering
\begin{subfigure}{\figurewidth}
    \includegraphics[width=\textwidth]{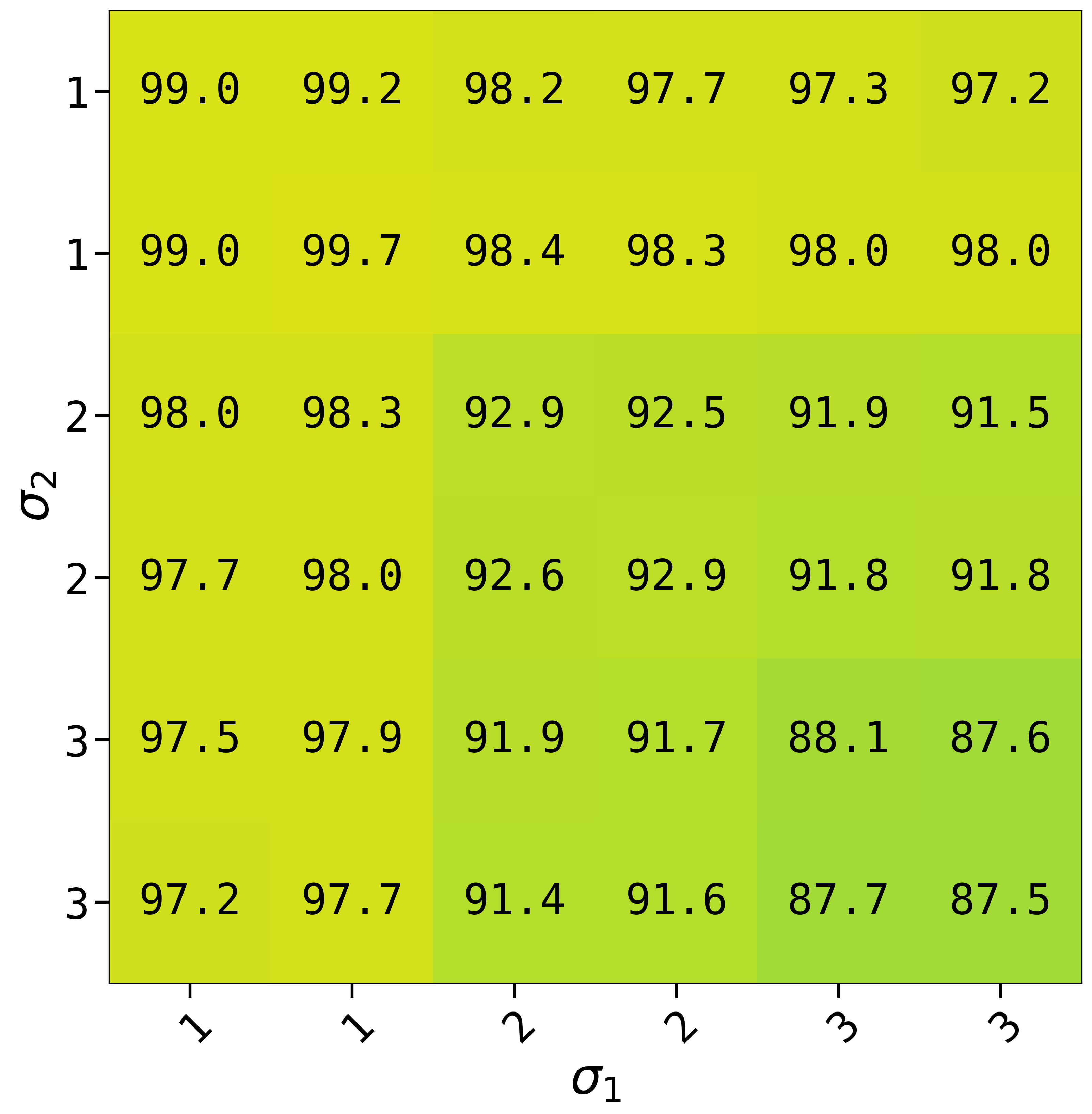}
    \caption{$z=0.5$.}
\end{subfigure}
    \figurehspace
\begin{subfigure}{\figurewidth}
    \includegraphics[width=\textwidth]{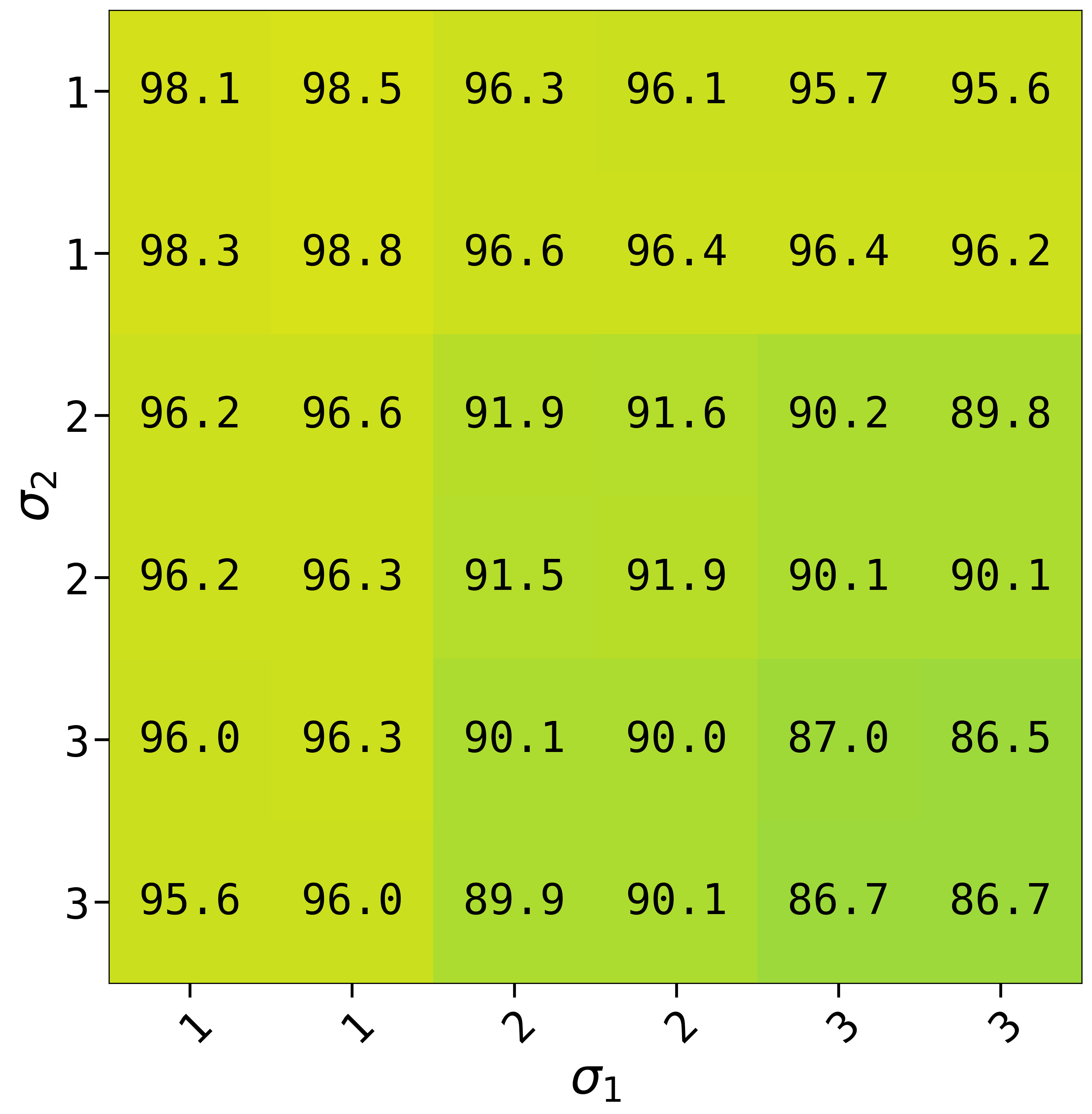}
    \caption{$z=2$.}
\end{subfigure}
\caption{Meta-NZSC trained agents can maintain their performance to a great degree even under potential misspecification of each other's noise model.}
\label{fig:robustness.uncertain.noise.model}
\end{figure}

\begin{figure}
\centering
\begin{subfigure}{\figurewidth}
    \includegraphics[width=\textwidth]{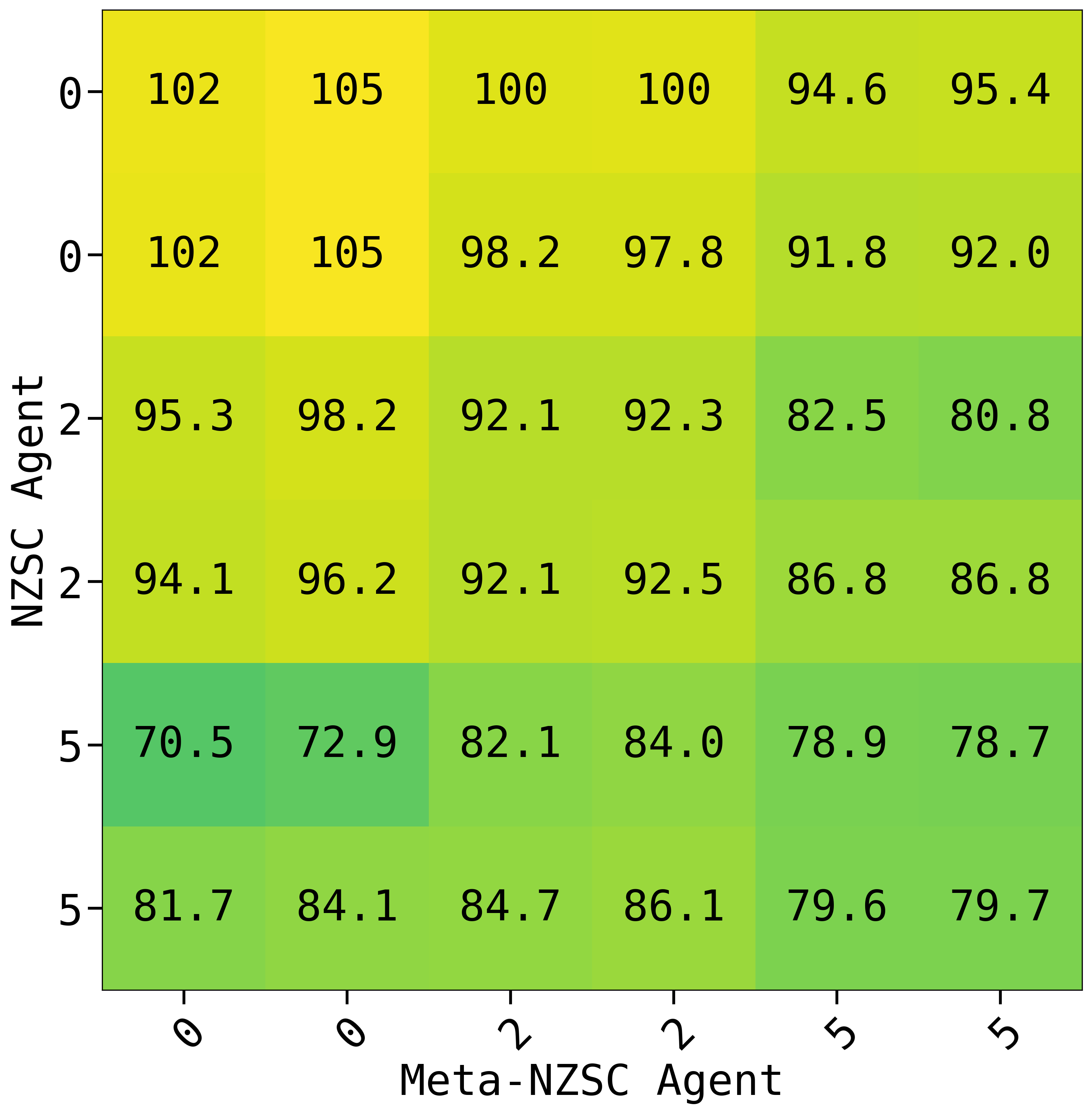}
    \caption{I-NLG.}
\end{subfigure}
    \figurehspace
\begin{subfigure}{\figurewidth}
    \includegraphics[width=\textwidth]{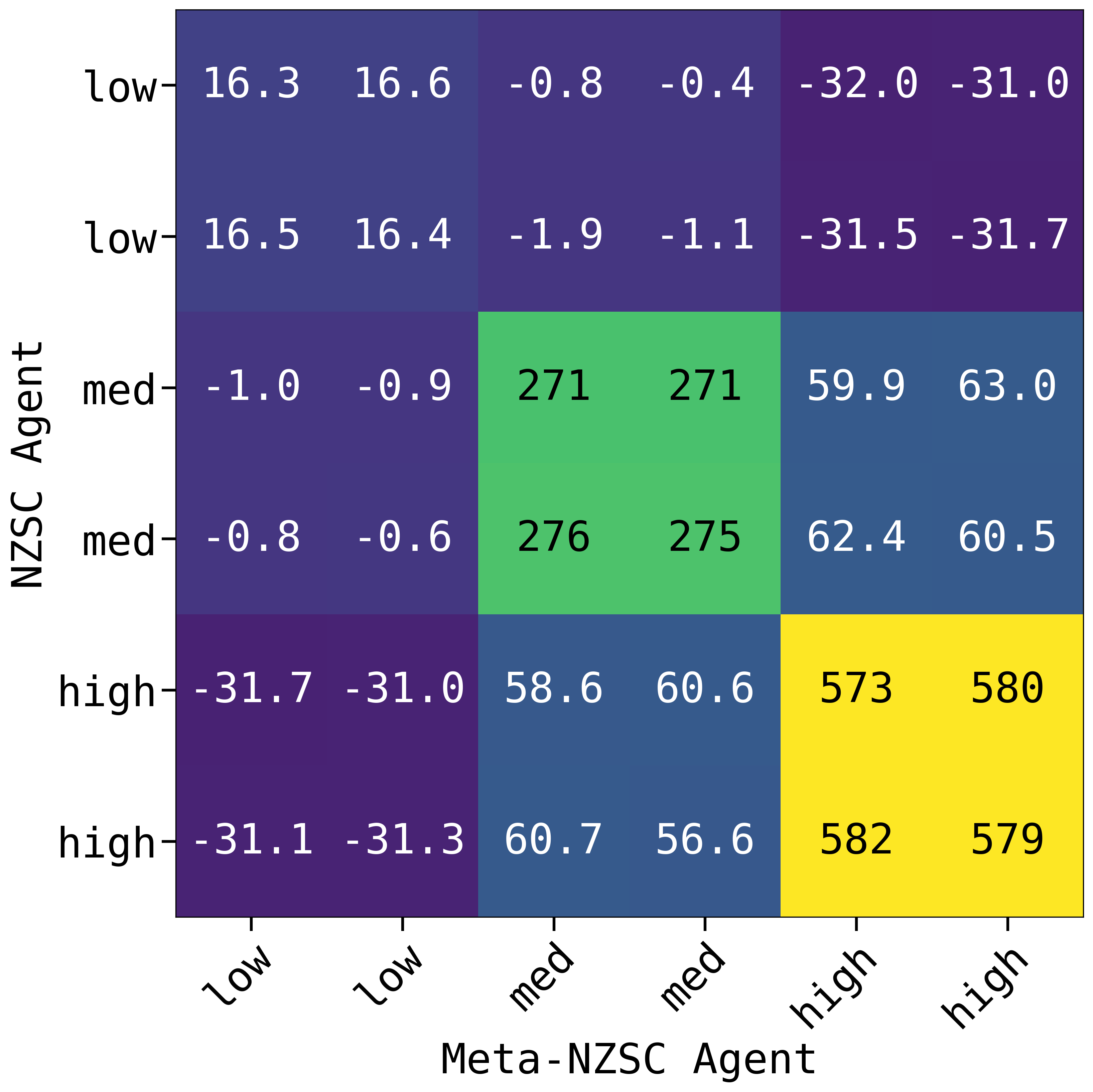}
    \caption{CEE.}
\end{subfigure}
\caption{Coordination performance of a pair of meta-NZSC trained agents and NZSC-trained agents is not markedly better than a pair of NZSC-trained agents.}
\label{fig:one-sided.meta-nzsc.training}
\end{figure}

\textbf{Robustness to misspecification of the noise model}: We have so far assumed that the noise model of both the agents is common knowledge, however, this assumption may be hard to meet in certain scenarios.
To evaluate the potential implications of violating this assumption, we consider a modified version of the I-NLG.
In the modified version of I-NLG, the agent $A_1$ still observes the true value of $\sigma_1$, but instead of observing $\sigma_2$, $A_1$ observes a random variable $X \sim N(\sigma_2, z)$ and vice versa for the agent $A_2$.
This creates uncertainty over the noise model of the partner agent.
In Figure~\ref{fig:robustness.uncertain.noise.model}, we show  results for meta-NZSC trained agents for two values of $z=0.5$ and $z=2.0$ showing that the agents are able to still coordinate well despite this additional uncertainty over each others noise models.
This indicates that the assumption that the noise models be common knowledge can be potentially relaxed and should be explored in future works.

\begin{figure}[t]
    \centering
    \begin{subfigure}{\figurewidth}
        \includegraphics[width=\textwidth]{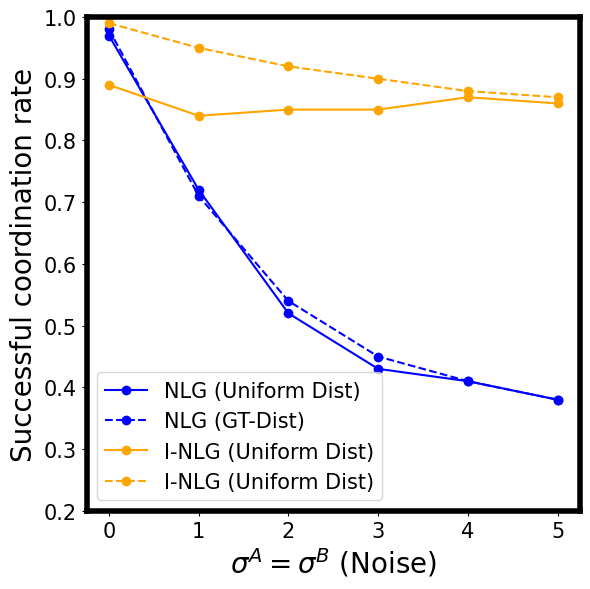}
    \end{subfigure}
    \figurehspace
    \begin{subfigure}{\figurewidth}
        \includegraphics[width=\textwidth]{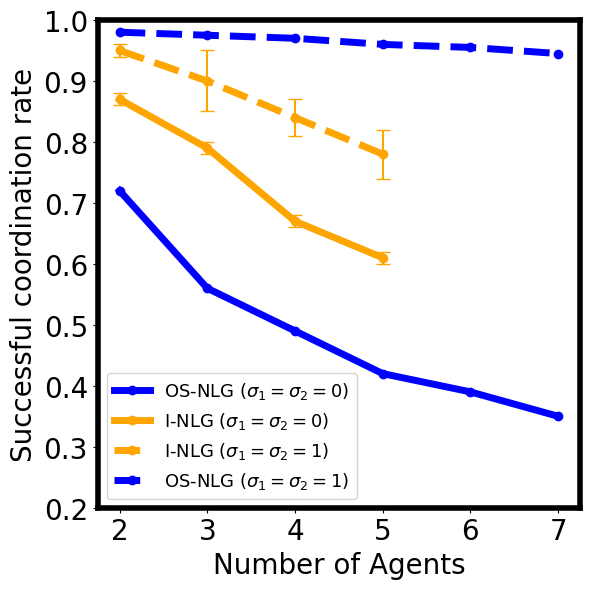}
    \end{subfigure}
    \caption{Left figure shows that successful coordination rate is similar for NZSC agents trained using uniform distribution or ground-truth distribution, indicating robustness to the misspecification of the ground truth distribution. Right figure shows the effect of the number of NZSC-trained agents involved in I-NLG on the coordination performance. For I-NLG, we could only manage to train agents in environments that had at most 5 agents.}
    \label{fig:n_agent_lg}
\end{figure}

\textbf{Robustness to Misspecification of the Ground-Truth Distribution}:
We have so far assumed that the ground-truth distribution over the coordination problems, i.e., $P(E)$ is known and common knowledge.
In practice, ground-truth distribution may not be known precisely.
If so, the principals may assume a uniform distribution (i.e., a flat prior) over the coordination problems.
In Figure~\ref{fig:n_agent_lg}, we show the coordination rate for agents in I-NLG (under ground truth distribution) trained via NZSC when assuming knowledge of ground truth distribution and when only assuming a flat prior.
We can see that in I-NLG, the NZSC training is surprisingly robust to assuming a flat prior over all coordination problems.
However, we note that in practice this may depend on how qualitatively similar flat prior is to the ground truth distribution.

\textbf{Does one-sided meta-NZSC training help?}
In practical situations, agents may differ in the \textit{extent} to which they have been trained to handle noise in the problem specification.
One such case could be where one agent has been trained via meta-NZSC and the other agent has been trained via NZSC.
In Figure~\ref{fig:one-sided.meta-nzsc.training}, we show the cross-play return between NZSC-trained agents and meta-NZSC trained agents.
Unfortunately, our results indicate that one-sided meta-NZSC training does not help improve coordination performance significantly; though in some cases it can produce minor improvements (e.g., a pair of NZSC and meta-NZSC trained, low speed and medium speed agents in CEE coordinate a bit better relative to pair of NZSC only trained agents).

\textbf{Effect of number of agents}:
Real-world coordination problems can involve more than two agents.
To better understand how the number of agents may impact ZSC performance in the presence of noise, we created a generalized version of NLG environment with $n$ agents.
Similar to the two-agent NLG environments described in Section~\ref{sec:environments}, this modified version of NLG environment requires agent to coordinate on a \textit{single} lever.
In Figure~\ref{fig:n_agent_lg}, we can see that as number of agents increases, a general degradation can be observed -- even for noiseless case -- though this degradation tends to be more noticeable for noisy case.

\section{Conclusion}
This work is primarily a critique of the commonly used setup of ZSC in which the coordination game is presumed to be common knowledge.
The unrealistic nature of this assumption is potentially a hurdle in real-world application of the ZSC research.
We propose NZSC framework which relaxes this assumption considerably, and our empirical results indicate that performant ZSC agents can be trained under this framework.
However, our investigations are quite preliminary and many avenues of future work exist.
In particular, this work highlights the need to develop more realistic benchmarks for ZSC research, so that the ZSC research is not confined to just games-like environments of Overcooked and Hanabi.


\subsubsection*{Acknowledgments}
We thank Chris Lu and Dmitrii Krasheninnikov for helpful discussions.
UA is supported by OpenPhil AI Fellowship and Vitalik Buterin Fellowship in AI Existential Safety.
JF is partially funded by the UKI grant EP/Y028481/1 (originally selected for funding by the ERC). JF is also supported by the JPMC Research Award and the Amazon Research Award.

\bibliography{references}
\bibliographystyle{iclr2025_conference}

\clearpage
\appendix

\section{Formulation}
\label{appendix:theory_stuff}
In the main text, we presented the reduction for the special case of two agent setting and all Dec-POMDPs sharing the state space. Here, we show how to handle the general cases.
\subsection{$n$-agents NZSC Reduction}
The reduction for the two-agent setting can be extended to $n$ agents straightforwardly by making following modifications: 
\begin{itemize}
\item The set of agents $\tilde D = {A_1,..,A_n}$.
\item $\mathcal{\tilde O}_{A_i} = \mathcal{O}_{A_i} \times E$ is the observation space for $i$th agent $A_i$.
\item $\tilde{\Omega}_A: \tilde{\mathcal{O}}_A \times  \tilde{S} \times A \rightarrow [0,1]$ gives the observation probability function for agent $A$ where 
    $$\tilde{\Omega}_A(\tilde o, \tilde s, a) = \tilde{\Omega}_A((o, E_i), (s,E_i),a) = [\Omega_A^{E_i}(o,s,a), E_A]$$ 
    where $E_A$ is the noisy DecPOMDP observed by agent $A$ and $\Omega_A^{E_i}$ denotes the observation function of agent $A$ in the Dec-POMDP $E_i \in \mathcal{E}$.
\end{itemize}

\subsection{Different State Spaces}
If not all Dec-POMDPs in $\mathcal{E}$ share the identical state space, then if the state space in the meta-DecPOMDP is naively set to the state space of ground truth Dec-POMDP, this may leak information about the ground truth Dec-POMDP. This is not permitted in NZSC. Thus, to prevent this the state space in the meta-dec POMDP is set to be the union of state spaces of all Dec-POMDPs in $\mathcal{E}$ i.e., $\tilde S = (\cup_{i\in I} \Ss_i)\times \E$ where $\E$ is the space of Dec-POMDPs and $I$ is an index set over $\E$


\section{Environments}
\label{appendix:environments}
\subsection{Noisy Lever Game}
\label{subsec:one_shot_nlg}

\subsubsection{Label Symmetry In The Noisy Lever Game}
Recall that state and action space in the noisy lever game consists of labels $\{1,2,3\}$ for identifying the levers. As all the levers in the noisy lever game have identical reward distribution, the \textit{labels} on any two levers can be flipped without meaningfully altering the game. For example, we can flip the labels on the first and second lever without altering the Dec-POMDP in a meaningful way.

\begin{lemma}
The following is a symmetry in the noisy lever game where $\phi_s, \phi_a, \phi_o$ are permutation maps of the state space, the action space, and the observation space respectively.
\begin{align*}
    &\phi_s((R_1,R_2,R_3)) \doteq (R_2,R_1,R_3)\\
    &\phi_a(2)\doteq 1; \phi_a(1) \doteq 2;\phi_a(3)\doteq 3\\
    &\phi_o((o_1,o_2,o_3)) \doteq (o_2,o_1,o_3) \\
\end{align*}
\end{lemma}

\clearpage
\section{Training Details}
\label{appendix:training_details}
\subsection{Training Methodology}
In OS-NLG and I-NLG, we simply use IPPO with self-play objectives and other-play objectives.
For CEE and SSE environments, we use population based training, whose details we give below.

Our training methodology comprises three progressive stages: self-play, NZSC, and Meta-NZSC. In the self-play training phase, we utilized Independent Proximal Policy Optimization (IPPO). Each agent independently learns using Proximal Policy Optimization (PPO) while sharing a common environment. IPPO adapts PPO for multi-agent settings by treating each agent's policy optimization as a distinct PPO problem. This enables independent learning while retaining the advantages of PPO's stable optimization framework. To enhance diversity among the policies, we trained 8 different seeds for each environment.

For the NZSC training stage, we built upon the population of 8 self-play trained agents, focusing on training additional agents to coordinate effectively with this varied set of policies. This approach facilitated the development of robust coordination strategies that can handle different partner behaviors.

In the Meta-NZSC phase, we expanded the training population by incorporating all successful NZSC agents. This resulted in training populations ranging from 9 to 18 agents, depending on the specific environment, thereby presenting an even broader array of coordination challenges.

To address the partial observability inherent in our POMDP settings, we employed an LSTM-based network architecture. In the CEE and SSE environments, our architecture directly embeds a 2-channel local grid view and incorporates global reward mean information into the observation vector for the LSTM. 

\subsection{Experiment Hyperparameters}
We utilized the following hyperparameters, which were tailored to each environment due to their distinct reward structures. The alpha parameter was set to 0.01 and the beta parameter to 3, in accordance with the recommendations from MEP.

\begin{table}[h]
\centering
\caption{Training Hyperparameters for Each Environment}
\label{tab:hyperparameters}
\resizebox{\textwidth}{!}{%
\begin{tabular}{|l|c|c|c|c|}
\hline
\textbf{Parameter} & \textbf{One-Shot NLG} & \textbf{Iterated NLG} & \textbf{CEE} & \textbf{SyncSight} \\
\hline
Learning Rate & $5 \times 10^{-4}$ & $5 \times 10^{-4}$ & $5 \times 10^{-4}$ & $5 \times 10^{-4}$ \\
\hline
Number of Environments & 1024 & 1024 & 1024 & 2048 \\
\hline
Steps per Environment & 2 & 32 & 256 & 128 \\
\hline
Total Timesteps & $2 \times 10^7$ & $2 \times 10^7$ & $3 \times 10^8$ & $5 \times 10^8$ \\
\hline
PPO Update Epochs & 4 & 8 & 16 & 8 \\
\hline
Number of Minibatches & 4 & 16 & 8 & 32 \\
\hline
Discount Factor ($\gamma$) & 0.99 & 0.99 & 0.99 & 0.99 \\
\hline
GAE Lambda & 0.95 & 0.95 & 0.95 & 0.95 \\
\hline
PPO Clip Range & 0.2 & 0.3 & 0.2 & 0.1 \\
\hline
Entropy Coefficient & 0.01 & 0.01 & 0.05 & 0.03 \\
\hline
FC Layer Dimension & 16 & 16 & 128 & 768 \\
\hline
LSTM Hidden Dimension & 16 & 16 & 128 & 768 \\
\hline
Environment Time Horizon & 1 & 16 & 32 & 16 \\
\hline
Prioritization Beta & - & - & 3 & 3 \\
\hline
MEP Entropy Coefficient ($\alpha$) & - & - & 0.01 & 0.01 \\
\hline
\end{tabular}%
}
\end{table}

\clearpage
\section{Additional Results}
\label{appx:sec.additional_results}

\renewcommand{\figurewidth}{0.35\textwidth}
\newcommand{\addspace}{
\ifbool{singcol}{}{\protect\\[\baselineskip]}
}

\subsection{One-Shot Noisy Lever Game}
In the following figure, we show cross-play return and final step coordination rate for different types of agents trained in one-shot noisy lever game.
Note that there is very minimal difference in NZSC training results, and meta-NZSC training results, indicating that agents require multiple timesteps to be able to exchange information between them and influence each other's policy.

\ifbool{singcol}{}{\vspace{5em}}
\begin{figure}[H]
\centering
\begin{subfigure}{\figurewidth}
\includegraphics[width=\textwidth]{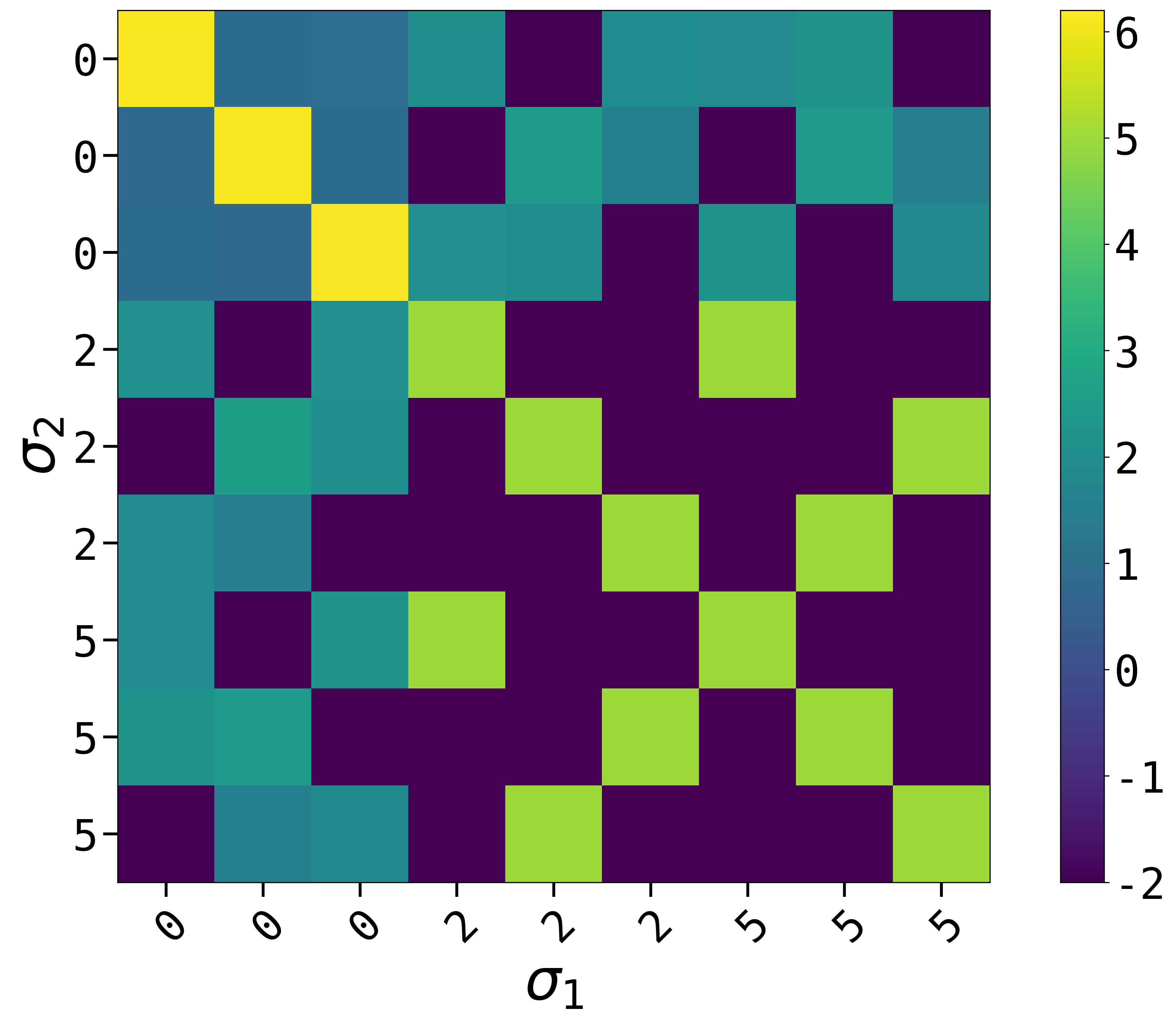}
\caption{Self Play: Cross-Play Return\addspace}
\label{fig:ape_self_play_rew}
\end{subfigure}
\hspace{25pt}
\begin{subfigure}{\figurewidth}
\includegraphics[width=\textwidth]{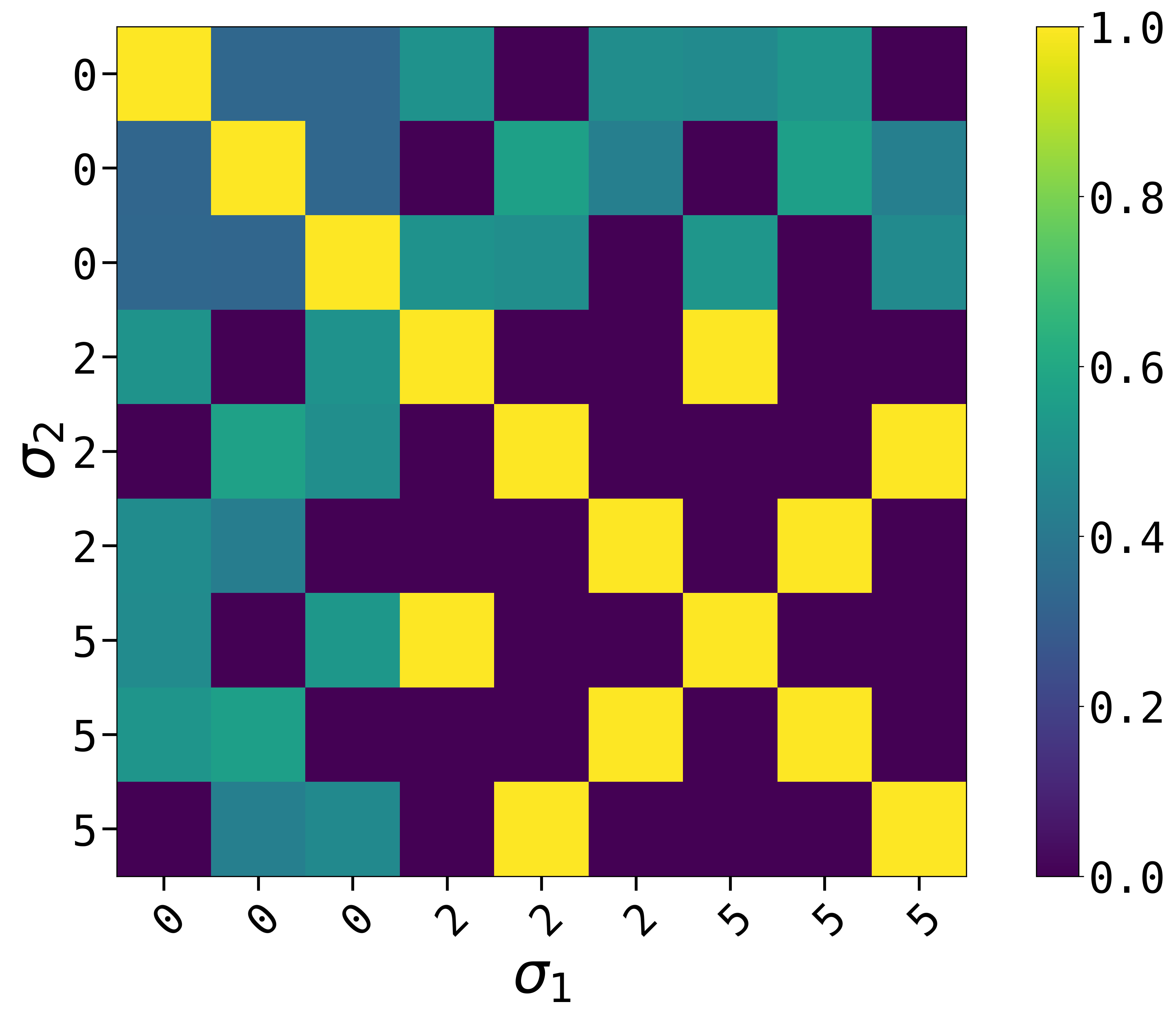}
\caption{Self Play: Terminal Step Coordination Rate}
\label{fig:ape_randomization_rew2}
\end{subfigure}
\begin{subfigure}{\figurewidth}
\includegraphics[width=\textwidth]{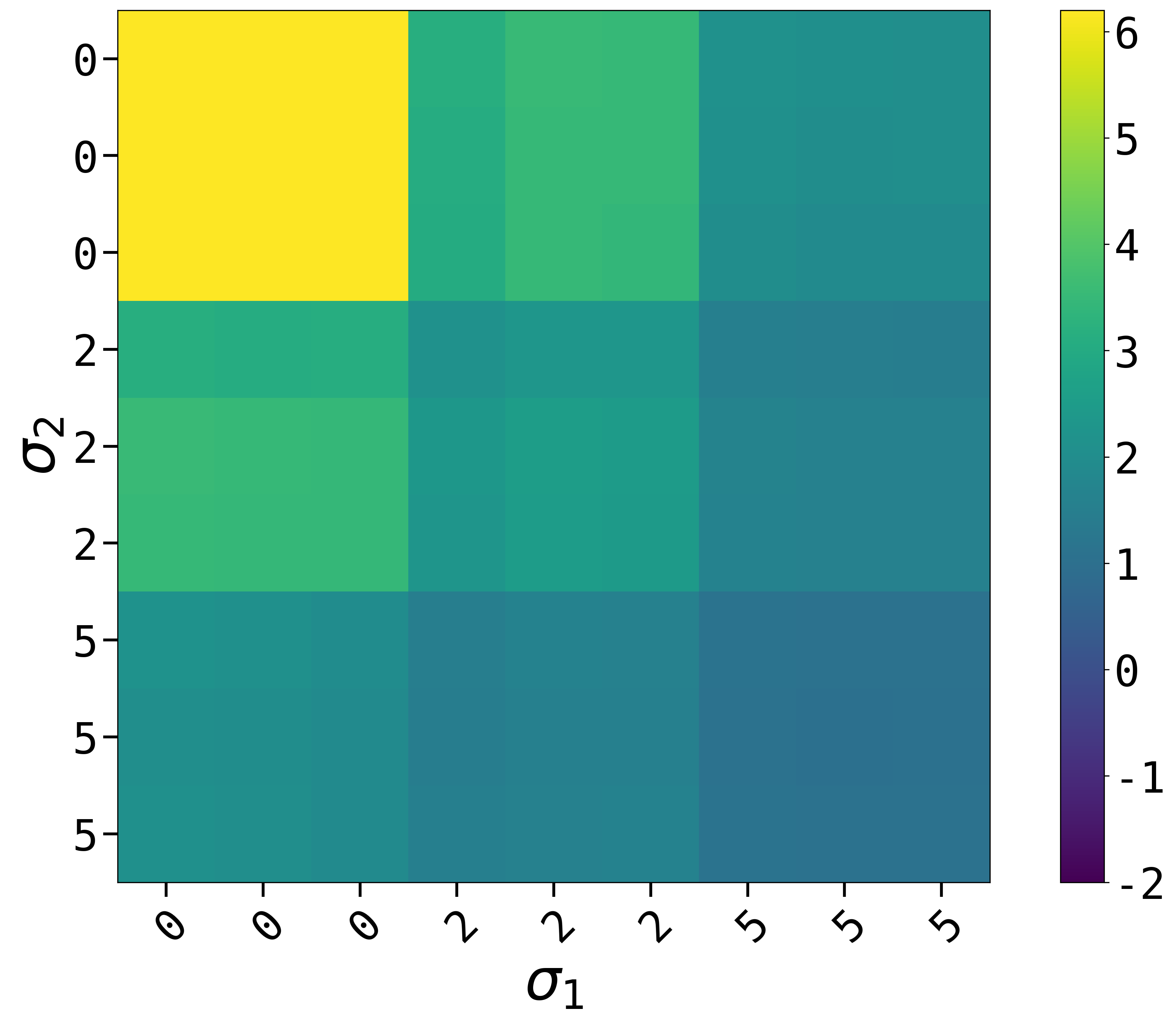}
\caption{NZSC: Cross-Play Return\addspace}
\label{fig:ape_self_play_rew_op}
\end{subfigure}
\hspace{25pt}
\begin{subfigure}{\figurewidth}
\includegraphics[width=\textwidth]{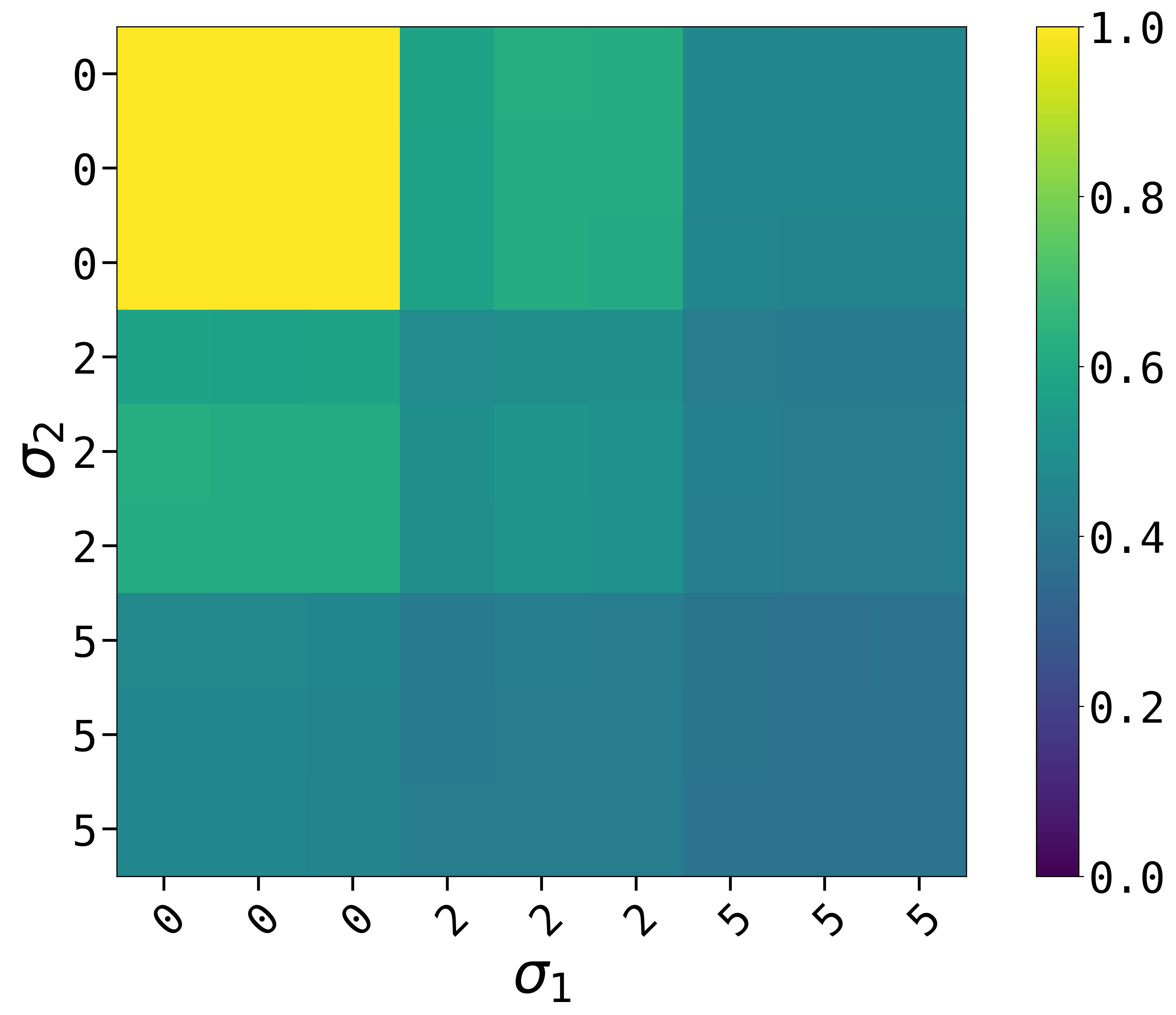}
\caption{NZSC: Terminal Step Coordination Rate}
\label{fig:ape_randomization_rew2_op}
\end{subfigure}
\begin{subfigure}{\figurewidth}
\includegraphics[width=\textwidth]{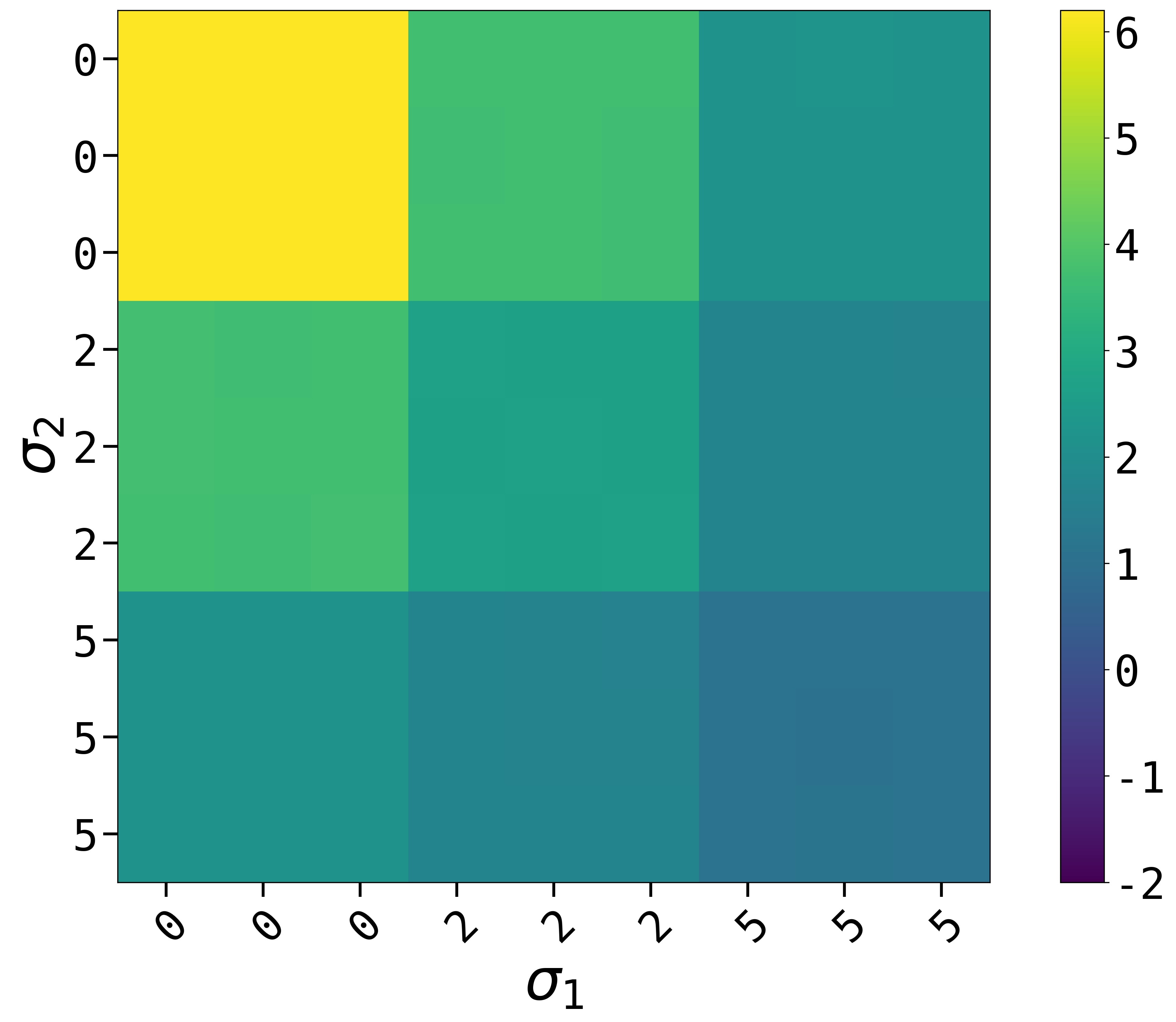}
\caption{Meta-NZSC: Cross-Play Return\addspace}
\label{fig:ape_self_play_rew_rand}
\end{subfigure}
\hspace{25pt}
\begin{subfigure}{\figurewidth}
\includegraphics[width=\textwidth]{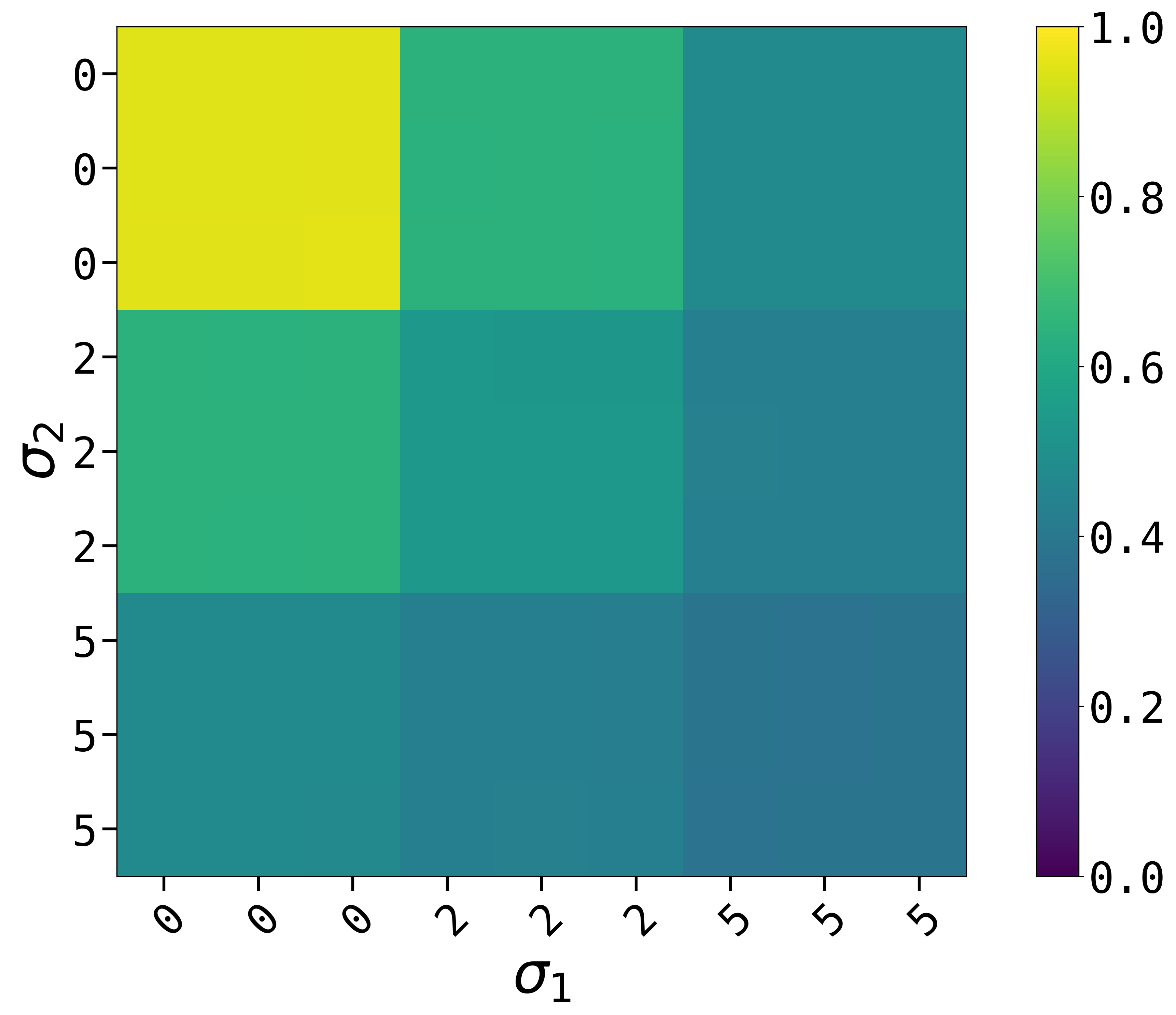}
\caption{Meta-NZSC: Terminal Step Coordination Rate}
\label{fig:ape_randomization_rew2_rand}
\end{subfigure}
\caption{One-shot noisy lever game results.}
\end{figure}

\clearpage
\subsection{Iterated Noisy Lever Game}
In the following figure, we show cross-play return and final step coordination rate for different types of agents trained in iterated noisy lever game.
Notably, some pairings of NZSC trained agents (e.g., $\sigma_1=5$, $\sigma_2=0$) can fail at coordinating by even the last timestep,
but all pairings of meta-NZSC trained agents are successful at coordinating by the final timestep almost all the time.

\ifbool{singcol}{}{\vspace{5em}}
\begin{figure}[H]
\centering
\begin{subfigure}{\figurewidth}
\includegraphics[width=\textwidth]{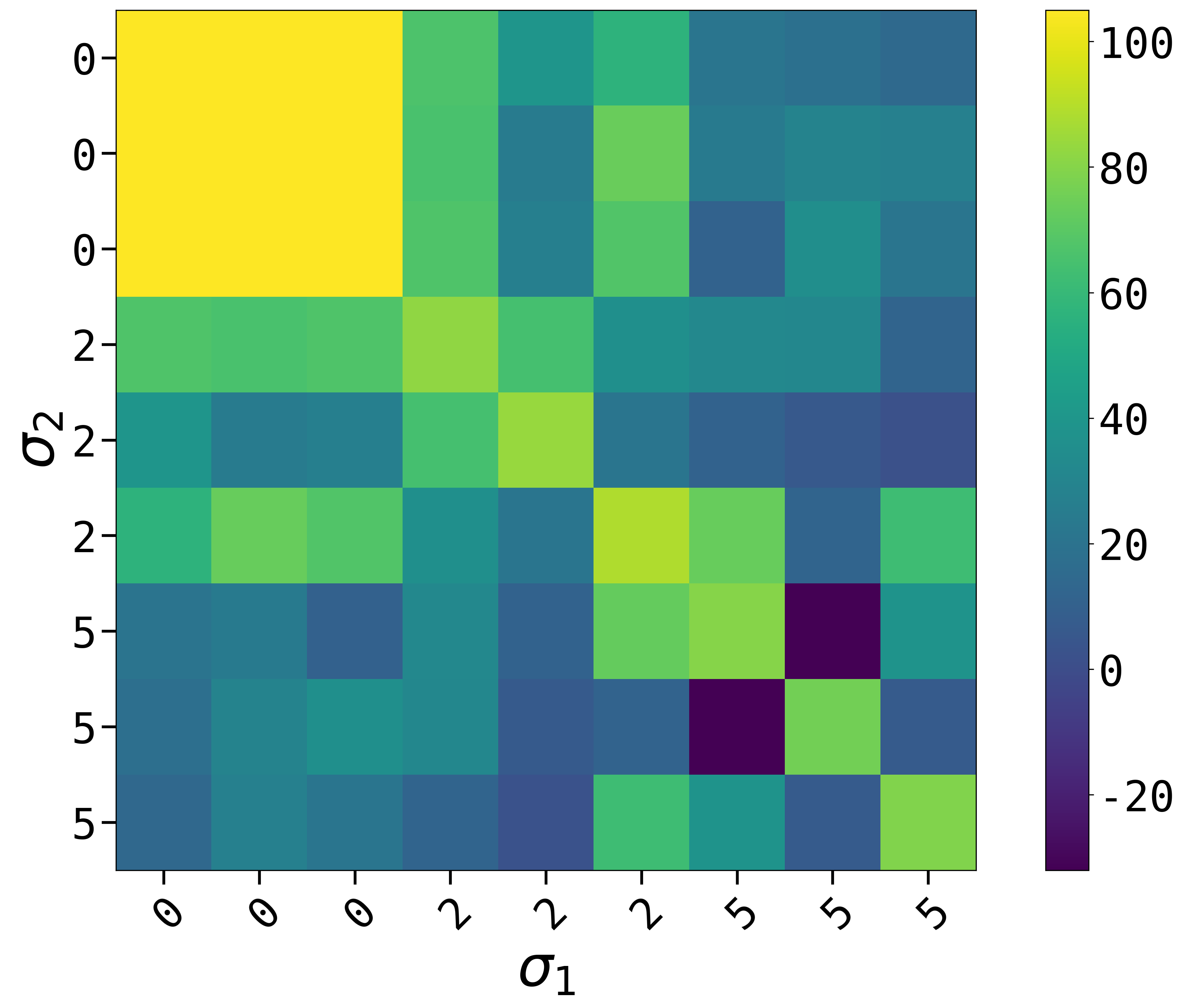}
\caption{Self Play: Cross-Play Return\addspace}
\label{fig:ape_self_play_rew}
\end{subfigure}
\hspace{25pt}
\begin{subfigure}{\figurewidth}
\includegraphics[width=\textwidth]{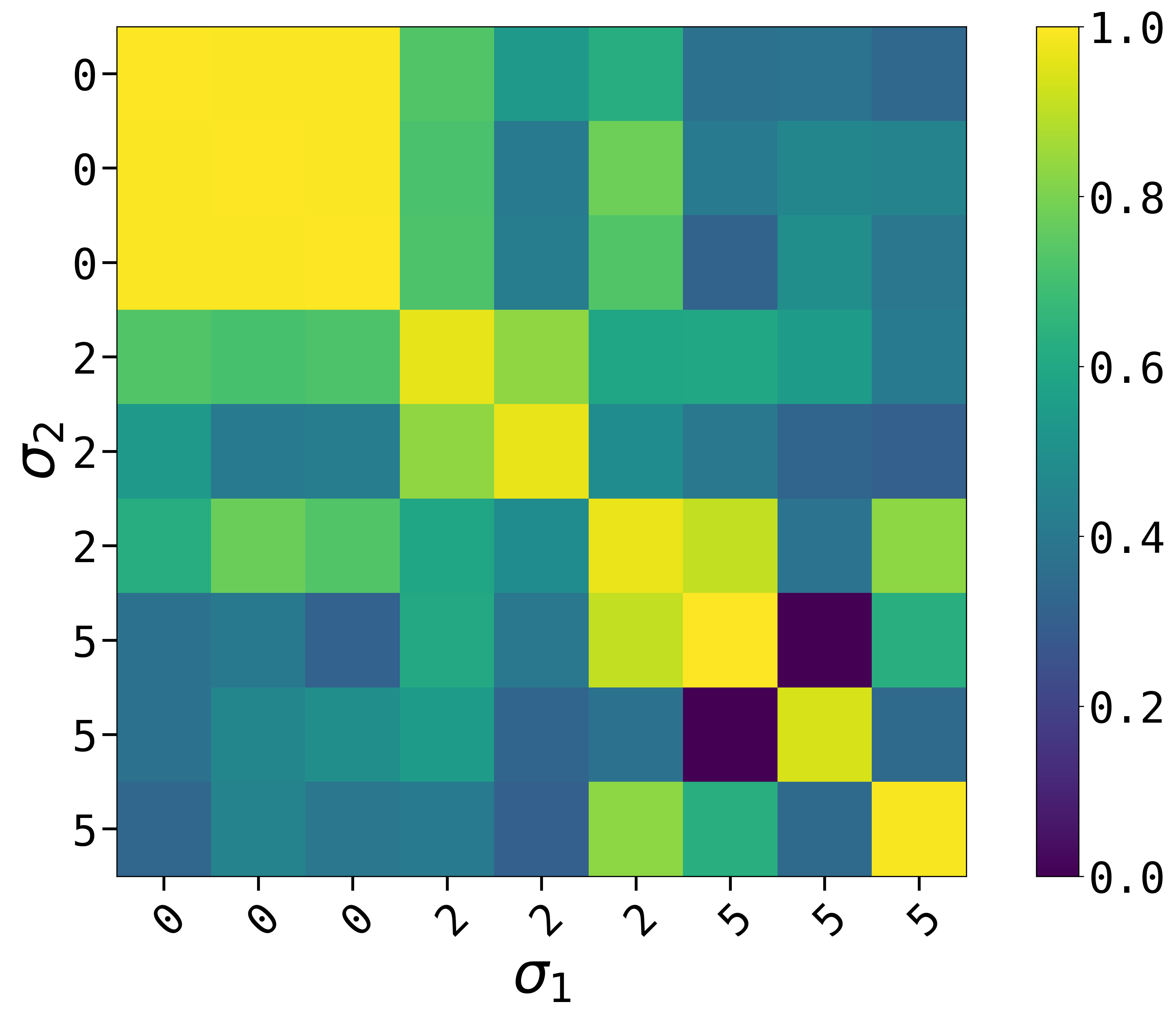}
\caption{Self Play: Terminal Step Coordination Rate}
\label{fig:ape_randomization_rew2}
\end{subfigure}
\begin{subfigure}{\figurewidth}
\includegraphics[width=\textwidth]{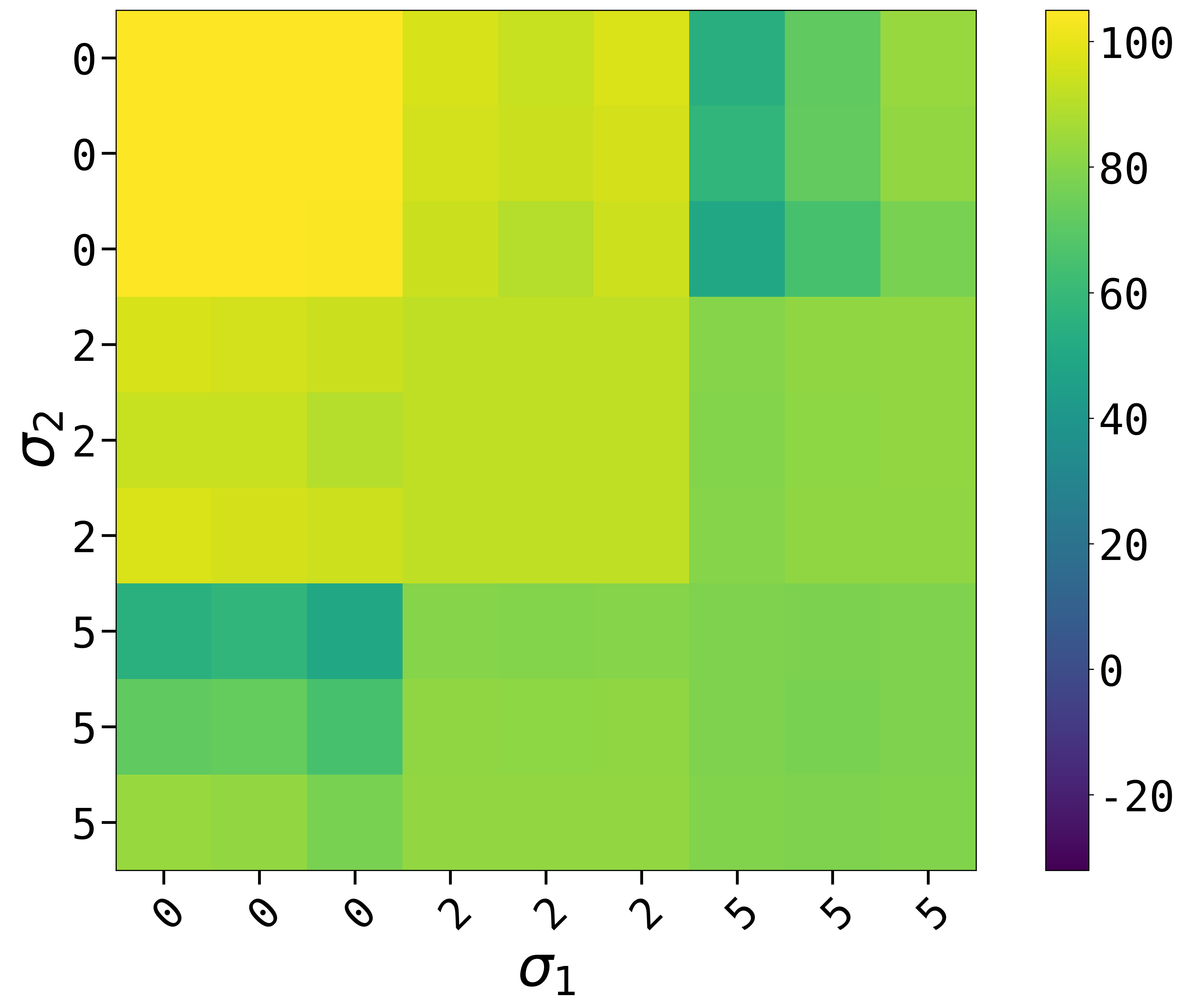}
\caption{NZSC: Cross-Play Return\addspace}
\label{fig:ape_self_play_rew_op}
\end{subfigure}
\hspace{25pt}
\begin{subfigure}{\figurewidth}
\includegraphics[width=\textwidth]{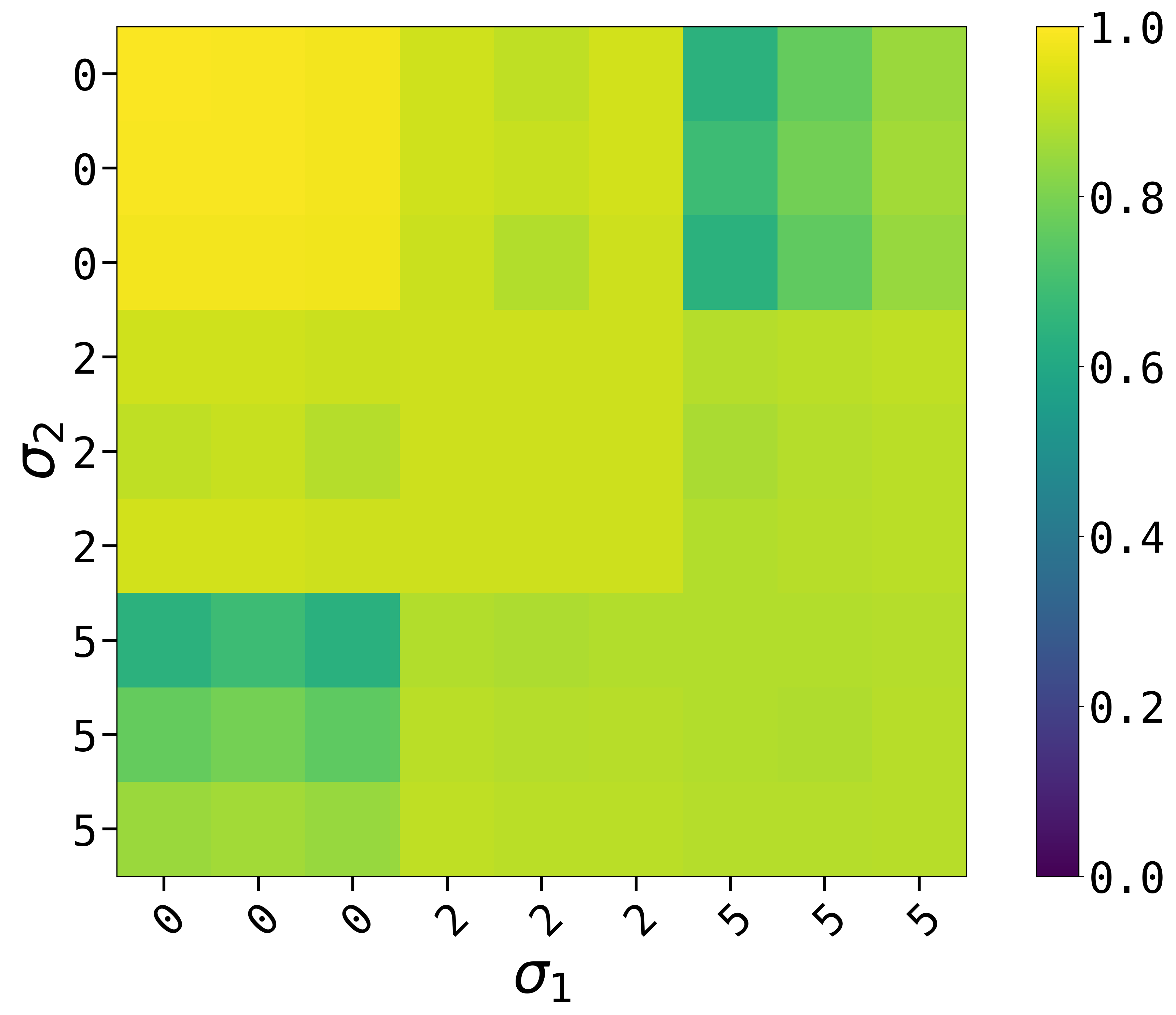}
\caption{NZSC: Terminal Step Coordination Rate}
\label{fig:ape_randomization_rew2_op}
\end{subfigure}
\begin{subfigure}{\figurewidth}
\includegraphics[width=\textwidth]{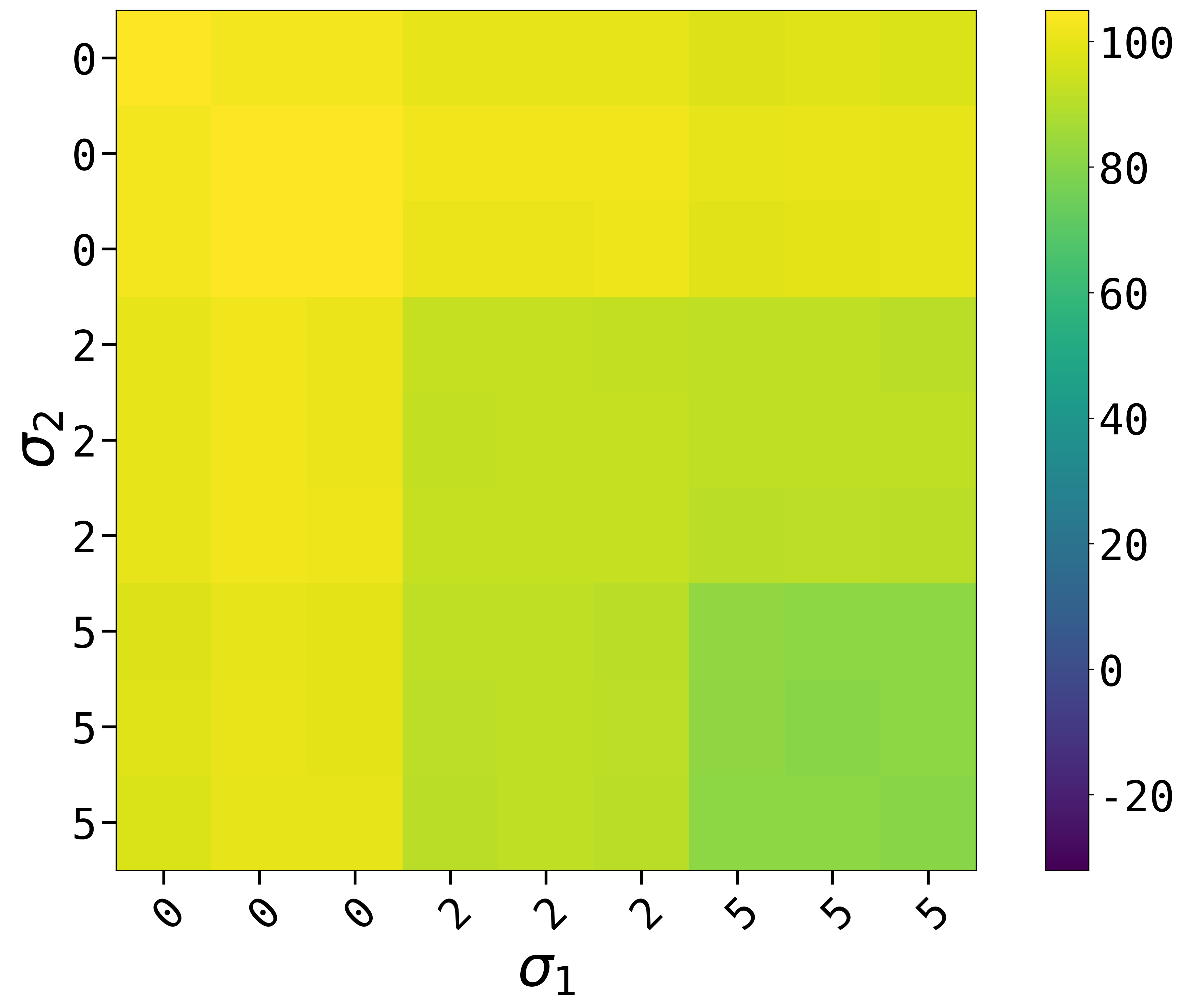}
\caption{Meta-NZSC: Cross-Play Return\addspace}
\label{fig:ape_self_play_rew_rand}
\end{subfigure}
\hspace{25pt}
\begin{subfigure}{\figurewidth}
\includegraphics[width=\textwidth]{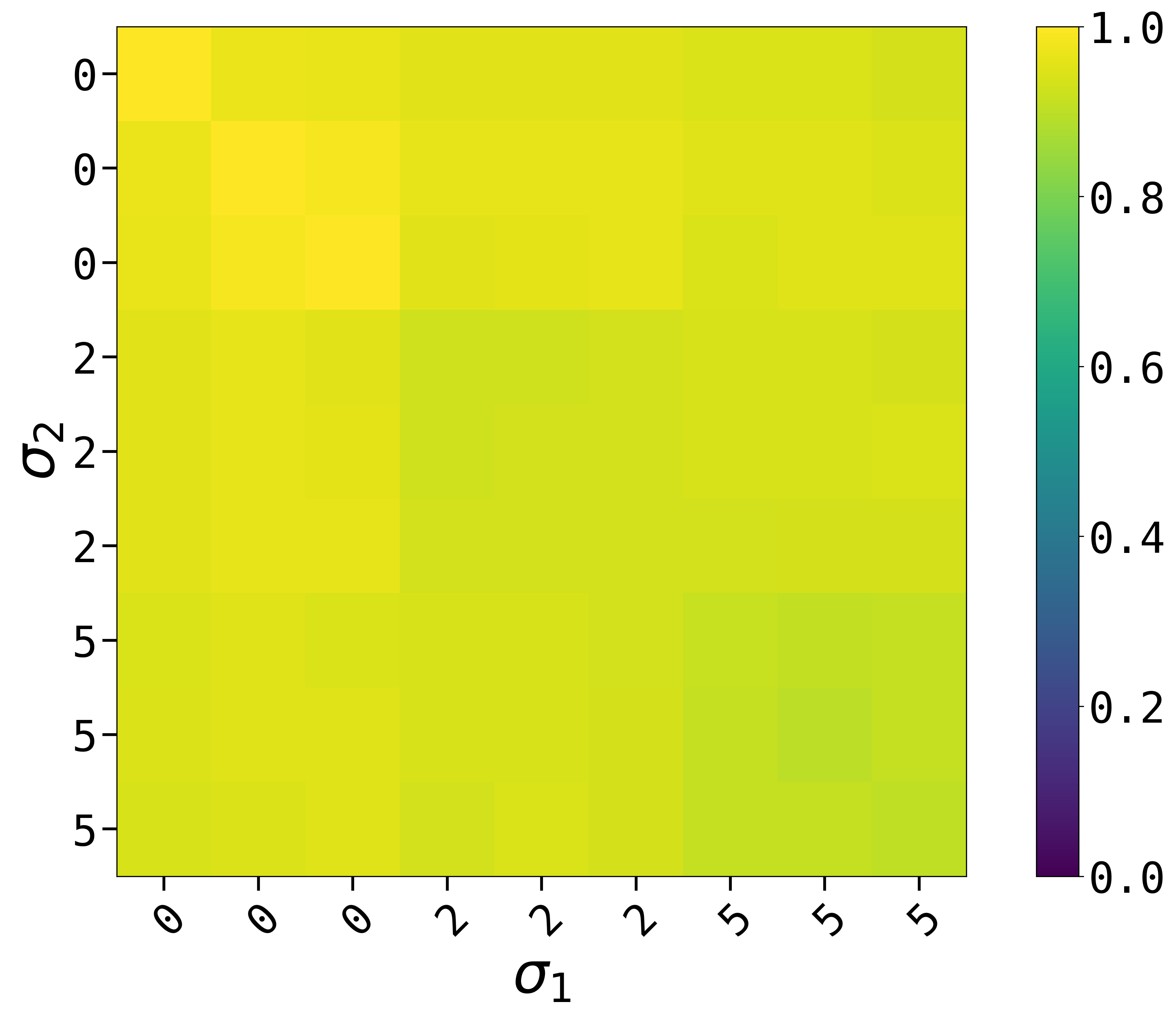}
\caption{Meta-NZSC: Terminal Step Coordination Rate}
\label{fig:ape_randomization_rew2_rand}
\end{subfigure}
\caption{Iterated noisy lever game results.}
\end{figure}

\clearpage
\subsection{Coordinated Exploration Environment}
In the following figure, we show cross-play return and final step coordination rate for different types of agents trained in coordinated exploration environment.
Note that while meta-NZSC trained agents are better at coordinating relative to other type of agents, even they are not able to perfectly coordinate in this environment by the final timestep all the time.

\ifbool{singcol}{}{\vspace{5em}}
\begin{figure}[H]
  \centering
  \begin{subfigure}{\figurewidth}
    \includegraphics[width=\textwidth]{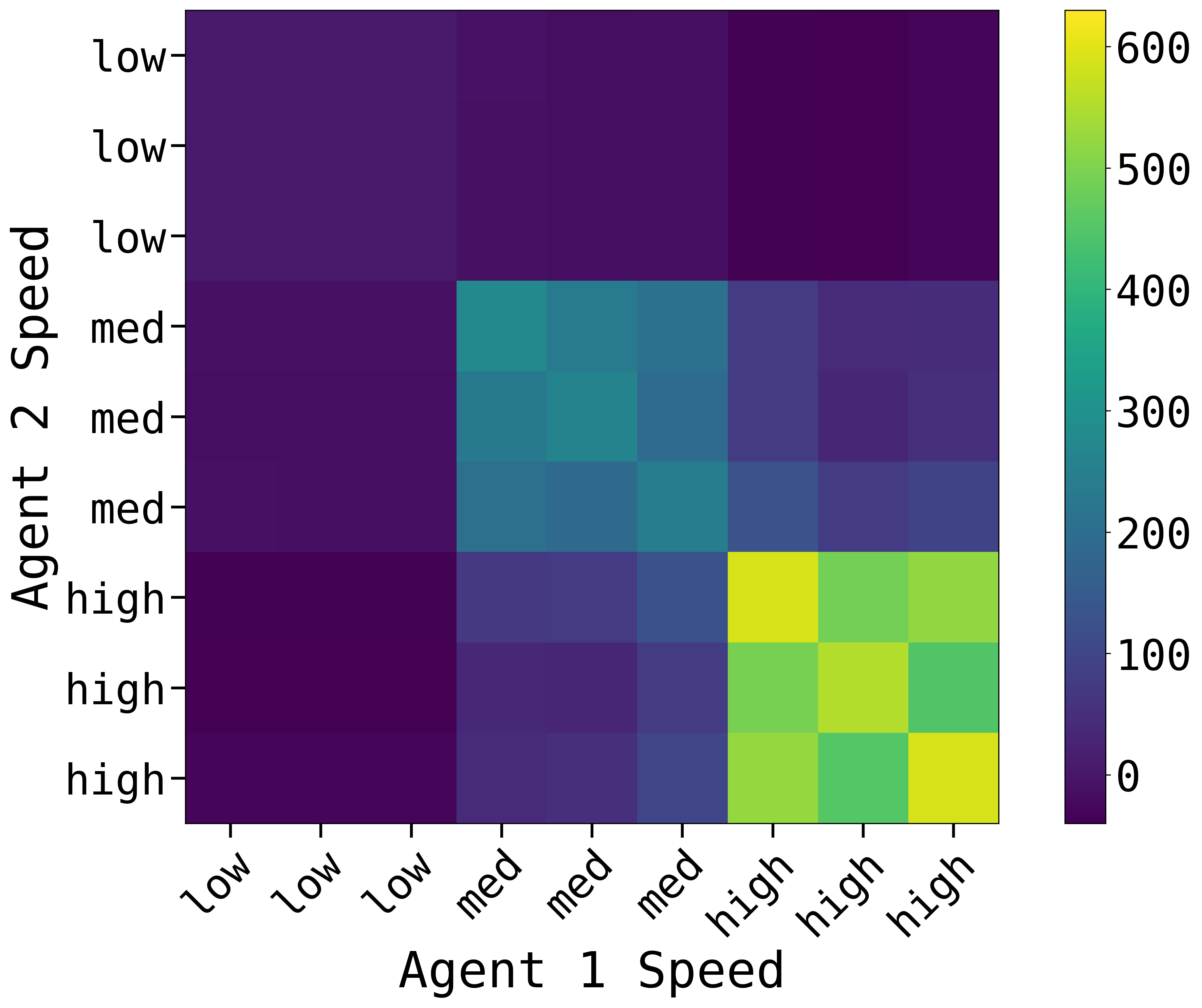}
    \caption{Self Play: Cross-Play Return\addspace}
    \label{fig:ape_self_play_rew}
  \end{subfigure}
  \hspace{25pt}
  \begin{subfigure}{\figurewidth}
    \includegraphics[width=\textwidth]{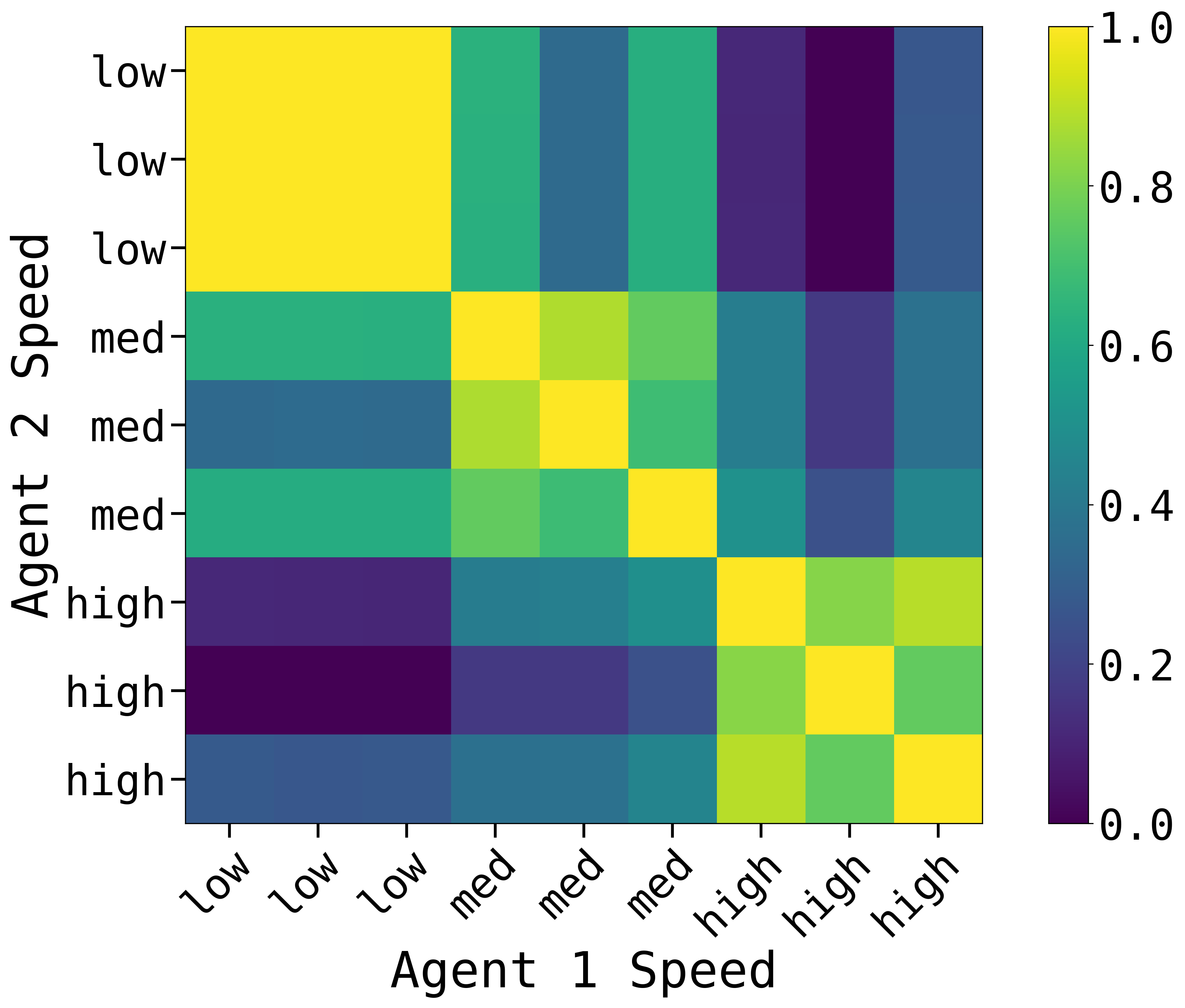}
    \caption{Self Play: Terminal Step Coordination Rate}
    \label{fig:ape_randomization_rew2}
  \end{subfigure}
  
  \begin{subfigure}{\figurewidth}
    \includegraphics[width=\textwidth]{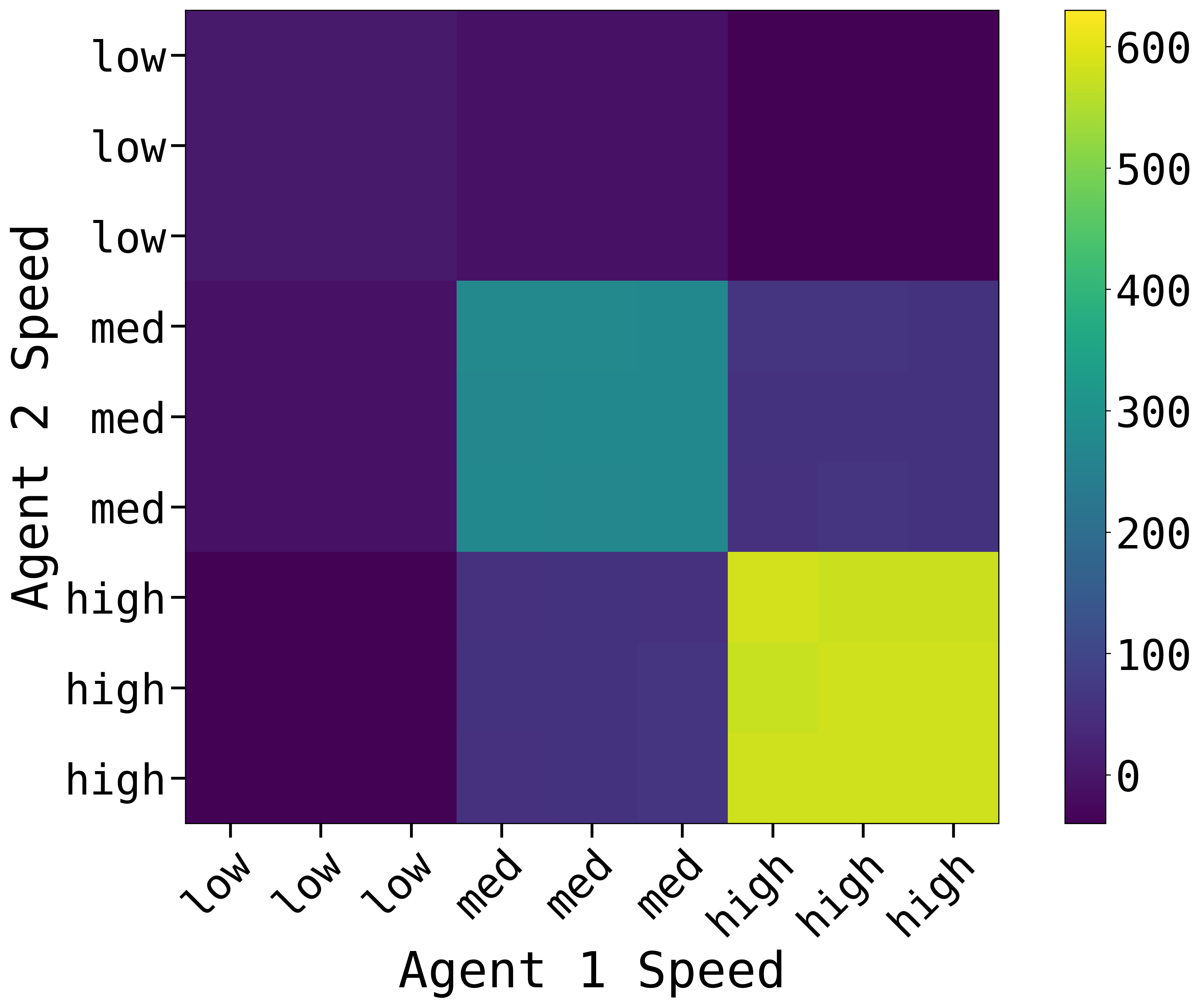}
    \caption{NZSC: Cross-Play Return\addspace}
    \label{fig:ape_self_play_rew_op}
  \end{subfigure}
  \hspace{25pt}
  \begin{subfigure}{\figurewidth}
    \includegraphics[width=\textwidth]{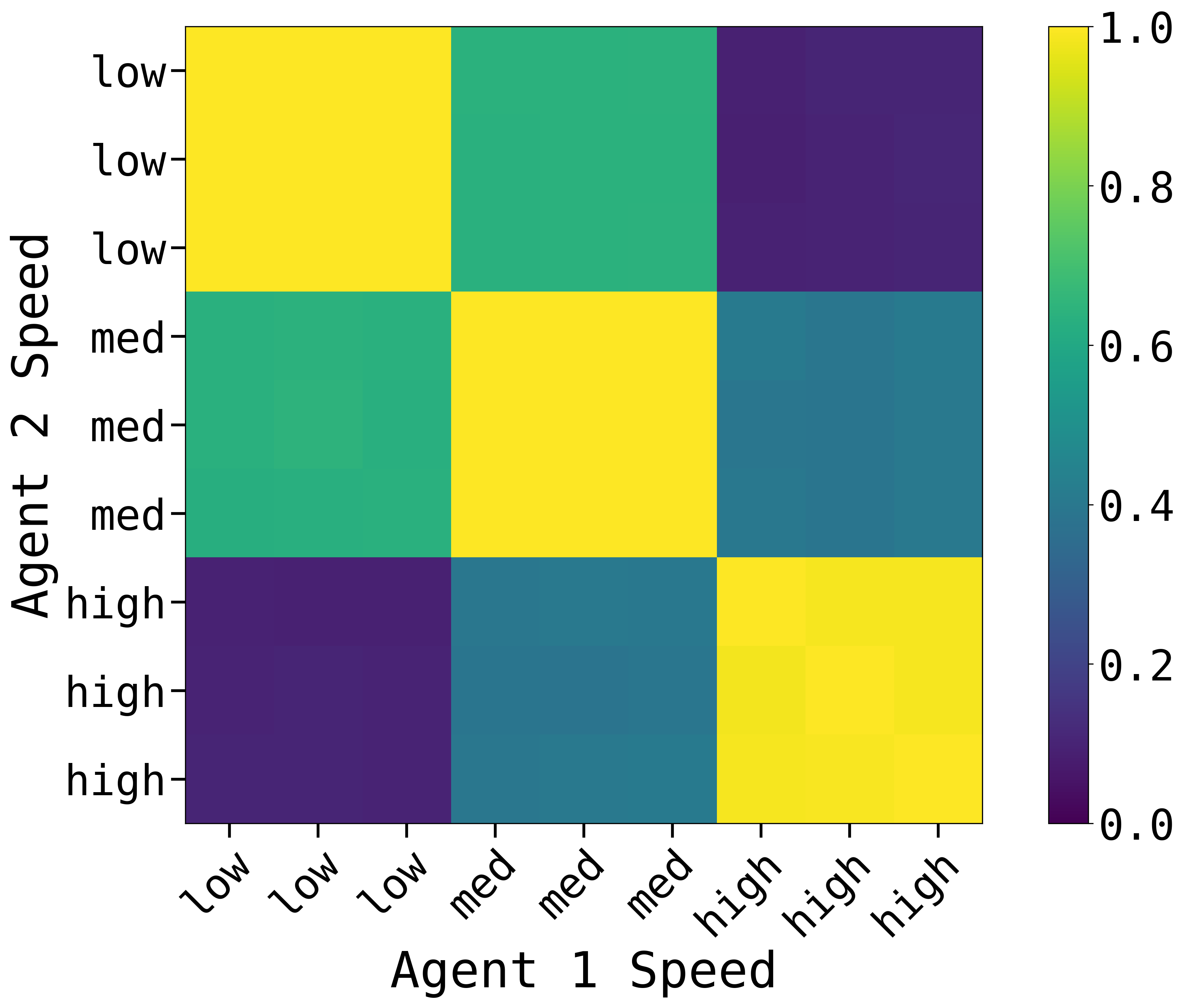}
    \caption{NZSC: Terminal Step Coordination Rate}
    \label{fig:ape_randomization_rew2_op}
  \end{subfigure}
  
  \begin{subfigure}{\figurewidth}
    \includegraphics[width=\textwidth]{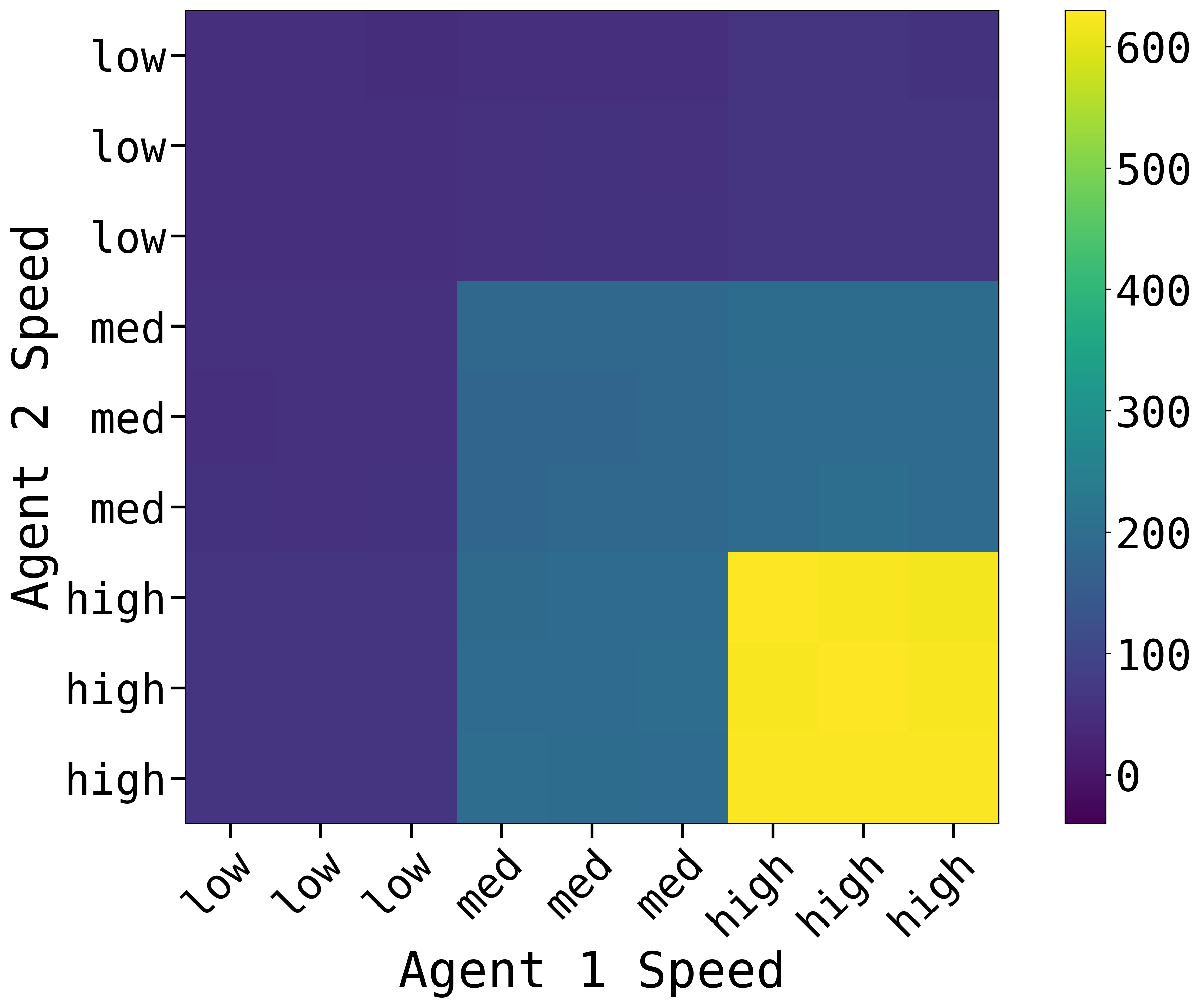}
    \caption{Meta-NZSC: Cross-Play Return\addspace}
    \label{fig:ape_self_play_rew_rand}
  \end{subfigure}
  \hspace{25pt}
  \begin{subfigure}{\figurewidth}
    \includegraphics[width=\textwidth]{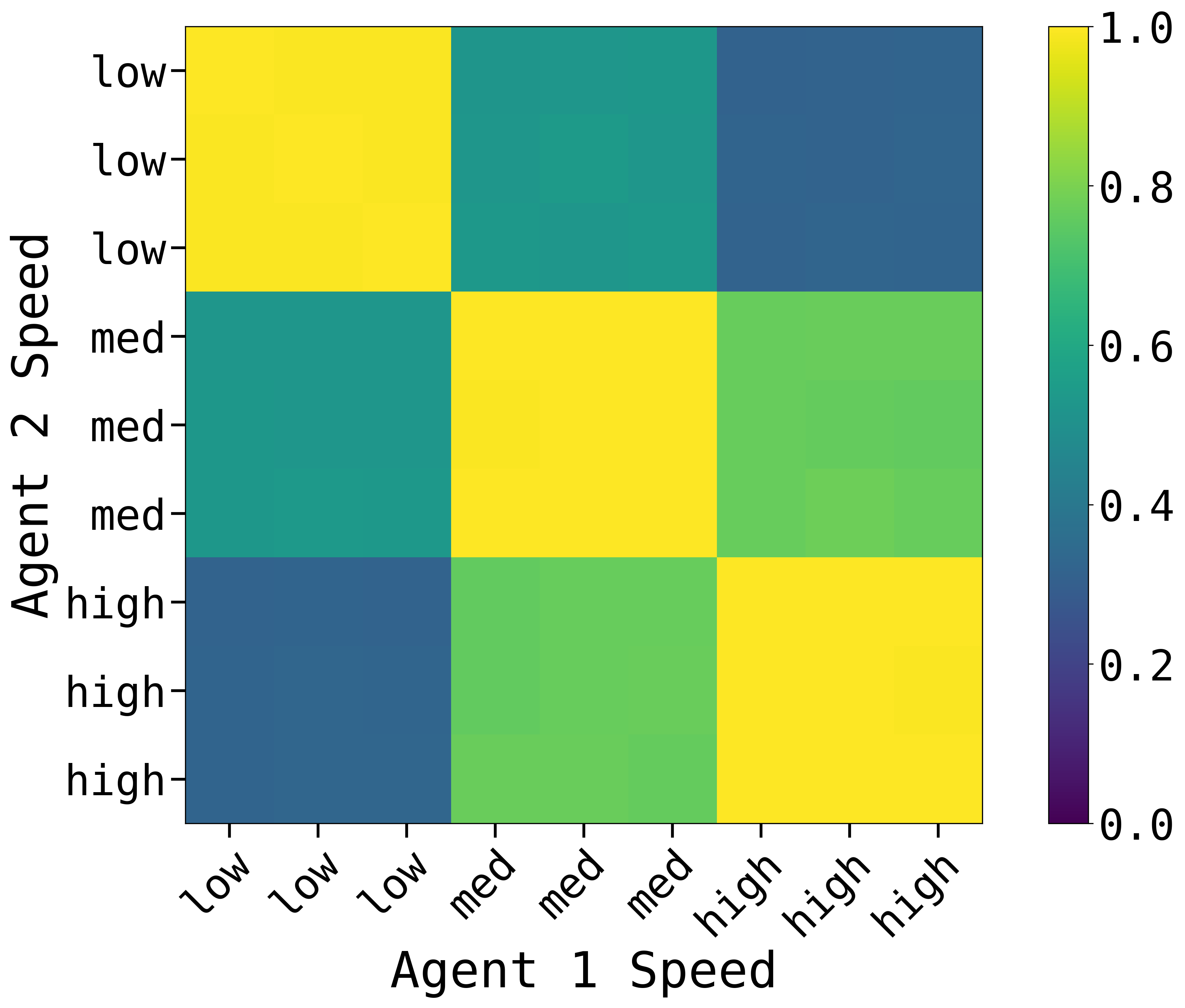}
    \caption{Meta-NZSC: Terminal Step Coordination Rate}
    \label{fig:ape_randomization_rew2_rand}
  \end{subfigure}
  \caption{Coordinated Exploration Environment results.}
\end{figure}

\clearpage
\subsection{SyncSight Environment}
In the following figure, we show cross-play return and final step coordination rate for different types of agents trained in syncsight environment.

\ifbool{singcol}{}{\vspace{5em}}
\begin{figure}[H]
  \centering
  \begin{subfigure}{\figurewidth}
    \includegraphics[width=\textwidth]{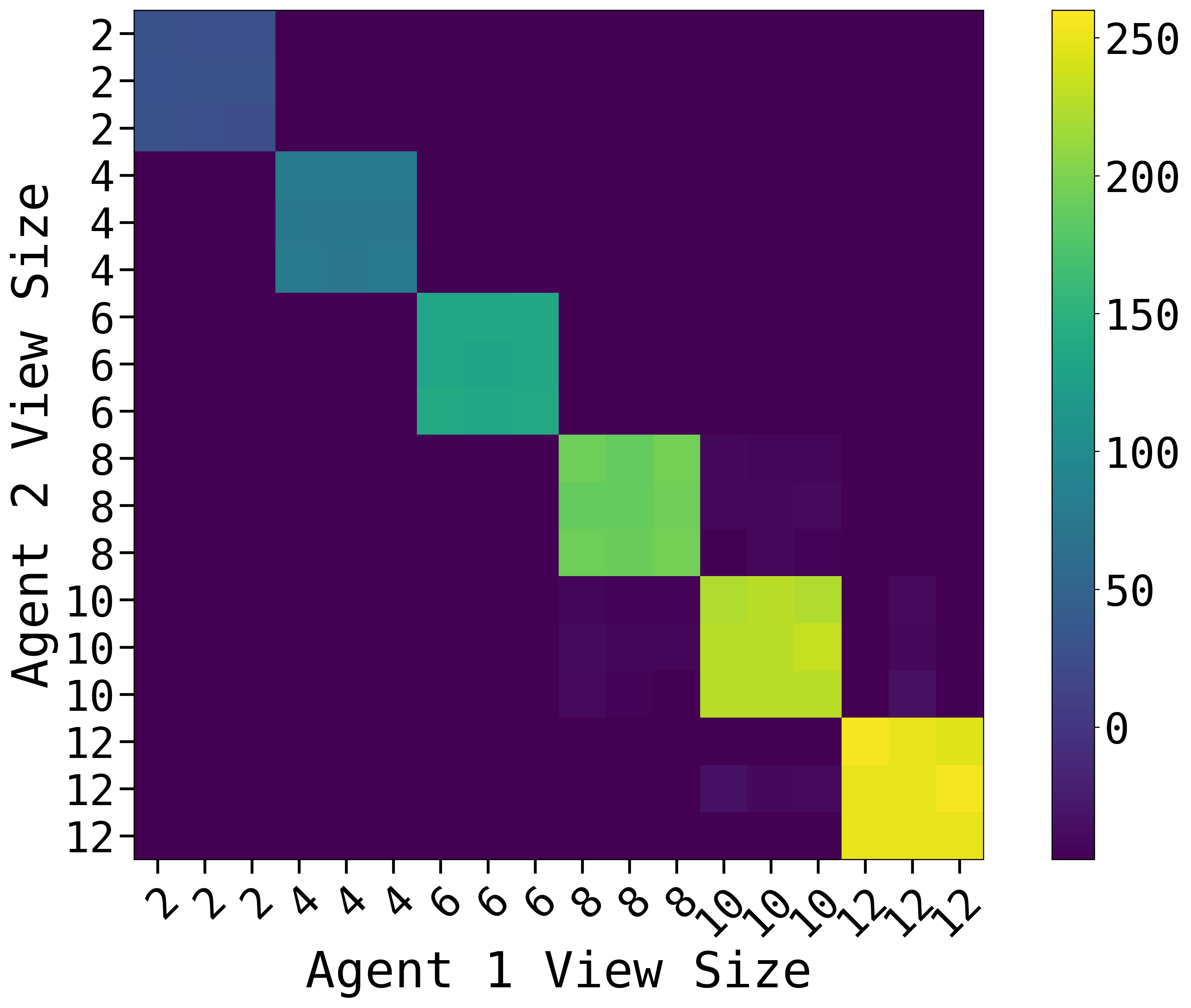}
    \caption{Self Play: Cross-Play Return\addspace}
    \label{fig:ape_self_play_rew}
  \end{subfigure}
  \hspace{25pt}
  \begin{subfigure}{\figurewidth}
    \includegraphics[width=\textwidth]{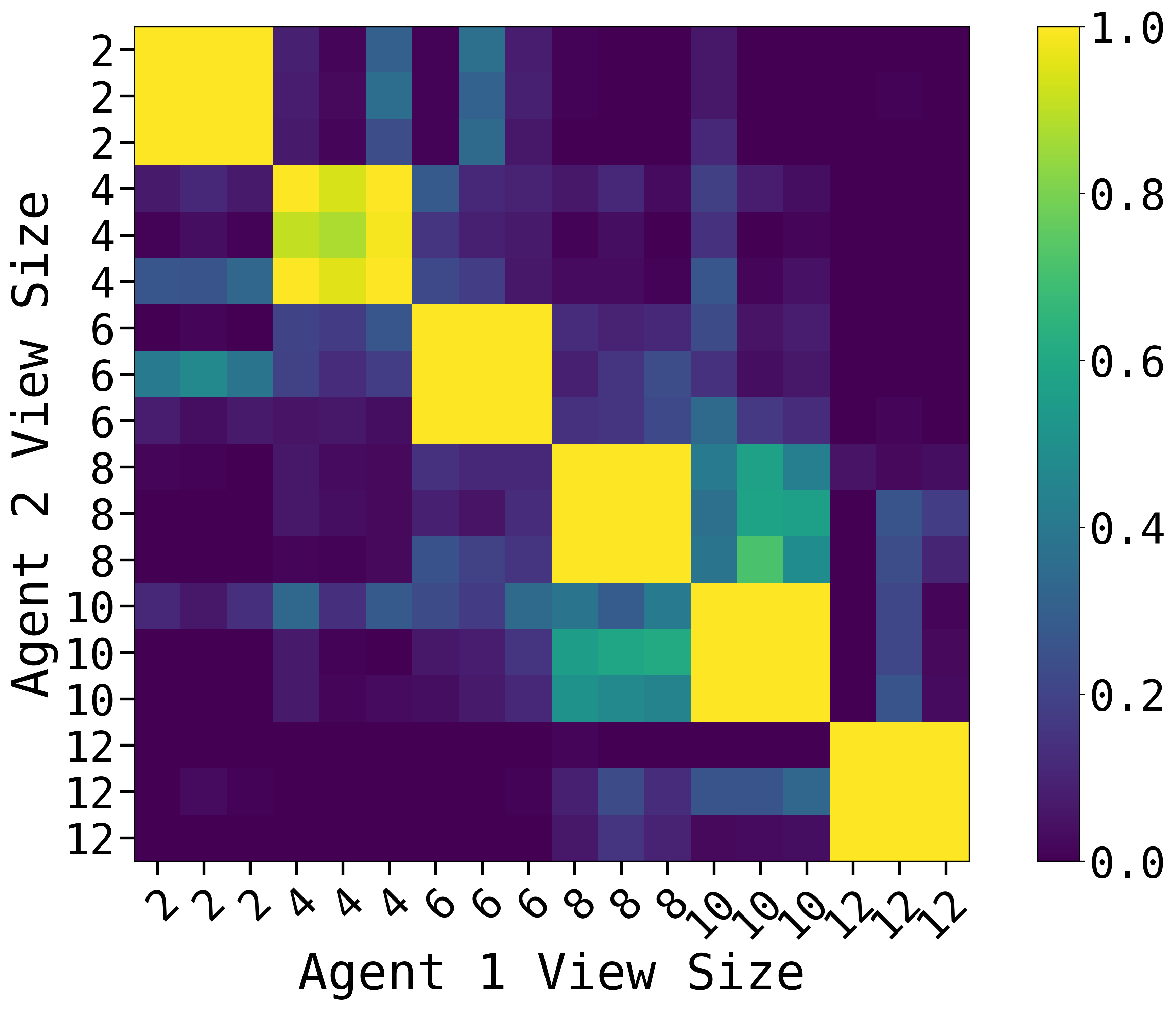}
    \caption{Self Play: Terminal Step Coordination Rate}
    \label{fig:ape_randomization_rew2}
  \end{subfigure}
  
  \begin{subfigure}{\figurewidth}
    \includegraphics[width=\textwidth]{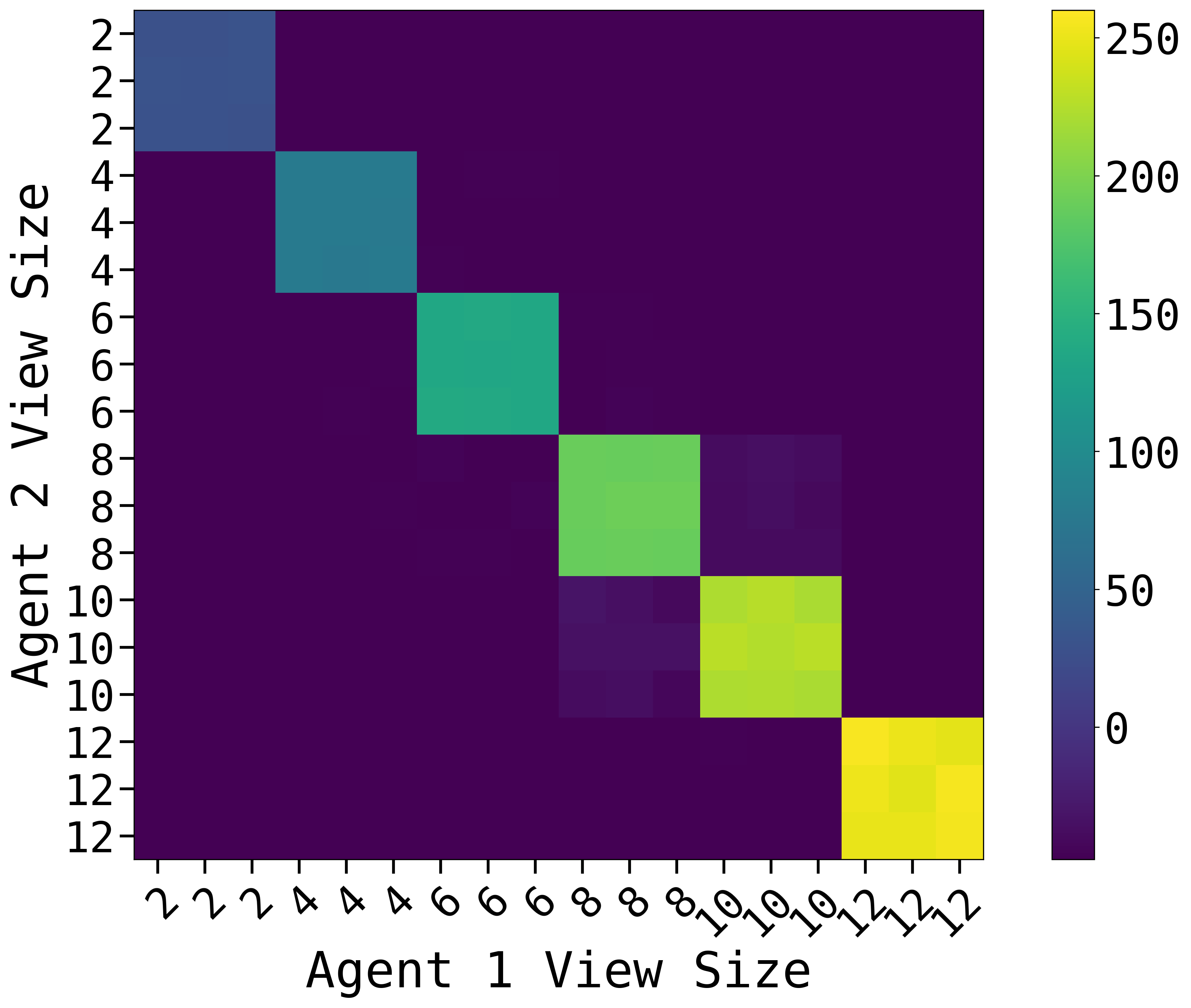}
    \caption{NZSC: Cross-Play Return\addspace}
    \label{fig:ape_self_play_rew_op}
  \end{subfigure}
  \hspace{25pt}
  \begin{subfigure}{\figurewidth}
    \includegraphics[width=\textwidth]{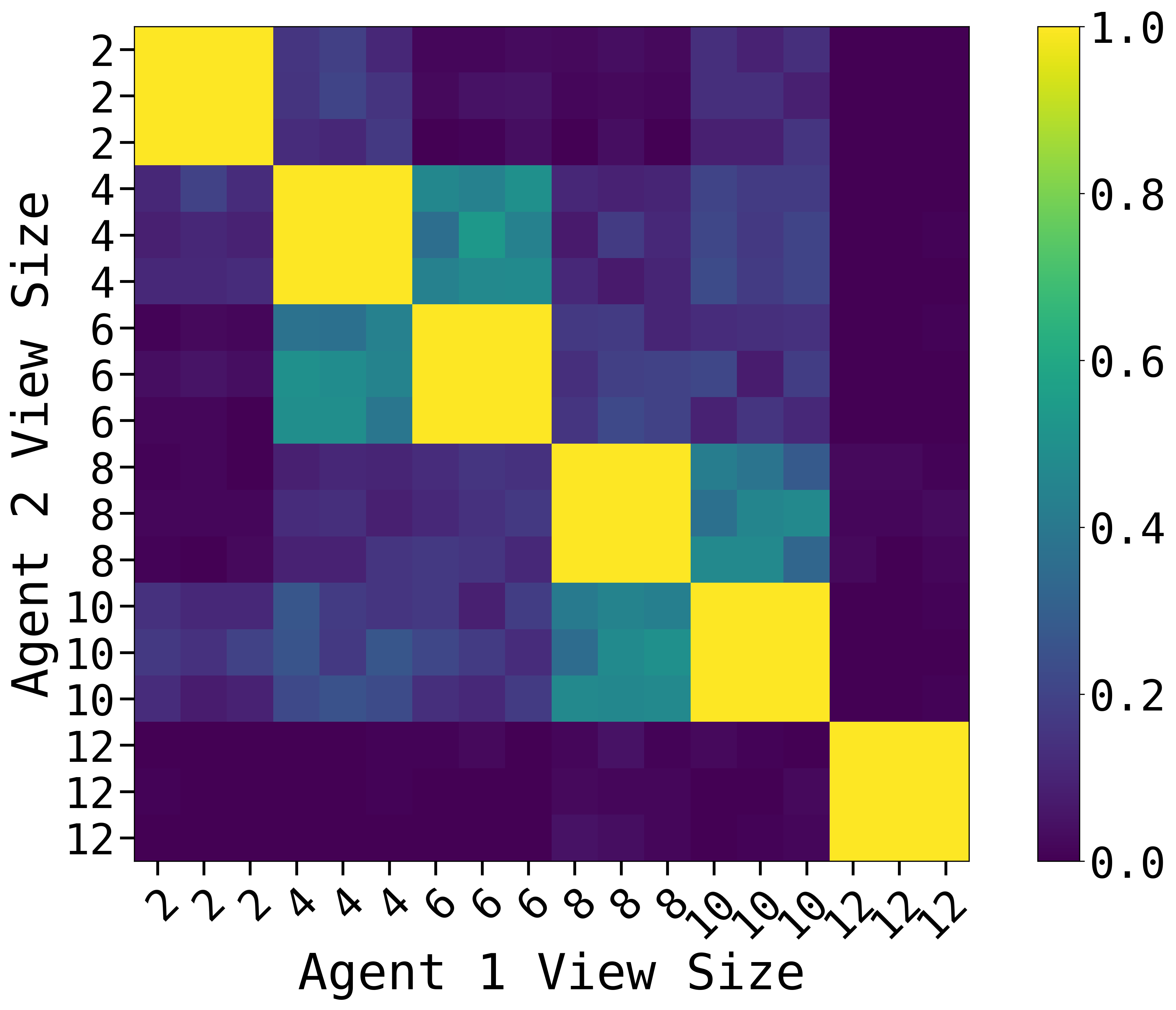}
    \caption{NZSC: Terminal Step Coordination Rate}
    \label{fig:ape_randomization_rew2_op}
  \end{subfigure}
  
  \begin{subfigure}{\figurewidth}
    \includegraphics[width=\textwidth]{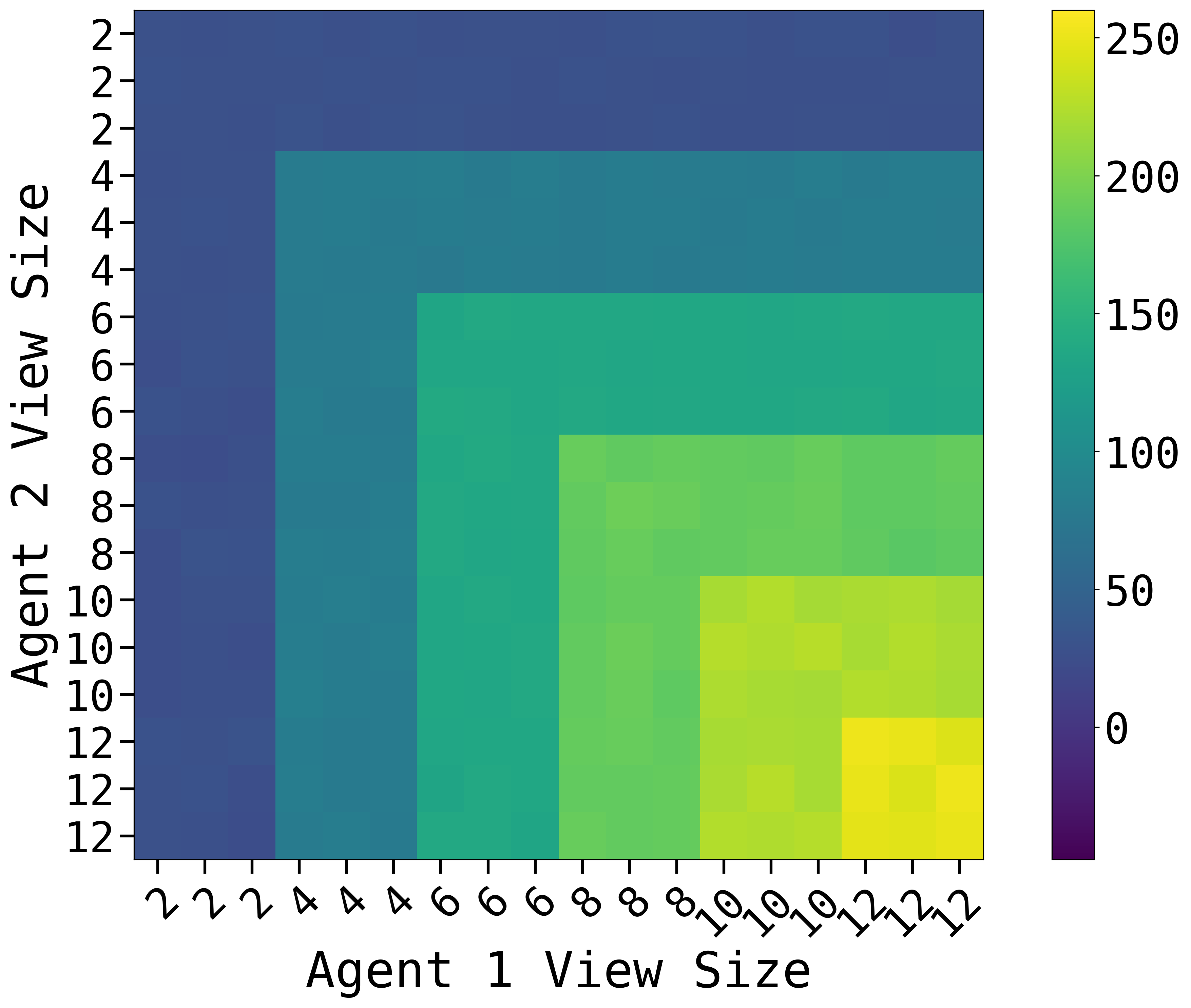}
    \caption{Meta-NZSC: Cross-Play Return\addspace}
    \label{fig:ape_self_play_rew_rand}
  \end{subfigure}
  \hspace{25pt}
  \begin{subfigure}{\figurewidth}
    \includegraphics[width=\textwidth]{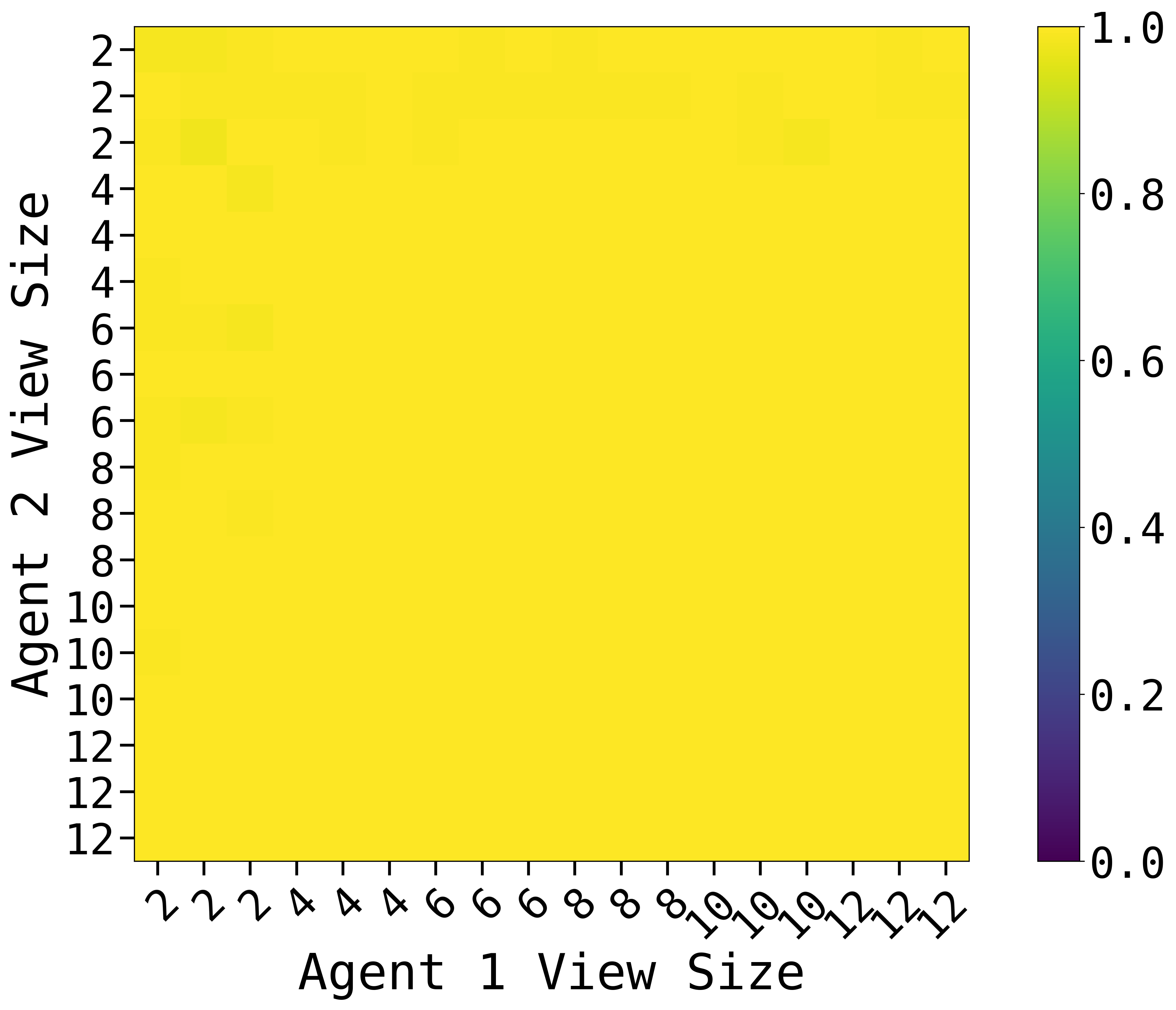}
    \caption{Meta-NZSC: Terminal Step Coordination Rate}
    \label{fig:ape_randomization_rew2_rand}
  \end{subfigure}
  
  \caption{SyncSight Environment results.}
\end{figure}

\end{document}